%% file: thesis.tex
\documentclass[12pt]{thesis}
\usepackage{graphicx} 
\usepackage[linesnumbered,ruled]{algorithm2e}
\usepackage{amsmath}
\usepackage{amssymb}
\usepackage{amsthm}
\usepackage{subcaption}
\usepackage{diagbox}
\usepackage{array}
\usepackage{booktabs}
\usepackage{multirow}
\usepackage{graphicx}
\usepackage{longtable}
\usepackage{comment}
\usepackage{todonotes}
\usepackage{ulem}
\usepackage[flushleft]{threeparttable}
\usepackage{bm}
\usepackage{url}

\newtheorem{theorem}{Theorem}[section]
\newtheorem{lemma}{Lemma}[section]
\newtheorem{corollary}{Corollary}[lemma]
\newtheorem{definition}{Definition}[section]
\newtheorem{remark}{Remark}

\begin{document}

\hyphenpenalty=1000
\tolerance=1000
\exhyphenpenalty=1000
\uchyph=0
\looseness=2



\abstract{\input{abstract}}
\input{approval}

\input{biography}
\dedication{\input{dedication}}
\acknowledgment{\input{acknowledgment}}

\initthesis


\tableofcontents
\listoftables
\listoffigures


\startbody

\input{chapter1}

\input{chapter2}

\input{chapter3}

\input{chapter4}

\input{chapter5}

\input{chapter6}


\bibliography{references}

\end{document}

%% file: abstract.tex
\noindent
The world has never been more connected, led by the information technology revolution in the past decades that has fundamentally changed the way people interact with each other using social networks. Consequently, enormous human activity data are collected from the business world and machine learning techniques are widely adopted to aid our decision processes. Despite of the success of machine learning in various application scenarios, there are still many questions that need to be well answered, such as optimizing machine learning outcomes when desired knowledge cannot be extracted from the available data. 
This naturally drives us to ponder if one can leverage some side information to populate the knowledge domain of their interest, such that the problems within that knowledge domain can be better tackled. 

In this work, such problems are investigated and practical solutions are proposed. 
To leverage machine learning in any decision-making process, one must convert the given knowledge (for example, natural language, unstructured text) into representation vectors that can be understood and processed by machine learning model in their compatible language and data format. The frequently encountered difficulty is, however, the given knowledge is not rich or reliable enough in the first place. In such cases, one seeks to fuse side information from a separate domain to mitigate the gap between good representation learning and the scarce knowledge in the domain of interest. This approach is named Cross-Domain Knowledge Transfer. It is crucial to study the problem because of the commonality of scarce knowledge in many scenarios, from online healthcare platform analyses to financial market risk quantification, leaving an obstacle in front of us benefiting from automated decision making. From the machine learning perspective,  the paradigm of semi-supervised learning takes advantage of large amount of data without ground truth and achieves impressive learning performance improvement. It is adopted in this dissertation for cross-domain knowledge transfer.

Furthermore, graph learning techniques are indispensable given that networks commonly exist in real word, such as taxonomy networks and scholarly article citation networks. These networks contain additional useful knowledge and are ought to be incorporated in the learning process, which serve as an important lever in solving the problem of cross-domain knowledge transfer. This dissertation proposes graph-based learning solutions and demonstrates their practical usage via empirical studies on real-world applications. Another line of effort in this work lies in leveraging the rich capacity of neural networks to improve the learning outcomes, as we are in the era of big data. 

In contrast to many Graph Neural Networks that directly iterate on the graph adjacency to approximate graph convolution filters, this work also proposes an efficient Eigenvalue learning method that directly optimizes the graph convolution in the spectral space. This work articulates the importance of network spectrum and provides detailed analyses on the spectral properties in the proposed EigenLearn method, which well aligns with a series of GNN models that attempt to have meaningful spectral interpretation in designing graph neural networks. The dissertation also addresses the efficiency, which can be categorized in two folds. First, by adopting approximate solutions it mitigates the complexity concerns for graph related algorithms, which are naturally quadratic in most cases and do not scale to large datasets. Second, it mitigates the storage and computation overhead in deep neural network, such that they can be deployed on many light-weight devices and significantly broaden the applicability. Finally, the dissertation is concluded by future endeavors. 

%% file: approval.tex

\advisor{Dantong Yu}{Associate Professor, Martin Tuchman School of Management, NJIT}

\member{Ioannis Koutis}
       {Associate Professor, Department of Computer Science, NJIT}

\member{Baruch M. Schieber}
       {Professor and Chair, Department of Computer Science, NJIT}
	
\member{Yi Chen}
       {Professor, Martin Tuchman School of Management, NJIT}
	
\member{Junmin Shi}
       {Associate Professor, Martin Tuchman School of Management, NJIT}

%% file: biography.tex



\title{Graph Enabled Cross-domain Knowledge Transfer}

\author{Shibo Yao}

\major{Business Data Science}

\defensedate{May}{9}{2022}



\department{Martin Tuchman School of Management}


\degree{Bachelor of Management Science,}
       {University of Science and Technology of China, Hefei, 2015}

\degree{Master of Science,}
       {Stony Brook University, Stony Brook, NY, 2016}
       

\publications{
        {Shibo Yao, Dantong Yu and Keli Xiao, 
        ``Enhancing Domain Word Embedding with Latent Semantic Imputation,'' 
        {\em Proceedings of the 25th ACM SIGKDD International Conference on Knowledge Discovery \& Data Mining}, 2019.}, 
        {Shibo Yao, Dantong Yu and Xiangmin Jiao,
        ``Perturbing Eigenvalues with Residual Learning in Graph Neural Networks,''
        {\em Proceedings of The 13th Asian Conference on Machine Learning}, PMLR 2021.},
        {Uras Varolgüneş, Shibo Yao, Yao Ma and Dantong Yu,
        ``Embedding Imputation with Self-Supervised Graph Neural Networks,''
        {\em IEEE Access}, 2023.},
	{Shibo Yao, Dantong Yu and Ioannis Koutis,
	 ``Neural Network Pruning as Spectrum Preserving Process,''
	{\em submitted to ACM TKDD}, 2021.},
	{Shibo Yao, Scott Ferson and Keli Xiao
	 ``Does weather affect financial markets? Evidence in dew point temperature and market volatility,''
	{\em Imprecision Friday}, seminar presentation, unpublished work, 2016.},
	{Shibo Yao, and Dantong Yu
	 ``Quantifying Heterogeneity in Financial Time Series for Improved Prediction,''
	{\em The 6th Applied Financial Modeling Conference}, 2018.},	
}

%% file: dedication.tex
\singlespace
\begin{center}
\begin{minipage}{4in}
{\em Simplicity is the final achievement.}
\flushright
Frederic Chopin
\end{minipage}      
\end{center}

%% file: acknowledgment.tex
\noindent
First and foremost, I would like to express my sincere gratitude to my advisor, Prof. Dantong Yu. 
Throughout my Ph.D. journey, Prof. Yu patiently listened my thoughts on various research topics and gave me numerous pieces of priceless advice. He gave me substantial flexibility to explore and broaden my research interests, and shaped me into a better researcher and person  over the years. I could not be more thankful for his support, guidance and encouragement to complete this journey. I would like to thank my committee members, Baruch M. Schieber, Ioannis Koutis, Yi Chen and Junmin Shi, for their kindness and insightful feedback during my journey on the dissertation. I would like to thank Prof. Schieber and Prof. Koutis for the generous office discussions and making complicated stuff simple and fun. 

Many thanks go to Professor Keli Xiao, Professor Xiangmin Jiao, Dr. Dimitrios Katramatos, Professor Eden Figueroa and Professor Yao Ma. I am grateful for the collaboration opportunities and benefit significantly from their guidance on my research exploration and technical writing. I am thankful for Dr. Yufei Ren, Professor Hao Zhong and Dr. Mingda Li giving me kind advice on doctoral study. I would also like to express my gratitude to Dr. Youzhong Wang who was my internship manager at Facebook and my colleague Haipeng Guan, for making my first industry intern experience awesome. Among many, Professor Scott D. Ferson shed light on my life with his wisdom. 

Hearty thanks are owed for the research support from the Provost Assistantship throughout my first two years of PhD study. 

%% file: chapter1.tex
\chapter{Introduction}


\section{Background Story}
The early investigation of this dissertation was dedicated to the study of financial market behaviors and most of the time was spent on a Bloomberg terminal. With the large amount of data retrieved from the largest financial database and the newly emerging machine learning techniques, we witness  encouraging results in predicting financial market trends given the seasonal firm disclosures, especially when the heterogeneous time series is considered  and appropriately quantified~\cite{ml-fin}~\cite{weather2016}. As a natural continuation, we managed to incorporate more information from the Bloomberg terminal, such as the useful information contained in the indulgent textual data. And that was where a more challenging problem kicked in. 

The key precursor for natural language processing is to find appropriate representations in a vector space for words and phrases such that models understand our communication system built upon these basic units. 
It was not long after the publication of the famous work that introduces word2vec, a technique that uses neural networks to assign human words semantically meaningful vectors for natural language processing tasks. To further improve our work~\cite{ml-fin}, we tried to acquire word embeddings for financial terminologies using word2vec and wikipages as the training corpus. It turned out that the general embedding technique was unable to handle a large number of terminologies because of the existing low-frequency words. When we visualized the terminology embeddings in a 2-d plane using TSNE, the terms supposed to have similar semantic meanings were not close to each other. This  observation is counter intuitive and indicates the low quality embedding. 

The problem hindered the progress, until we discovered another dataset that included most of the financial terminologies and their well organized financial attributes. The text corpus used for training embedding and the financial attributes appeared to be from two unrelated domains. Nevertheless, we were curious if one can be fused to the other. After all, they somehow constituted two different sets of knowledge that are complimentary to each other. After some serious brainstorming, we instantiated the investigation of cross-domain knowledge transfer, dived deeper into the methodology aspect in searching for better solutions to the problem, and found the practical usage in financial domain applications and many more. 

The larger background in nowadays' machine learning practice is that network-based approach is the key to many problems. 
\begin{itemize}
    \item When we want to advertise products on social networks, we can train a dedicated model that predicts if a user will click on the advertisement or not based on their recent activities. For those users that are not active recently, we can leverage the social graph and user identity information to infer what their taste is and what they may be in need of. 
    \item In financial market risk quantification, one can utilize supply chain network, company information and stock market performance to reveal hidden facts about companies. One example is that if we visualize companies using the aforementioned information, we can find that Walmart is closer to financial companies. And the reason underneath is how the company profits in a similar way to financial companies. This way, we have better assessment on financial risks for certain firms. 
    \item Natural language processing in healthcare could be difficult since there are many terminologies and abbreviations that do not have reliable representations for language models to read. In this case, one can treat taxonomies built by healthcare experts as graphs and transfer domain knowledge to enhance the representations.  
\end{itemize}

The research motivation will start in the following sections with more technical background. 

\section{Research Motivation}

In the past decade, machine learning has seen prosperity in both academic research and the daily operation in many technology tycoons such as Google and Facebook, due to huge amount of data generated by users and the growing computing power. Depending on the context, machine learning practice is also referred to as personalization techniques, ranking system etc. Ever since human are connected by the powerful social networks, almost every niche problem where machine learning is involved requires a large-scale system. However, repeating end-to-end learning processes for each individual task has been found a waste of time and computation. A smarter solution that has been already adopted is to learn some sort of general representations or embeddings for the given data that can be shared by a category of related downstream learning tasks for better personalization. A concrete example is that in multimedia content consumption, we usually have general pre-trained embeddings (representation vectors) for sentences, pictures, videos, or even abstract objects such as accounts and user cohort.

The learning of such general data representations usually takes rich and high-quality prior knowledge, which is not always readily available in the desired domain. For example, in natural language processing, we need to learn the good embedding vectors for words based on large amount of textual data such that machine learning models can understand human sentences only once they are converted to numbers. However, even the advanced embedding techniques including word2vec, can have missing words and unreliable embeddings, especially when facing domain specific tasks, such as chemistry and healthcare, due to thousands or millions of terminologies and abbreviations. In such situations, it is of key importance if we can transfer some rich knowledge from one domain to another, which in turn aids the representation learning. Besides, the knowledge transfer also relies on semi-supervised learning and efficient graph learning that is capable of quantifying the nonlinearity in data.

\section{Problem Definition and Challenges}

\subsection{Representation Learning and Knowledge Transfer}
Representation learning is the process of finding an appropriate representation, usually a real-valued vector, for the data sample. By ``appropriate" it means two data samples that should be semantically close are positioned near each other in the representation space. It is also subject to the learning context nevertheless. Representation learning serves as an indispensable step in any machine learning system because only this way, the machine learning model can recognize a puppy image, an English poetry or a funny short video.
Word embedding is the most representative representation learning problem, in which we try to associate real-valued vectors to words such that they are meaningfully positioned in the semantic space. Obviously, ``dog" and ``cat" should have the embedding vectors close to each other enough since they are both mammal accompanying human. 

The obstacle in many representation learning problems is the prior knowledge in the given domain is not rich enough. Take, again, word embedding as an example. We often come across domain-specific language tasks such as chemistry and healthcare text analyses, which may involve thousands or even millions of terminologies and abbreviations. These terminologies and abbreviations are hard to learn in the semantic space given the fact that they are low-frequency words in corpus. What if there is some knowledge available in another domain? Can we transfer such knowledge to better learn the embedding vector in the semantic space? This dissertation will tackle the challenge with semi-supervised learning. 

\subsection{Semi-supervised Learning Given Limited Ground Truth}
Another challenge in the problem of knowledge transfer from one domain to another is that there is often a limited number of ground truths. In the case of domain language task, the number of missing embeddings can be significantly larger than the known ones, where the efficacy of regular supervised learning deteriorates. To address this challenge, we can leverage the large amount of samples without ground truth and semi-supervised learning paradigm. 

Formally, given $\{x_q\}$ and $\{x_p\}$ which denote the sets of feature vectors of unlabeled and labeled samples respectively, 
and $\{y_p\}$ which denotes the set of labels associated with $\{x_p\}$,
we want to infer the labels $\{y_q\}$ for $\{x_q\}$. By ``label" it is not necessarily the label defined within the context of machine learning, but rather a more general term that points to some representation vector defined in another domain. Then we are left with the question of what semi-supervised learning models can solve the problem effectively and efficiently. 

\subsection{Graph, Nonlinearity and Efficiency}
\begin{figure}[htbp]
    \centering
    \includegraphics[scale=0.6]{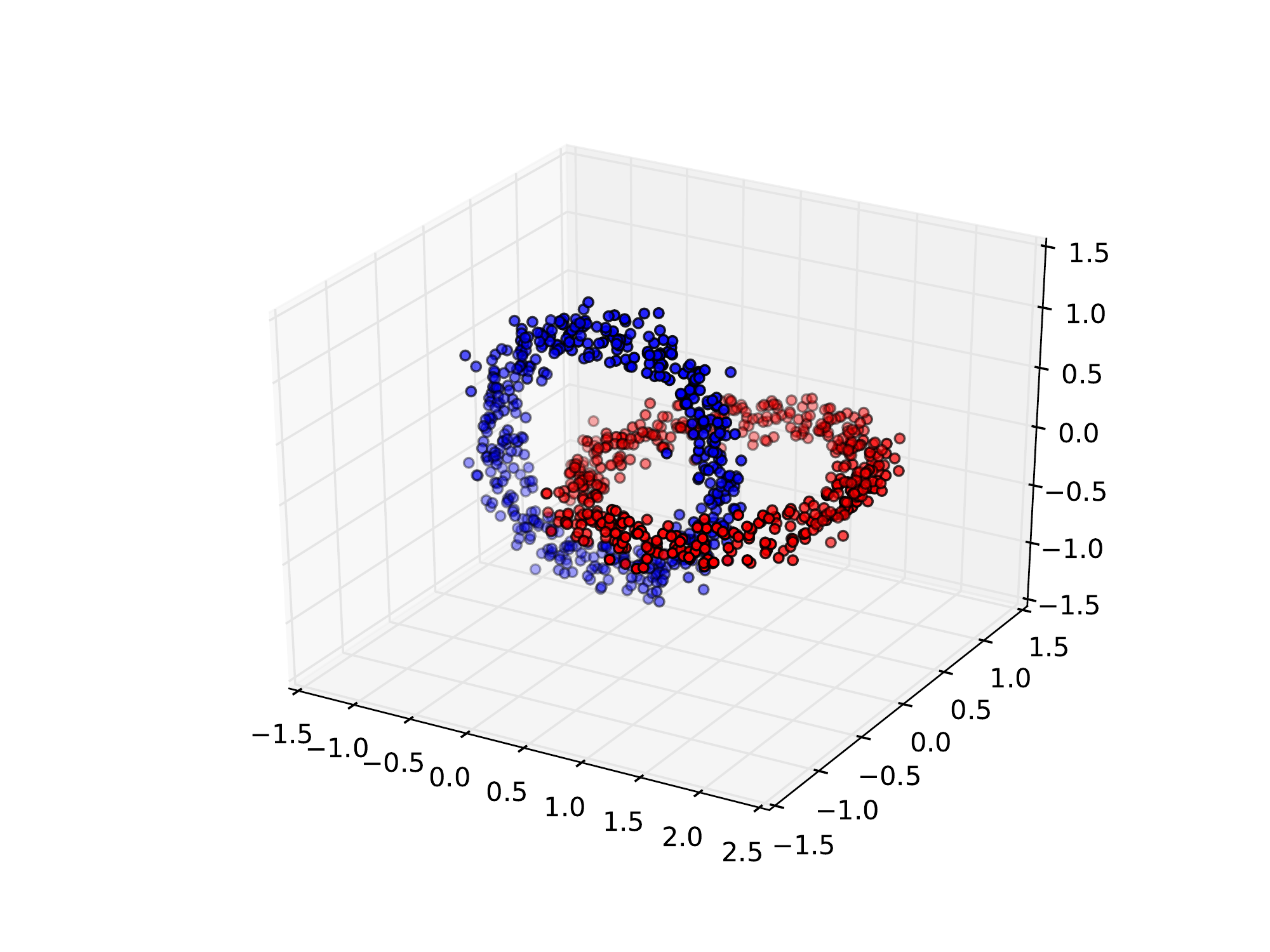}
    \caption{The Twin-Ring example of manifold.}
    \label{fig:twin3d}
\end{figure}

\begin{figure}[htbp]
    \centering
    \includegraphics[scale=0.6]{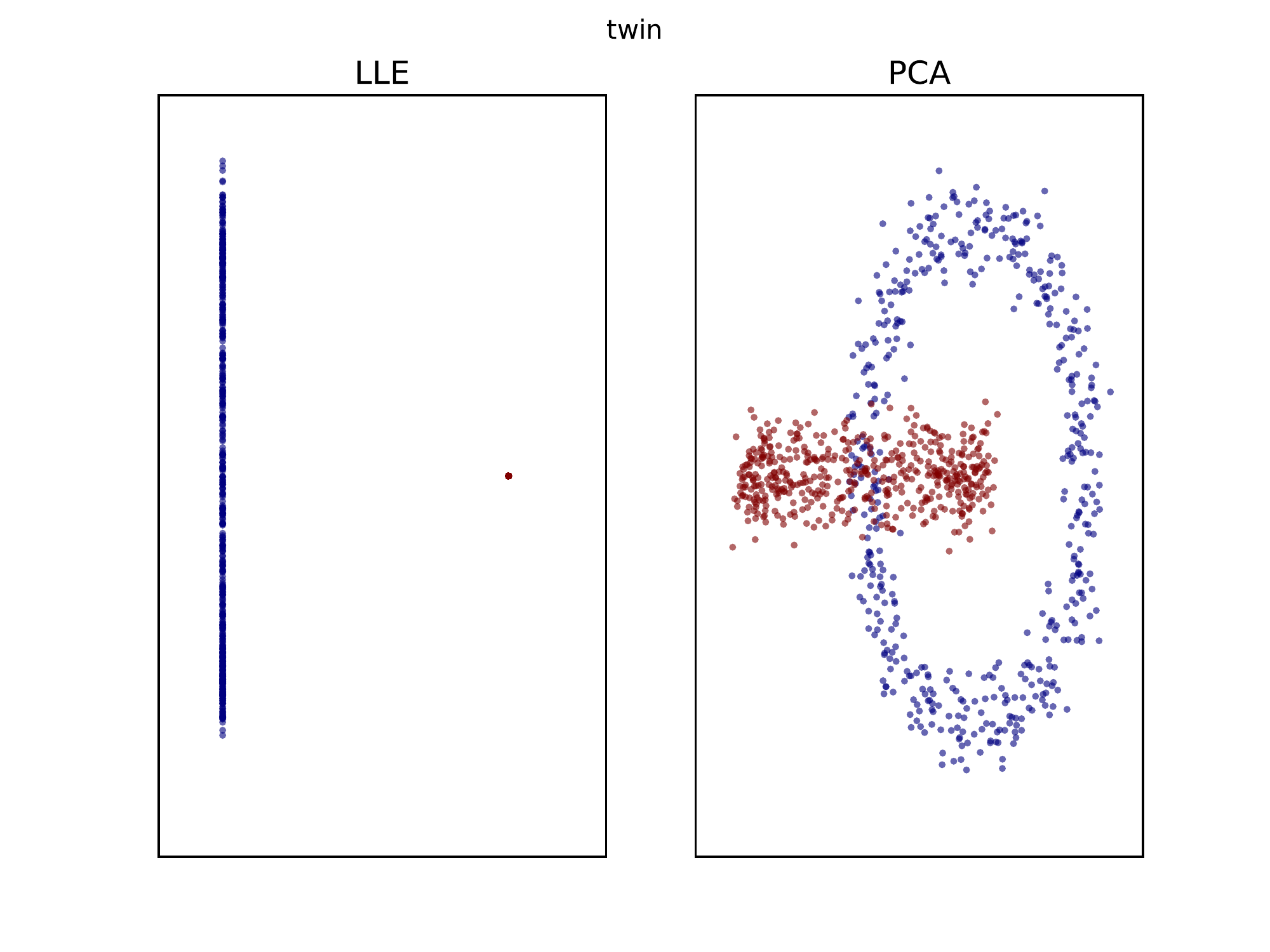}
    \caption{Visualize Twin-Ring with LLE and PCA.}
    \label{fig:twin}
\end{figure}

On one hand, semi-supervised learning takes advantage of the distribution of large amount of unlabeled samples to improve model performance by incorporating extra explicit self-supervision, or so-called regularization. On the other hand, it has been argued and examined that many high-dimension data in fact lie in a low-dimension manifold and we need nonlinear methods to capture the complex data distribution, where graph kicks in. 
For example, can we build a linear classifier to deal with the two-class problem (``Twin-Ring") as shown in \textbf{Figure~\ref{fig:twin3d}}?  Such challenges require careful problem formulation and efficient solutions.
Compare the visualization~\cite{LLE-SVM} given by Locally Linear Embedding (LLE) that involves graph and by the linear method Principle Component Analysis (PCA), as shown in \textbf{Figure~\ref{fig:twin}}. LLE uses graph to twist the space and makes the case easily separable as a clear contrast to PCA. 

Formally, graph-based learning involves a graph $\mathcal{G}=(\mathcal{V}, \mathcal{E}, A)$ 
that describes the pairwise relation of the samples (i.e., the nodes in $\mathcal{V}$), where $\mathcal{V}$ is the node set, $\mathcal{E}$ is the edge set and $A$ is the square adjacency matrix. Let $n$ be the number of nodes. Then $|\mathcal{V}| = n$ and $A\in \mathbb{R}^{n\times n}$. 

Another challenge in graph-involved learning is complexity. For a sparse graph we have $|\mathcal{E}|$ linear in $n$ while for a dense graph $|\mathcal{E}|$ quadratic in $n$. Hence, $A$ is filled with elements and when the graph is large any operations on $A$ becomes unmanageable. Therefore, it is critical to seek efficiency in effective graph-based methods in order to realize their practical value. 

Furthermore, on the one hand, introducing neural units to graph-based learning can significantly increase the model capacity and therefore promote the model performance especially given large amount of data. On the other hand, neural networks also induce computation burden during inference and the memory and storage overhead. We also need to address the efficiency issue in this aspect by making the neural networks lightweight and fast while preserving the performance.

\section{Dissertation Contributions and Overview}

\subsection{Formulate the Problem of Cross-domain Knowledge Transfer}
The first major contribution of this dissertation is that we formulated the problem of cross-domain knowledge transfer~\cite{LSI}. This part of work span off from the finding of unreliable and missing embedding features during the early exploratory investigation in machine learning aided financial market prediction~\cite{ml-fin}. We formulate and illustrate the problem of transferring knowledge from one domain to another as a semi-supervised learning problem, and demonstrate with the case of knowledge transfer between financial market historical trading information and word embedding in the semantic space, which seem completely unrelated but in fact can be well connected with the proposed solution~\cite{LSI}. The significance of this part of work is to enable us solve the missing and unreliable embedding issue when there is some side information. An example of the practical usage is to enhance the general pretrained emebedding for improved performance in domain-specific learning tasks. 

\subsection{Propose Solutions with Provable Properties}
We propose multiple solutions to solve the aforementioned problem and provide thorough analyses. In the initial work~\cite{LSI}, we propose a graph-based semi-supervised learning approach with provable spectral properties. In the followup work~\cite{yao2021}, we leverage the recent advance in graph neural network and organically combine it with fast graph construction techniques to better solve the embedding imputation problem.  

\subsection{Further Improve the Solutions on Efficacy and Efficiency}
Furthermore, we dive deep into the methodology itself and investigate how the solutions can be improved in general. Specifically, we inject a residual unit to achieve effective and efficient eigenvalue perturbation to the graph filter matrix in graph convolutional neural network~\cite{eiglearn}. We also study of problem of neural network pruning, and propose a solution~\cite{pruning} to minimize the computation requirement and storage overhead in its serving stage. 

%% file: chapter2.tex
\chapter{Graph-based Semi-supervised Learning} 

\section{From Missing Embedding to Cross-domain Knowledge Transfer}
\label{ssl definition}

Word embedding is the process of learning a compact real valued vector representation for word or phrase based on large corpus. This was traditionally done via matrix factorization based on word-word or word-document co-occurrence statistics, e.g., Latent Semantic Analysis~\cite{deerwester1990indexing}. In recent years, neural network based approaches with sampling~\cite{mikolov2013distributed}\cite{bojanowski2017enriching}\cite{pennington2014glove} have shown promising results given the large amount of textual data and computing power. However, the two approaches are essentially in the same spirit~\cite{levy2014neural}--the neural network approaches are in fact redefining the metric based on which the matrix is constructed. 

Word embedding techniques have been very useful in many natural language processing tasks but there are still questions not well answered in the literature. For example, it is difficult to generate reliable word embeddings if the corpus size is small or  indispensable words have relatively low frequencies~\cite{bojanowski2016enriching}. Such cases can happen in various domain-specific language tasks, e.g., chemistry, biology, and healthcare, where thousands of domain-specific terminologies exist. To be more specific, in some domain language tasks, there could be thousands or even millions of terminologies or abbreviations where embedding might be unreliable. A concrete example could be biochemical terms, healthcare terms or company names in financial market. However, usually we have some prior information for those terms. For instance, we know some physical properties for the biochemical terms and these are in fact the representations defined in a feature space, which is $\{x_i\}_{i=1}^{p+q}$. We also know some of the reliable embedding vectors in the semantic space, which is $\{y_i\}_{i=1}^p$. And we seek to learn the unknown embedding vectors in the semantic space, which is $\{y_j\}_{j=1}^q$. 

This challenge naturally drives us to find an effective way to leverage available useful information sources to enhance the embedding vectors of words and phrases in domain-specific NLP tasks. In this chapter, I will discuss how such cross-domain knowledge transfer problems can be solve by graph-based semi-supervised learning, by starting with the motivation of semi-supervised learning. 

In large-scale machine learning problems, one of the issues to be addressed has been the costly labeling process. It is desired to use a relatively small amount of labeled samples and take advantage of large amount of unlabeled samples to achieve comparable learning outcomes. An existing approach is semi-supervised learning with transductive inference~\cite{zhu2009introduction}, where the implicit supervision comes from a small amount of labeled samples and the explicit regularization comes from large amount of unlabeled samples. Formally, the embedding imputation can be formulated as a semi-supervised learning problem as
\begin{equation}
\label{ssl}
X=
\begin{bmatrix}
    X_p \\
    X_q \\
\end{bmatrix}
=
\begin {bmatrix}
    x_1^\top \\
    \vdots \\
    x_l^\top \\
    x_{p+1}^\top \\
    \vdots \\
    x_{p+q}^\top \\
\end {bmatrix}
\rightarrow
\begin {bmatrix}
    y_1^\top \\
    \vdots \\
    y_p^\top \\
    \hat{y}_{p+1}^\top \\
    \vdots \\
    \hat{y}_{p+q}^\top \\
\end {bmatrix}
=
\begin{bmatrix}
    Y_p \\
    Y_q \\
\end{bmatrix}
=
Y
\end{equation}
where $\{x_i\}_{i=1}^q$ are feature vectors for the unknown samples, 
$\{x_j\}_{j=1}^p$ are feature vectors for the known samples, 
and $\{y_j\}_{j=1}^p$ are the known embeddings. 
The goal is to infer $\{y_i\}_{i=1}^q$ for the unknown samples, as displayed in Equation~\ref{ssl}.

There are variations of semi-supervised learning~\cite{zhu2005semi}, among which graph-based methods have been demonstrated to be effective with clear spectral explanations~\cite{zhu2005graph}. I will illustrate how graph-based semi-supervised learning is able to describe the data point pairwise relation and therefore capture the 
nonlinearity using well established spectral graph theory. I will also introduce a new graph-based semi-supervised learning model that was previously published~\cite{LSI} where the embedding imputation problem is formulated and solved.

\section{Graph-based Approach}

\begin{figure}[t]
  \centering
  \graphicspath{{./figures/}}
  \includegraphics[width=.6\textwidth]{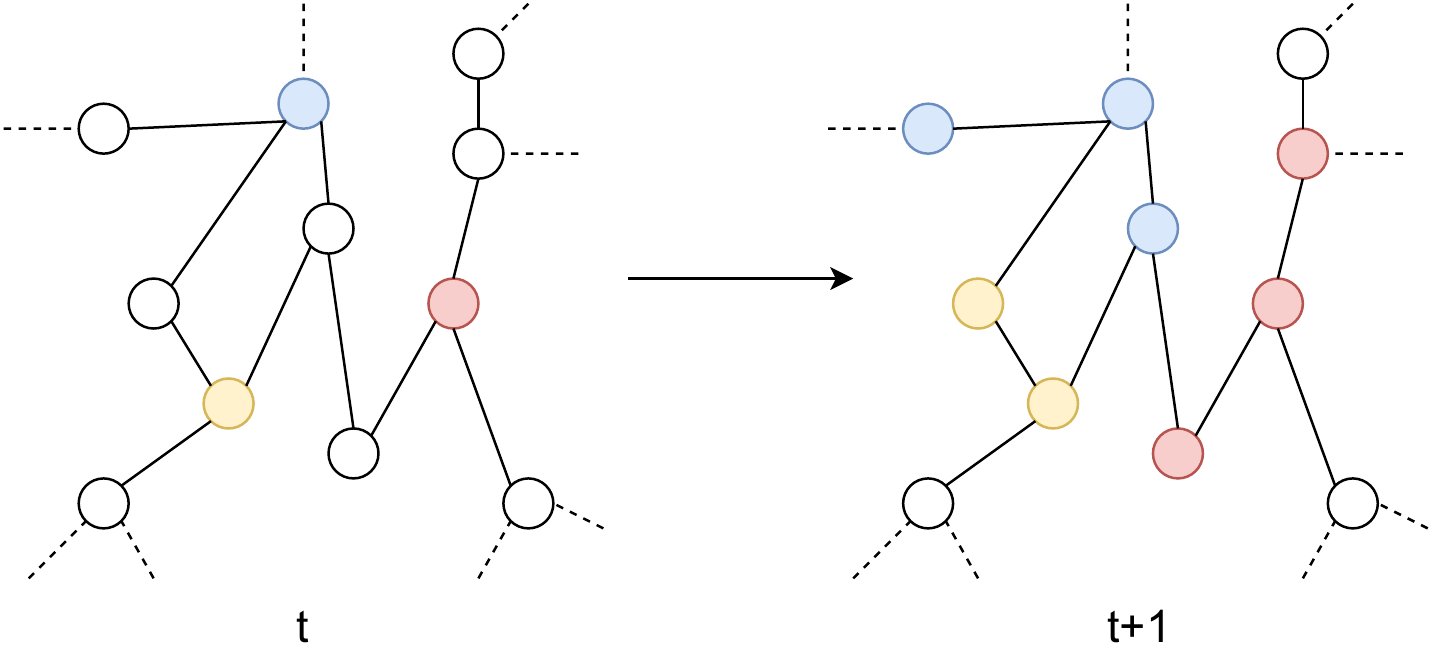}
  \caption{Label diffusion on graph.}
  \label{fig:LP}
\end{figure}

The assumption behind the graph-based approach is that the complex relation of data can be captured by a graph $\mathcal{G}=\{\mathcal{V}, \mathcal{E}\}$, where $\mathcal{V}$ is the node set and $\mathcal{E}$ is the edge set. We also use $A$ to denote the adjacency matrix that contains the edges. Existing graph-involved methods often assume that the representation of a point is some kind of weighted sum of its neighbors~\cite{roweis2000nonlinear}, in both the feature space $\mathbb{R}^n$ and the label space $\mathbb{R}^L$. This is similar to the idea of k-means clustering or Gaussian Mixture model with Expectation Maximization. The difference is that in the clustering process the labels are not predetermined while in semi-supervised learning with graph some samples have predetermined labels and remain unchanged (in some models even the sample with known labels can slightly change their labels). 

A high-level description of graph-based semi-supervised learning (sometimes also called graph transductive learning, label propagation~\cite{zhu2002learning} or label diffusion) usually includes three steps, the graph construction, the weight matrix (normalized weighted adjacency matrix) construction and the unknown label inference with random walk. The label diffusion depicted in \textbf{Figure}~\ref{fig:LP} can be explained as propagating information within the neighboring nodes on a graph, such as at $t$-step, there are three colored nodes on the graph and at $(t+1)$-step, the neighboring uncolored nodes intake information from the colored ones and transform to the same colors as of their neighboring colored nodes.

\section{A Typical Semi-supervised Learning with Graph}

\subsection{The Graph Construction}
\label{graph construction}

In graph-based semi-supervised learning, the first step is to construct the graph, i.e., to figure out the $E$ in $\mathcal{G}$, where $E$ is the edge set indicating the connections among samples. Given $\{x_i\}_{i=1}^p$ and $\{x_j\}_{j=1}^q$, we are able to construct a graph $\mathcal{G} = (V,E,A)$, where $V$ is the node set, $E$ is the edge set and $A$ is the adjacency matrix describing the weighted edges, based on some metric, $\phi(x_i,x_j)$, to describe the data point pairwise relation. 

An intuitive example is to apply the Gaussian kernel, $\phi(x_i,x_j) = exp(-\frac{\|x_i-x_j\|_2^2}{\sigma})$, on all node (sample) pairs $(x_i, x_j), \forall i\neq j$. Therefore, we are left with a pairwise affinity matrix $A$ where the larger the matrix element the closer the pair of nodes. Another example is the inverse of Euclidean distance, $\phi(x_i,x_j) = \frac{1}{\|x_i-x_j\|_2}$. 

Note that the aforementioned graph is complete, i.e., there are exactly $n\times n$ elements in $A$ where $n$ is the number of nodes. Many previous works~\cite{zhu2009introduction} have pointed out that a sparse graph, e.g., a $k$-nearest-neighbor graph or $\delta$-nearest-neighbor graph, shows better learning outcomes. The reasons are two-fold: (1) The data are distributed on a low-dimension manifold embedded in the original high-dimension space and the locality assumption is helpful~\cite{saul2003think} (2) A sparse graph can make the learning process significantly faster. Hence, throwing away some edges from the complete graph is a common practice. 

\subsection{The Weight Matrix Construction}

Once the adjacency matrix is constructed, we want to transfer the pairwise relation from the feature space to the label space, which is usually done by random walk~\cite{spielman2007spectral}. Define the degree matrix $D = diag(A\textbf{1})$ where $\textbf{1}$ is the one vector. The random walk matrix (sometimes also called probability matrix or transition matrix, in this draft it is also referred as weight matrix) is defined as $M = D^{-1}A$. There are also other ways to build the weight matrix and I will discuss shortly in the following section.

\subsection{Solving for the Unknown Labels with Random Walk}

For notational simplicity, let matrix $Y$ include the label vectors for both the known labels and the unknown labels, where $y_i$ is ordered consistently with $M$. The idea of random walk is $Y_t = MY_{t-1}$. In plain English, we are taking the weighted average of each sample's neighbors iteratively, until convergence. However, since the known labels should be fixed, some modifications on $M$ are necessary. 

Let us start with the convergence analysis of a random walk on a positive undirected graph $\mathcal{G} = (V,E)$. Recall the corresponding transition matrix for $\mathcal{G}$, $M = D^{-1}A$. Take a vector $\mathbf{z} \in \mathbb{R}^{p+q}$ (it could be a dimension of $Y$). The random walk process is depicted as $\mathbf{z}_t = M \mathbf{z}_{t-1}$. Recall the definition of eigenvalue $\lambda_i$ and eigenvector $\mathbf{v}_i$ of a real matrix $M$, $M\mathbf{v}_i = \lambda \mathbf{v}_i$. Expand $\mathbf{z}$ as a linear combination of the eigenvectors, $\mathbf{z} = \sum_{i=1}^{p+q}c_i\mathbf{v}_i$, where $c_i$ is the coefficient. One step of random walk is then $\mathbf{z}_t = M \sum_{i=1}^{p+q}c_i\mathbf{v}_i$. Therefore, we have 
\begin{equation}
    \lim_{t\rightarrow\infty} \mathbf{z}_t 
    = \lim_{t\rightarrow\infty} M^t \sum_{i=1}^{p+q}c_i\mathbf{v}_i
    = \lim_{t\rightarrow\infty} \sum_{i=1}^{p+q}c_i M^t\mathbf{v}_i 
    = \lim_{t\rightarrow\infty} \sum_{i=1}^{p+q}c_i \lambda_i^t\mathbf{v}_i
\end{equation}
So we know that the convergence of the random walk depends on the spectral radius of $M$, where $\rho(M) = max\{|\lambda_i|\}$. Recall we have the following theorem\cite{pillai2005perron}
\begin{theorem}\label{bound spectral radius}
For any nonnegative square matrices, the spectral radius is bounded by the minimum row sum and maximum row sum. 
\end{theorem}
In other words, $\forall B, B_{ij} \geq 0$
\begin{equation}
    min(\sum_{j}B_{ij}) \leq \rho(B) \leq max(\sum_{j}B_{ij}). 
\end{equation}
Since $min(\sum_{j}M_{ij}) = 1$ and $max(\sum_{j}M_{ij}) = 1$, we have $\rho(M) = 1$. Hence, $\lim_{t\rightarrow\infty} \mathbf{z}_t = \lim_{t\rightarrow\infty} \sum_{i=1}^{e}c_i \lambda_i^t\mathbf{v}_i$ where $e$ is the multiplicity of dominant eigenvalue. As for whether the random walk converges, it depends on whether there is any eigenvalue being -1 (the corresponding graph would be bipartite). And to mitigate the issue of eigenvalue being -1, we often adopt lazy random walk\cite{spielman2007spectral}, i.e., let $M_{Lazy} = 1/2(I+M)$ where $I$ is identity matrix. This ensures the eigenvalues of $M_{Lazy}$ is distributed between 0 and 1. The graph explanation is that we add self-loops to all the nodes to ensure the convergence of random walk.   

As for the weight matrix for semi-supervised learning, we want to make sure the known labels remain unchanged which can be done by modifying $M$. To ensure the modifications on $M$ serve our purpose, let us take a look at the meaning of the blocks within $M$. 
\begin{equation}
M = 
\left[
\begin{array}{c|c}
M_{pp} & M_{pq} \\
\hline
M_{qp} & M_{qq}
\end{array}
\right]
\rightarrow
\left[
\begin{array}{c|c}
I_p & 0 \\
\hline
M_{qp} & M_{qq}
\end{array}
\right]
\label{eqn:modified-weight-matrix}
\end{equation}
If we see the graph $\mathcal{G}$ as two separate components, where $\mathcal{G}_p$ contains all the labeled nodes and $\mathcal{G}_q$ contains all the unlabeled nodes, 
it is not hard to tell that $M_{pp}$ is the label diffusion within $\mathcal{G}_p$, $M_{qq}$ is the label diffusion within $\mathcal{G}_q$, $M_{pq}$ is the label diffusion from $\mathcal{G}_q$ to $\mathcal{G}_p$, and $M_{qp}$ is the label diffusion from $\mathcal{G}_p$ to $\mathcal{G}_q$. Because we do not want to change the known labels, we remove all incoming edges to $\mathcal{G}_p$, i.e., the edges within $\mathcal{G}_p$ and from $\mathcal{G}_q$ to $\mathcal{G}_p$, while retaining all self-loops within $\mathcal{G}_p$. Hence, $M_{pp}$ is replaced by an identity matrix and $M_{pq}$ is replaced by zero matrix. Next we need to show that the random walk with $M$ converges. Moreover, the stable distribution is irrelevant to the initialization of $Y$ (deterministic convergence). 

\begin{theorem}
If $M_{qq}$ is a convergent matrix, i.e., $\lim_{t \rightarrow \infty}M_{qq}^t = 0$, then random walk with $M$ guarantees deterministic convergence. 
\end{theorem}

\begin{proof}
Rewrite the walk as follows:
\begin{align}
  & \left[Y_p^{(t+1)}, Y_q^{(t+1)}\right] = \left[Y_P^{(t)}, M_{qp}Y_p^{(t)}+M_{qq}Y_q^{(t)}\right]. \\
 & \lim_{t \to \infty}Y_q^{(t)} = \lim_{t \to \infty}M_{qq}^tY_q^{(0)} + \left[\sum_{i=0}^{t-1}M_{qq}^{i-1} \right]M_{qp}Y_p.  
 \label{second-equation}
\end{align}

Given $\lim_{t \to \infty} M_{qq}^t=0$, the stable distribution is deterministic regardless of $Y_q^{(0)}$. 
\end{proof}

Recall that when $\mathcal{G}$ is a complete graph, $M$ is dense and $M_{qq}$ is a substochastic matrix where the row sum is strictly less than 1 (for example, when we use inverse Euclidean distance or Gaussian kernel and do not throw away any edges). In this case, $\rho(M_{qq}) < 1$ and $M_{qq}$ is a convergent matrix. Hence, such $M$ guarantees deterministic convergence. However, when $\mathcal{G}$ is of certain type and $M$ is of certain type, the convergence analysis could be a little bit more complex. We move the discussion to the next section.

\section{Latent Semantic Imputation}
\label{sec:LSI}

\subsection{The Algorithm and Properties}

\begin{figure}[t]
  \centering
  \graphicspath{{./figures/}}
  \includegraphics[width=.6\textwidth]{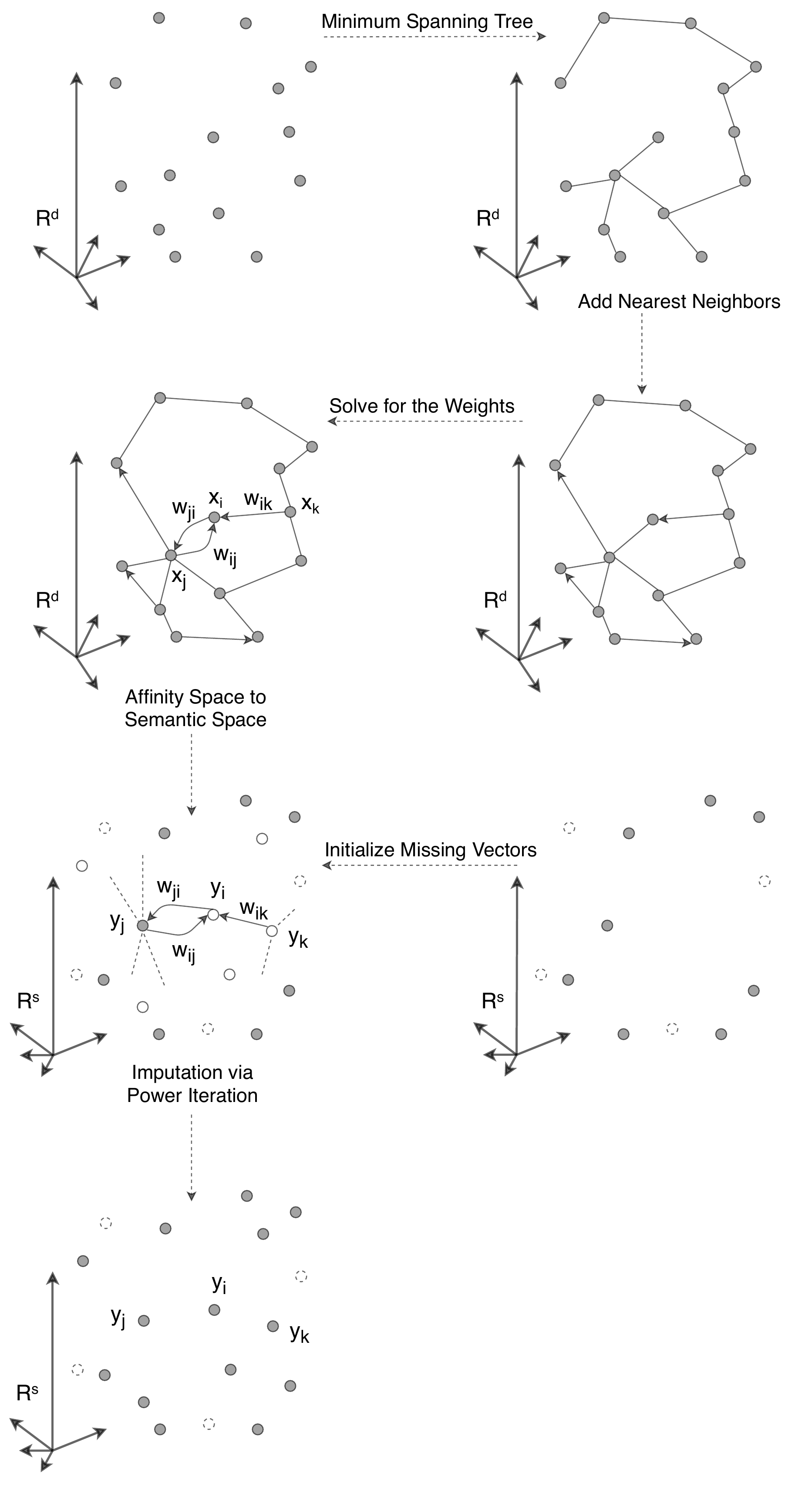}
  \caption{Schematic of Latent Semantic Imputation.}
  \label{fig:LSI}
\end{figure}

\textbf{Figure}~\ref{fig:LSI} depicts the process of Latent Semantic Imputation. We first build a graph by looking at the relative position between data points in the representation space. Then we apply non-negative least square to compute the weighted affinity matrix and apply power method to impute the missing vectors in the semantic space. Hence, the knowledge about the samples is transferred from one domain to another.

\begin{algorithm}[t]
    \SetKwFunction{Union}{Union}\SetKwFunction{FindCompress}{FindCompress}\SetKwInOut{Input}{Input}\SetKwInOut{Output}{Output}
 
    \Input{($X$, $\delta$) \tcp*{$\delta$:minimum degree}}
    \Output{$\mathcal{G}=(V,E)$ }
    \BlankLine
    $A = EuclideanDistance(X)$; \\
    $\mathcal{G}=(V,E)\gets Kruskal(A)$;   \\
    \For{$i\gets1$ \KwTo $|V|$}{ 
        $V_i\gets \{v_j\ |\ (v_j, v_i) \notin E \}$; \\ 
        \While{$deg^-(v_i)<\delta$ \tcp*{$deg^-$: in-degree}}{ 
            $v_j = argmin(v_j)\ d(v_i, v_j), v_j \in V_i$; \\
            $E \gets E \cup \{(v_j, v_i)\}$;\\
            $V_i \gets V_i \setminus \{(v_j, v_i)\}$;\\
        }
    }
\caption{MST-$k$-NN Graph}
\label{MST-kNN}
\end{algorithm}

As discussed in the previous section, many works introduced sparsity via kNN graph or $\epsilon$NN graph to semi-supervised learning, and found better learning outcomes. One major drawback of the aforementioned graphs is that they could be disconnected. Recall the initial motivation of semi-supervised learning with graph is that a data point's representation is some weighted average of its neighbors. If the graph is disconnected and within a connected component there is no labeled sample, the learning result for all the nodes in that connected component will be problematic -- there could be infinitely many optimal solutions. 

To mitigate the disconnection issue, we propose a Minimum-Spanning-Tree-k-Nearest-Neighbor graph (MST-kNN) in our work titled Latent Semantic Imputation shown in Algorithm~\ref{MST-kNN}. 
The idea is to maintain the locality with k-Nearest-Neighbor while ensuring the connectivity via a Minimum Spanning Tree. Note that the resulting graph is a directed graph and the associated adjacency matrix is asymmetric. 

Besides, inspired by nonlinear dimensionality reduction pioneer work, Locally Linear Embedding, we adopt least square to construct the weight matrix, or the walk matrix $M$ mentioned earlier. To ensure non-negativity we impose additional constraints in the objective shown below. 
\begin{equation}
\begin{aligned}
& \underset{\textit{M}}{\text{argmin}}
& & \sum_{i=1}^{n}\|\textbf{x}_i - \sum_{j=1}^{n}M_{ij}\textbf{x}_j \|^2 \\
& \text{s.t.}
& & (i,j) \in \mathcal{E} \\
& & & \sum_{j=1}^n M_{ij} = 1, i\neq j \\
& & &  M_{ij} \geq 0
\end{aligned}
\end{equation}

To solve this problem, note that it can be reduced to $n$ nonnegative least squares problems each of which tries to solve for an optimal weight vector with the same constraints, 
$
\underset{\textbf{m}_i}{\text{argmin}}
\|\textbf{x}_i - \sum_{j=1}^{n}M_{ij}\textbf{x}_j \|^2 ,
$
since solving for weight vector $\textbf{m}_i$ has no influence on solving for $\textbf{m}_j$, $\forall i \neq j$. 

In practice, during the matrix power process $Y_p$ is fixed, and only $Y_q$ needs to be updated during the iteration. Therefore, we set $M_p$ to identity, 
\begin{equation}
\left[
\begin{array}{c|c}
M_{pp} & M_{pq} \\
\hline
M_{qp} & M_{qq}
\end{array}
\right]
\rightarrow
\left[
\begin{array}{c|c}
I_p & 0 \\
\hline
M_{qp} & M_{qq}
\end{array}
\right]
\label{eqn:modified-weight-matrix}
\end{equation}
and then apply the power iteration to update the embedding matrix: $Y^{(t+1)} = MY^{(t)}$. The stopping criterion is the convergence of $Y_q$ when the $l_1$-norm changing rate of $Y_q$ between two iterations  falls under a predefined threshold $\eta$ or when the maximum number of iterations is reached. 
\begin{equation}
\frac{\left\|Y_q^{(t+1)} - Y_q^{(t)}\right\|_1}{\left\|Y_q^{(t)}\right\|_1} < \eta.
\end{equation}

Now we need to show that Latent Semantic Imputation guarantees deterministic convergence by showing $M_{qq}$ is a convergent matrix, using the same analysis paradigm. 

Due to the minimum spanning tree, we have the following lemma: 

\begin{lemma}\label{connectivity}
For every node in $\mathcal{G}_q$, there always exists a path from $\mathcal{G}_p$ to this node.
\end{lemma}

\begin{definition}[\textbf{Sink Node}]
Let $r_i = \sum_j(M_{qq})_{ij}$, the $i$-th row sum. A sink node in a substochastic matrix is one with $r_i < 1$.
\end{definition}
Given Lemma~\ref{connectivity}, we have the following corollary:
\begin{corollary}
For every node in $\mathcal{G}_q$, either it is a sink node or there exists a path from a sink node to it, or both.
\end{corollary} 

\begin{lemma}\label{substochastic}
For a substochastic matrix, for every non-sink node, if there exists a path from a sink node to this non-sink node, then the substochastic matrix is  convergent. 
\end{lemma}
\begin{proof}
To show $\lim_{t \to \infty} M_{qq}^t=0$, we need to show 
$$\forall i, \lim_{t \to \infty} r_i^{(t)} =  \lim_{t \to \infty} \sum_{j=1}^q \left(M_{qq}^{(t)}\right)_{ij} = 0,$$
or $\forall i,$ for a finite $t$
$$ \sum_{j=1}^q \left(M_{qq}^{(t)}\right)_{ij}  < 1.$$
 For every sink node $v_{k^*}$ in $\mathcal{G}_q$, we have $r_{k^*} < 1$. And $\forall t > 1,$
\begin{eqnarray*}
r_{k^*}^{(t)} &=& \sum_{k=1}^q\sum_{j=1}^q(M_{qq})_{k^*j} \left(M_{qq}^{(t-1)}\right)_{jk} \\
& = & \sum_{j=1}^q(M_{qq})_{k^*j}\sum_{k=1}^q\left(M_{qq}^{(t-1)}\right)_{jk} = \sum_{j=1}^q(M_{qq})_{k^*j}r_j^{(t-1)}. 
\end{eqnarray*}
Since we have $\forall i, \forall t>0, r_i^{(t)}\leq 1$, 
$$
r_{k^*}^{(t)} = \sum_{j=1}^q (M_{qq})_{k^*j}r_j^{(t-1)} \leq \sum_{j=1}^q (M_{qq})_{k^*j} = r_{k^*} < 1. 
$$
Thus, the convergence is apparently true for those sink nodes. Suppose the shortest path (with all positive edges) from a sink node $v_{k^*}$ to a non-sink node $v_i$ within $\mathcal{G}_q$ has $m$ steps. Then, we have 
$$
\left(M_{qq}^{(m)}\right)_{ik^*} > 0
$$
and 
$$
r_{k^*} < 1. 
$$
Hence, the following condition holds: 
$$r_i^{(m+1)} = \sum_{j=1}^q \left(M_{qq}^{(m)}\right)_{ij} r_j < \sum_{j=1}^q \left(M_{qq}^{(m)}\right)_{ij} = r_i^{(m)} \leq 1, i\neq k^*.$$
Because our graph is always finite, the convergence also holds for the non-sink nodes. 
\end{proof}

Combining Lemma~\ref{connectivity} and \ref{substochastic}, we conclude that  $M_{qq}$ is a convergent matrix  under our algorithm settings. Hence, LSI guarantees a deterministic convergence. This property does not always hold for a $k$-NN graph and it is also the reason why we start with a minimum spanning tree in the graph construction. 

Another benefit of using a sparse graph here is it reduces the complexity in the least squares step. A remark is that the final walk result is a series of linear combinations of the dominant eigenvectors of the walk matrix where the combination coefficients are determined by the graph construction. And this has its close relation with spectral clustering~\cite{von2007tutorial}. 

\subsection{Further Improvement}
The complexity of the presented MST-kNN graph is $O(Elog(v))$ due to the minimum spanning tree construction. This does not scale well when the original complete graph grows large. There are existing works that are able to construct approximate kNN graphs at sub-quadratic complexity~\cite{chen2009fast}\cite{zhang2013fast} without explicitly computing the Euclidean distance matrix. Therefore, a possible improvement would be to construct a kNN graph approximately based on the original sample representation vectors, and then fix the connectivity by adding a small amount of edges. 

Another line of further improvement lies in spectral sparsification~\cite{spielman2011spectral}\cite{koutis2015faster}. We are looking at constructing a graph sparsifier without computing the whole distance matrix, while still preserving some properties of the graph. Such graph can be applied to the semi-supervised learning process. An application study with the complexity issue mitigated is introduced in the next Chapter.

\section{Empirical Study}
\label{application}

The graph-based semi-supervised learning has been widely applied in learning labels. Nevertheless, its usage is not limited to this. In our previous work, we used Latent Semantic Imputation to fuse prior knowledge to enhance domain word embedding, which is the first attempt of its kind. 

\begin{figure}[h]
  \centering
  \graphicspath{{./figures/}}
  \includegraphics[width=\textwidth]{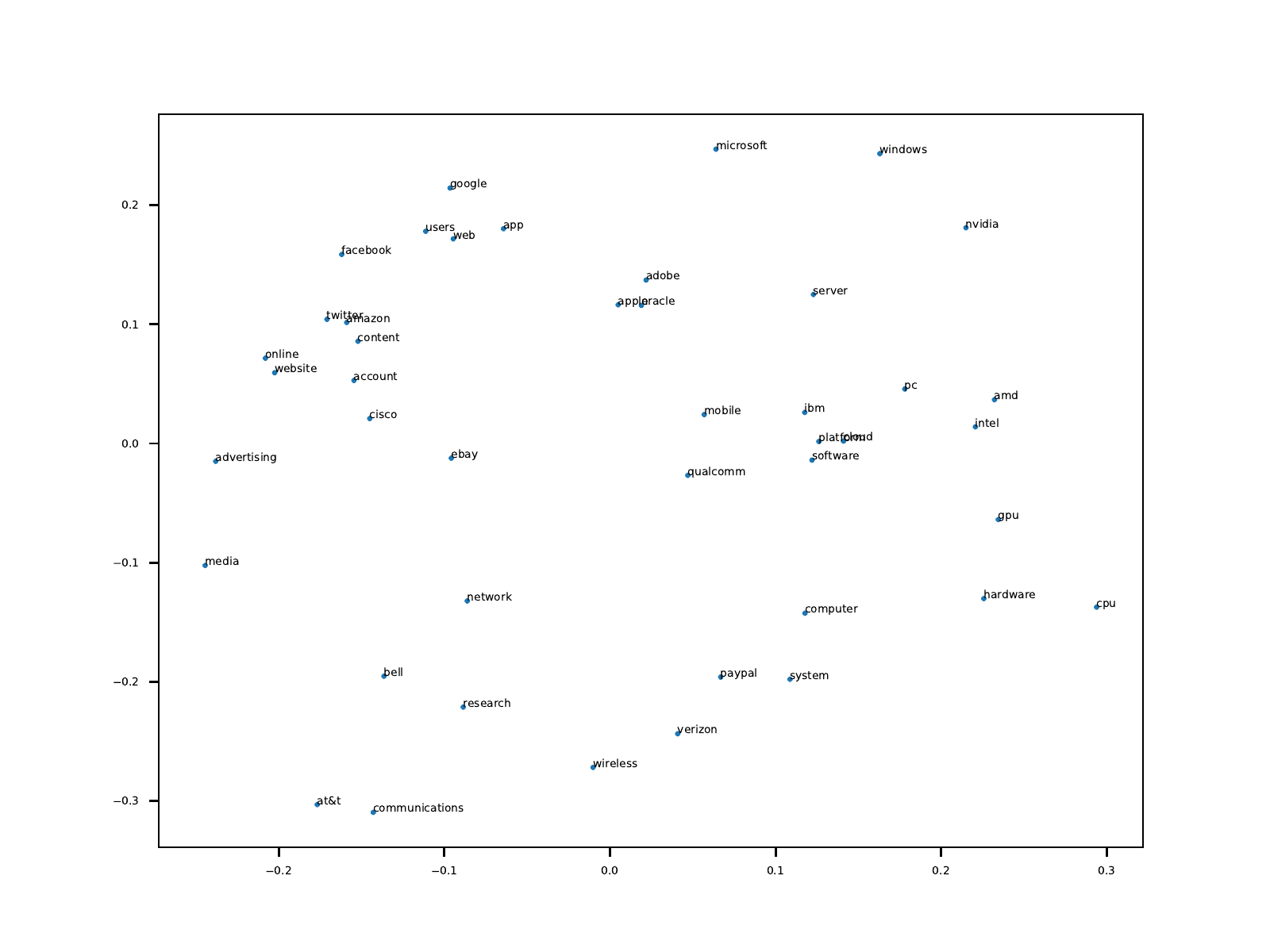}
  \caption{Visualizing company terms using word2vec on a plane.}
  \label{fig:visual_company}
\end{figure}

We focus on the financial terms that appeared in financial text and try to enhance the embedding for these terms. In the exploratory study, we crawled Wiki pages about S\&P500 companies and obtained the associated term embedding vectors using word2vec, and visualized the terms on a plane as shown in \textbf{Figure}~\ref{fig:visual_company}. The visualization indicates that the embedding captures the latent meaning of the terms. For example, the hardware company and terms such as nvidia, amd, intel gpu ,cpu and hardware are close to each other, while facebook, google, user, web and app form another clique because they are internet companies and highly rely on user network. However, by checking the statistics of these term we found they have relatively high frequency. For the low frequency words, their embedding vectors do not follow the pattern of latent meaning in terms of language. Hence, it's critical to improve the embedding quality for the low frequency words in order to use them in downstream language tasks.

\begin{figure}[h]
\graphicspath{{./figures/}}
\begin{subfigure}{.5\linewidth}
\centering
\includegraphics[width=\linewidth]{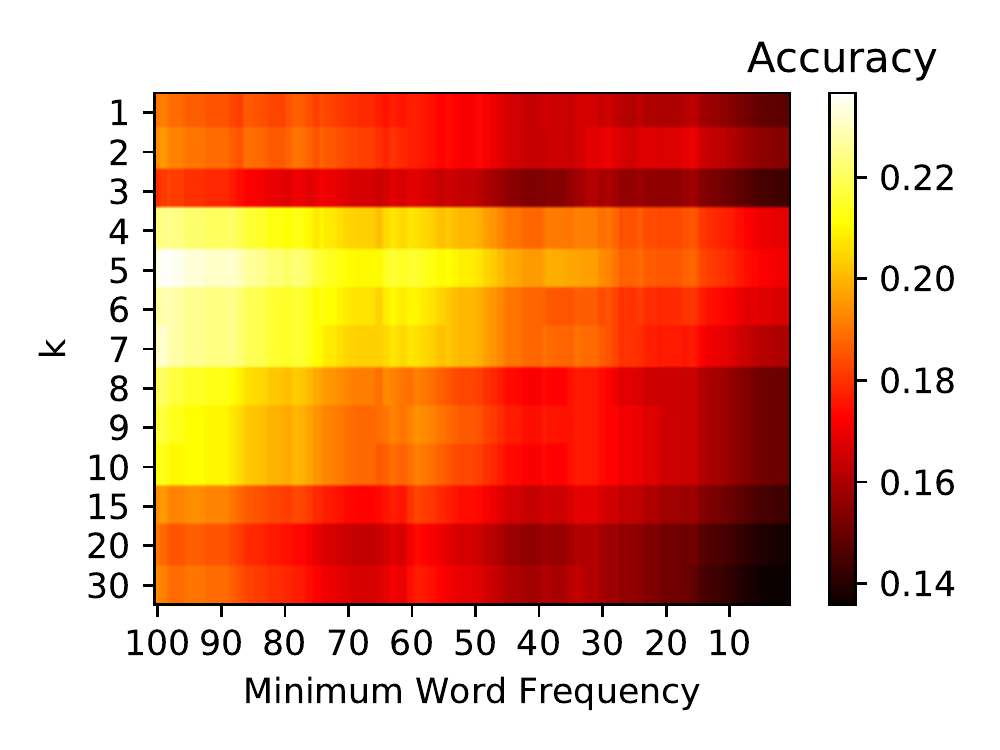}
\label{fig:heat_before}
\end{subfigure}
\begin{subfigure}{.5\linewidth}
\centering
\includegraphics[width=\linewidth]{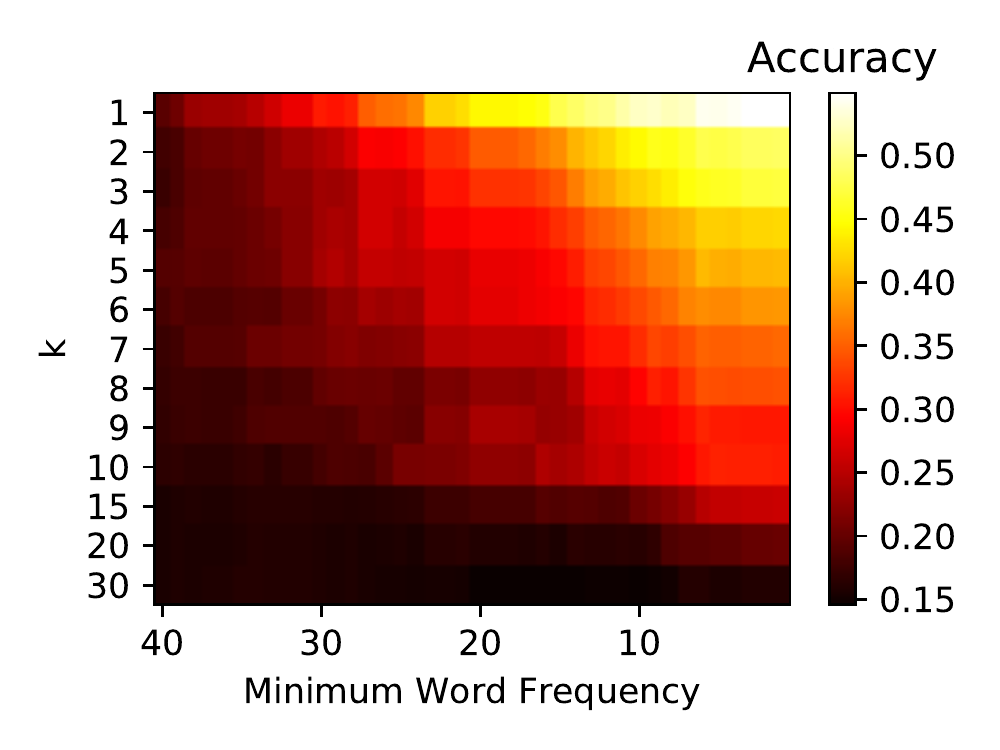}
\label{fig:heat_after}
\end{subfigure}
\caption{$k$-NN accuracy on word embedding vectors. (a) Self-trained embedding on Wiki corpus (b) Self-trained embedding on the same Wiki corpus combined with the side information via LSI.}
\label{fig:heat}
\end{figure}

We first demonstrate the embedding quality is negatively related to the word frequency via experiment and how fusing prior knowledge can improve the embedding quality, as shown in \textbf{Figure~\ref{fig:heat}}. \textbf{Table~\ref{tab:freq}} tracks the detailed classification accuracy  on different embeddings. For example, ``self" means self-trained embedding, ``self(hf)" means self-trained embeddings for high-frequency words only, ``+aff" means incorporating side information using LSI. 

\begin{table}
  \centering
  \caption{$k$-NN Accuracy (\%) on Embedding Vectors}
  \scalebox{0.9}{
  \begin{tabular}{c | c c c c c c c c }
    \toprule
    \diagbox[]{$E$}{$k$} & 2 & 5 & 8 & 10 & 15 & 20 & 30\\
    \midrule
    self & 0.154 & 0.170 & 0.150 & 0.150 & 0.144 & 0.138 & 0.135 \\
    self(hf) & 0.180 & 0.190 & 0.172 & 0.167 & 0.157 & 0.157 & 0.157 \\
    \textbf{self(hf)+aff} & 0.556 & 0.472 & 0.396 & 0.359 & 0.302 & 0.261 & 0.187 \\
    Google & 0.220 & 0.297 & 0.271 & 0.305 & 0.280 & 0.280 & 0.186\\
    \textbf{Google+aff} & 0.838 & 0.803 & 0.784 & 0.768 & 0.725 & 0.678 & 0.626 \\
    Glove & 0.417 & 0.466 & 0.490 & 0.500 & 0.500 & 0.505 & 0.451 \\
    \textbf{Glove+aff} & 0.832 & 0.766 & 0.690 & 0.653 & 0.606 & 0.542 & 0.405 \\
    fast & 0.443 & 0.496 & 0.527 & 0.500 & 0.511 & 0.470 & 0.447 \\
    \textbf{fast+aff} & 0.811 & 0.749 & 0.713 & 0.684 & 0.641 & 0.608 & 0.595\\
    \bottomrule
  \end{tabular}}
  \label{tab:freq}
\end{table}

To verify that LSI is robust to its hyper-parameter $\delta$ and stopping criterion $\eta$, we did multiple investigations.  When we set $\eta = 1e^{-2}$ and let $\delta$ vary, we observed that LSI is relatively robust to a varying $\delta$ under the constraint that $\delta$ is not too large in which case the manifold assumption is significantly violated, or too small, which causes one or two neighbors to dominate.  When we set the minimum degree of the graph $\delta=8$ and let $\eta$ vary, we also had the same observation that the LSI is robust to the stopping criterion $\eta$.
\begin{figure}
\begin{subfigure}{.5\linewidth}
\centering
\includegraphics[width=\linewidth]{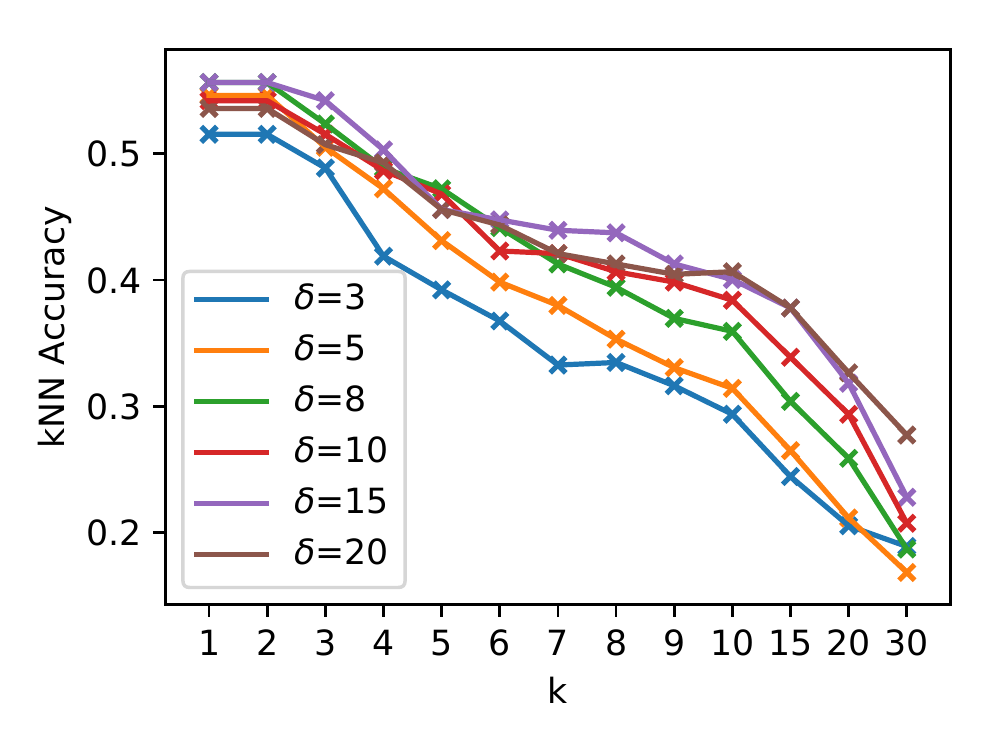}
\caption{Sensitivity of $\delta$}
\label{fig:delta sens}
\end{subfigure}%
\begin{subfigure}{.5\linewidth}
\centering
\includegraphics[width=\linewidth]{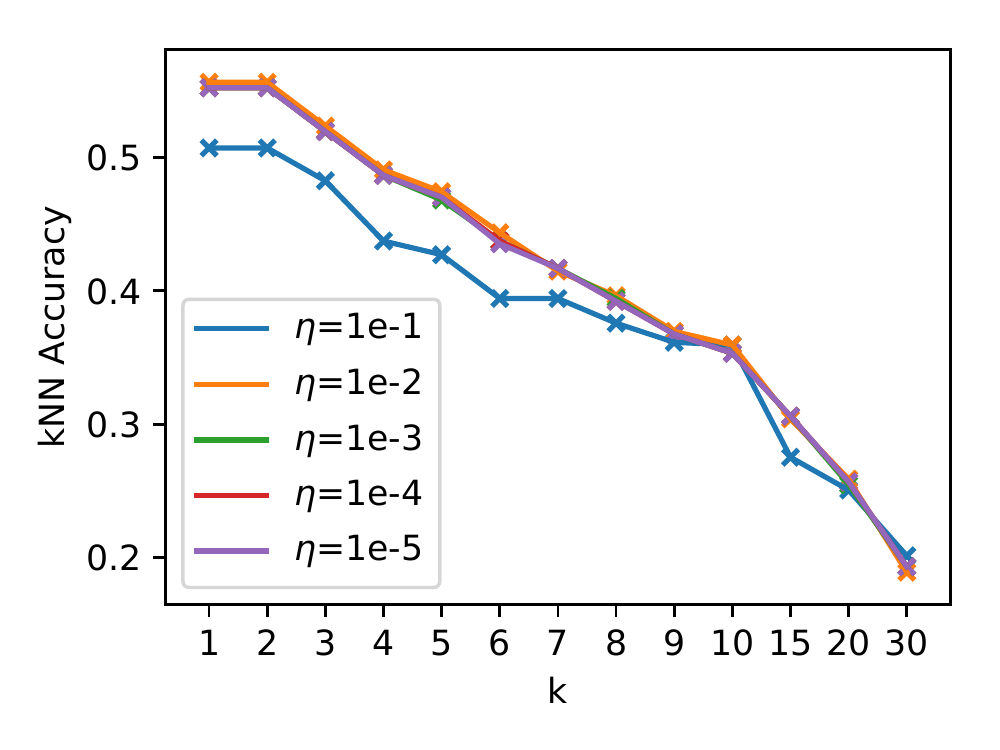}
\caption{Sensitivity of $\eta$}
\label{fig:eta sens}
\end{subfigure}
\caption{Sensitivity tests on node degree and stopping criterion for LSI.}
\label{fig:sensitivity test}
\end{figure}

\begin{figure}[h]
  \centering
  \graphicspath{{./figures/}}
  \includegraphics[width=\textwidth]{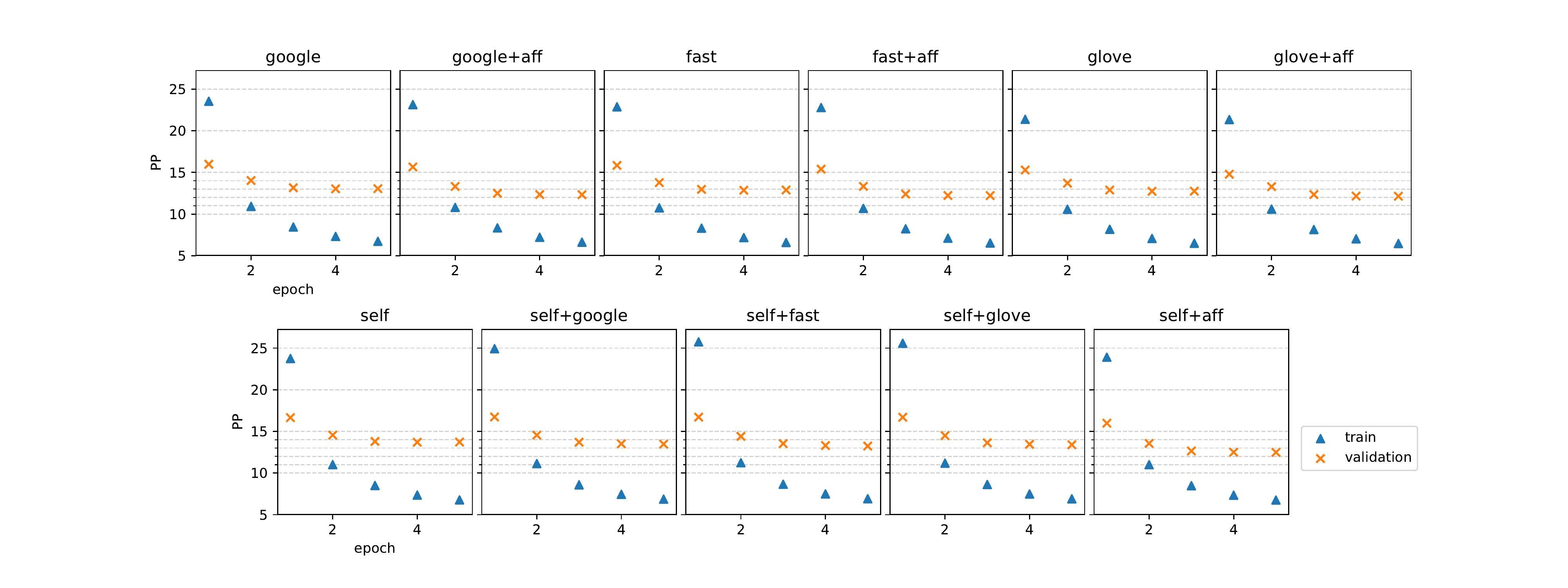}
  \caption{Training and validation perplexity with different embeddings.}
  \label{fig:train_pp}
\end{figure}

And then we use the enhanced embedding in an LSTM-based language model, and show that the enhanced embedding leads to be better language modeling~\cite{bengio2003neural} performance measured by perplexity. \textbf{Table~\ref{tab:10run}} shows the detailed testing perplexity for different embeddings over 10 runs. \textbf{Figure~\ref{fig:train_pp}} displays the training and validation perplexity during the LSTM learning process for different embeddings. 
For more details please see \cite{LSI}. 

\begin{table}
\centering
  \caption{Language Model Test Perplexity for 10 Runs}
  \scalebox{0.7}{
  \begin{tabular}{c | c c c c c c c c c c c}
    \toprule
    \diagbox[]{$E$}{$round$} & 1 & 2 & 3 & 4 & 5 & 6 & 7 & 8 & 9 & 10\\
    \midrule
    \textbf{google+aff} &11.617&11.676&11.676&11.57&11.615&11.557&11.6&11.743&11.731&11.677 \\
    google &12.315&12.527&12.37&12.473&12.391&12.363&12.434&12.448&12.535&12.454 \\
    \textbf{self+aff} &11.838&11.956&11.914&11.858&11.822&11.92&11.912&11.922&11.892&11.8 \\
    self &12.934&13.153&13.157&13.038&13.168&13.031&13.112&12.987&13.124&13.228 \\
    \textbf{self+google} &12.742&12.849&12.828&12.705&12.76&12.676&12.759&12.697&12.762&12.645 \\
    \textbf{self+glove} &12.639&12.626&12.59&12.675&12.752&12.62&12.63&12.635&12.681&12.61 \\
    \textbf{self+fast} &12.456&12.518&12.516&12.35&12.441&12.546&12.502&12.418&12.485&12.536 \\
    fast &12.168&12.105&12.162&12.303&12.258&12.193&12.211&12.252&12.152&12.35 \\
    \textbf{fast+aff} &11.626&11.629&11.617&11.689&11.662&11.675&11.667&11.552&11.639&11.622 \\
    \textbf{glove+aff} &11.524&11.564&11.446&11.469&11.468&11.618&11.526&11.439&11.491&11.55 \\
    glove &12.265&12.136&12.146&12.26&12.309&12.188&12.155&12.25&12.308&12.167 \\
    \bottomrule
  \end{tabular}}
  \label{tab:10run}
\end{table}


%% file: chapter3.tex
\chapter{Empowering Graph SSL with Neural Units}

Although baring clear spectral explanation, the power of conventional semi-supervised learning methods with graph is limited due to their model capacity, since there are no learnable parameters. This leads to the fact that they are better at problems that involve regular-size datasets, but are mostly inferior in dealing with large data which is the trend in nowadays machine learning practice. 

Graph Neural Networks (GNNs)~\cite{scarselli2008graph}\cite{defferrard2016convolutional}\cite{kipf2016semi} have achieved promising performance improvement in solving various problems in the past few years, by combining the idea of information diffusion on graph and the rich capacity of neural networks. In this chapter, I will discuss the relevant GNN models, as well as how they can be tailored with graph algorithms and applied to the problems mentioned in the previous chapter effectively and efficiently. 

\section{Graph Convolutional Neural Networks}

Graph Convolutional Neural Networks(GCN)~\cite{kipf2016semi}\cite{defferrard2016convolutional}\cite{bruna2013spectral} are defined on graphs as an extension of convolutional neural networks(CNN)~\cite{lecun1995convolutional} based on graph signal processing studies~\cite{shuman2013emerging}. CNN was designed based on discrete signal processing techniques to capture the spatial information. Concretely, in vision tasks, CNN learns better feature map of the given images by applying filter on the reception field which contains pixels close to each other. By adopting the same idea, GCN applies graph filter on locally neighboring nodes on a graph and generates better node representations and thereby the overall learning outcomes. 

Since GCN is an organic combination of graph learning and neural network, we need to first illustrate the idea of neural network. Given $\{x_i\}$ and $\{y_i\}$ as the feature vectors and the corresponding ground truths as the training data, we want to learn a mapping $y=\phi(x)$ such that it can be used for inference when a new sample $x_j$ is given. Take one dense layer as an example. The mapping $\phi(x)$ is parameterized as $\sigma(Wx)$, where $W$ is a trainable weight matrix containing the free parameters and $\sigma$ is the nonlinear activation function such as softmax. The training process is done by first defining a loss function $\mathcal{L}(y, \phi(x))$ such that we can measure the distance between the prediction and ground truth, and then taking the gradient $\frac{\partial \mathcal{L}}{\partial W}$ and performing gradient descent such that $\mathcal{L}$ is minimized. Now, we want to inject such a mechanism into graph semi-supervised learning. 

Recall the semi-supervised learning problem defined in Section~\ref{ssl definition}, where 
given $\{x_q\}$ and $\{x_p\}$ which denote the sets of feature vectors of unlabeled and labeled samples respectively, 
and $\{y_p\}$ which denotes the set of labels associated with $\{x_p\}$,
we want to infer the labels $\{y_q\}$ for $\{x_q\}$.
Graph-based learning involves a graph $\mathcal{G}=(\mathcal{V}, \mathcal{E}, A)$ 
that describes the pairwise relation of the samples (i.e., the nodes in $\mathcal{V}$). 
GCN belongs to the family of graph-based learning, which takes advantage of both $\mathcal{G}$
and the rich capacity of the neural network to improve model performance. 

A typical graph convolution layer is
$$
Z^{(l+1)} = \sigma(SZ^{(l)}W^{(l)})
$$
where $l$ denotes a certain layer, $W^{(l)} \in \mathbb{R}^{d^{(l)}\times d^{(l+1)}}$ is the trainable weight matrix in that layer, 
$Z^{(l)} \in \mathbb{R}^{n\times d^{(l)}}$ is the feature map matrix, $\sigma$ denotes some nonlinear activation function,
and $S \in \mathbb{R}^{n\times n}$ is the graph filter matrix constructed 
based on $A$. As an example, $S = \tilde{D}^{-1/2} \tilde{A} \tilde{D}^{-1/2}$ 
where $\tilde{A} = A + I$ and $\tilde{D} = diag(\tilde{A}\textbf{1})$, 
i.e., the so-called symmetrically normalized adjacency matrix with self-loop added.
The spectral convolution on the input signal $Z^{(l)}$ can be shown as: 
$$
SZ^{(l)} = \Sigma_{i=1}^n \textbf{v}_{i} \lambda_{i} \textbf{v}_{i}^{T} Z^{(l)}
$$
where $\lambda_i$ and $\textbf{v}_i$ are the eigenvalue and eigenvector of $S$, 
and $\{\textbf{v}_i\}$ also forms the bases of the graph signal. 

\begin{figure}[ht]
    \centering
    \includegraphics[width=\linewidth]{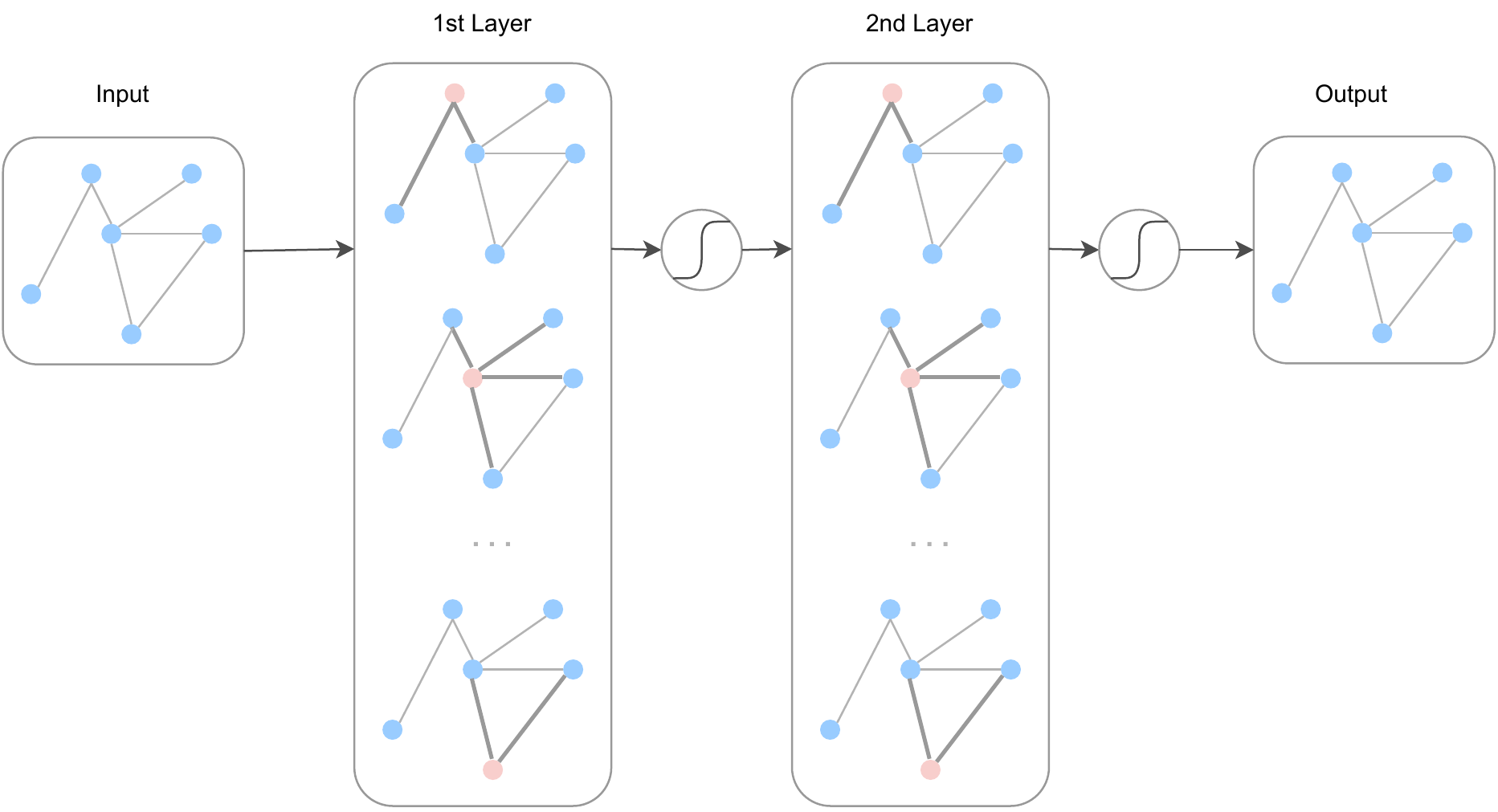}
    \caption{A two-layer GCN schema.}
    \label{fig:GCN}
\end{figure}

Usually there are multiple such layers stacked together in a model to capture the 
complex mapping from the input domain to the output domain, such as the two-layer GCN shown in \textbf{Figure~\ref{fig:GCN}}. However, studies have found that
deeper GCN models tend to suffer from overfitting. To propagate the node information 
further away, one can inject the idea of personalized page rank into GCN.

\section{Deeper Propagation with Personalized PageRank}

The Personalized PageRank Neural Prediction (PPNP) was proposed by~\cite{klicpera2018predict}
to further improve conventional GCNs. The motivation of PPNP is to take advantage of node information
further away by using personalized PageRank~\cite{page1999pagerank} with teleport. 
Derived from the stationary distribution, the PPNP layer can be described by
\begin{equation}
   Z = softmax(\alpha (I - (1-\alpha)S)^{-1}H) 
   \label{ppnp}
\end{equation}
where $\alpha$ is the hyper-parameter that controls the strength of teleport. 

Using Personalized PageRank in graph neural network is to mitigate the propagation limit on graph. In a regular GCN, there are usually two to three layers and hence the information propagation on graph is limited to two to three hops. Introducing more layers in the architecture has two main drawbacks, being (1) graph neural network losing track of the local information on graph (2) more layers lead to more trainable parameters and hence overfitting. Personalized PageRank mitigates such problems by introducing a teleport step which allows the walk back to the starting point. This ensures that propagation on graph incorporate multi-hop neighbors while preserving the local information of the starting point without making the graph neural neural network deep. 

Note that the Formula~\ref{ppnp} involves matrix inversion which is usually super-quadratic and does not scale to large graphs. Moreover, the resultant graph filter matrix is dense which makes the graph convolution evaluation quadratic. 
Hence, the authors of~\cite{klicpera2018predict} proposed the approximate solution (APPNP) 
based on the idea of power iteration which can be written as
\begin{equation}
\begin{split}
   & Z^{(0)} = H = f_{\theta}(X),\\
   & Z^{(l+1)} = (1-\alpha) SZ^{(l)} + \alpha H,\\
   & Z^{(L)} = softmax((1-\alpha) SZ^{(L-1)} + \alpha H) 
\end{split}
\label{eq:appnp}
\end{equation}
where $f_{\theta}$ is some densely connected neural network layers parameterized by trainable weight matrices. 
The benefit of adopting this approach
is to allow multiple propagation steps in a shallow GCN and avoid overfitting,
which allows the information propagate further on the graph 
without increasing the asymptotic complexity of that in a regular GCN~\cite{kipf2016semi}. I should point out that Equation~\ref{eq:appnp} is an explicit Neumann series and it resembles the shifted invert Laplacian filter.

\section{Anchor Sampling for Graph Construction}

To apply the graph-based methods for semi-supervised learning, we need the adjacency matrix $A$ that
describes the pairwise relation among the samples. However, the topology information $A$ is not 
always available. In such cases, the common practice is to use the features $\{x_i\}=\{x_q\} \cap \{x_p\}$ to construct $A$. 

A detailed illustration and some examples of graph construction are introduced in Section~\ref{graph construction}. The typical graph construction takes quadratic time and does not scale to large graphs. One can utilize anchor sampling process as a good approximte solution with linear complexity. There are multiple fast solutions to certain graph construction, including approximate kNN graph~\cite{zhang2013fast}\cite{chen2009fast} and graph sparsifier construction~\cite{alman2020algorithms}. We adopt a sampling based solution for approximate kNN graph~\cite{liu2010large} and find it easy to implement, effective and robust when combined with graph neural network. 

\subsection{A Customized Approximate Solution}

In a recent work~\cite{yao2021}, we adopt the idea of anchor-graph~\cite{liu2010large} to mitigate the scalability issue. The key idea is to reduce to constant the size of the node set from which one selects $k$ neighbors. 
To construct the anchor-$k$NN graph, a constant number $m$ anchor nodes are randomly sampled from the set of the $p$ nodes with known embeddings and their feature vectors are stacked into a matrix $X_m \in \mathbb{R}^{m\times d}$. 
Then we loop through the entire node set and compute the Euclidean distance between each node and the selected $m$ anchor nodes using their feature vectors. 
Based on the Euclidean matrix, $\delta$ nearest neighbors are selected for each node. 
And if node $i$ is a neighbour of node $j$, a directed edge is added from node $i$ to node $j$ and a directed edge from node $j$ to node $i$, i.e., set $A_{ij} = 1$ and $A_{ji} = 1$ in the adjacency matrix $A$.

As said, the essence of anchor-graph is to choose a constant number $m$ anchor nodes such that the nearest neighbor search space for one node is reduced from $n$ to $m$. Therefore, when one uses a partition function as in quick search to choose $\delta$ nearest neighbors for a node, 
the time complexity is reduced from $O(n)$ to $O(m)$, and the overall complexity of $k$NN search for $n$ nodes is reduced from $O(n^2)$ to $O(mn)$. Also, the overall complexity of distance computation as a precursor for $k$NN search is reduced from $O(dn^2)$ to $O(dmn)$. 
The graph construction is depicted in Algorithm~\ref{anchor-graph},
where $X_m$ is the matrix stacked from the feature vectors of the $m$ anchor nodes, and choice() is the random sampling process with uniform distribution. 
$choice(V, \delta)$ means randomly choose $\delta$ nodes from the node set $V$. 
Note that in Algorithm~\ref{anchor-graph} we explicitly construct a mutual $k$NN graph, 
which is equivalent to the common practice in GCN of converting directed graphs to undirected graphs. 

\begin{algorithm}[t]
    \SetKwFunction{Union}{Union}\SetKwFunction{FindCompress}{FindCompress}\SetKwInOut{Input}{Input}\SetKwInOut{Output}{Output}
 
    \Input{($X$, $\{p\}$, $\delta$, $m$) \tcp*{$X$:feature matrix; $\{p\}$:index set for samples with embedding;
                 $\delta$:desired node degree; $m$:number of anchors}}
    \Output{$\mathcal{G}=(V,E,A)$ }
    \BlankLine
    $X_{m} = choice(X_{p}, m)$ \\
    $C = EuclideanDistance(X, X_{m})$ where $C \in \mathbb{R}^{n\times m}$ \\
    $\mathcal{G}=(V,E,A)$, where $E=\phi$ and $A=\textbf{0}$  \\
    \For{$i\ \textbf{in}\ {n}$}{
        $\gamma = NN\_index(C_i, \delta)$ using a partition function; \\
        $A_{i\{\gamma\}} = \textbf{1}$; \\
        $E \gets E \cup \{(v_i, v_{\{\gamma\}})\}$; \\
        $A_{\{\gamma\}i} = \textbf{1}$; \\
        $E \gets E \cup \{(v_{\{\gamma\}}, v_{i})\}$; \\
    }
\caption{Anchor-$k$NN Graph}
\label{anchor-graph}
\end{algorithm}

One might question that Algorithm~\ref{anchor-graph} does not necessarily produce a connected graph, which seems contradictory to the argument of ensuring connectivity using Minimum Spanning Tree in Section~\ref{sec:LSI}. Indeed connectivity is not guaranteed in Algorithm~\ref{anchor-graph}. However, since the anchors are always chosen from the nodes with known embeddings, we are left with the situation where each connected component in the graph involves at least one node with known embedding. And this achieves a similar effect to that in Algorithm~\ref{MST-kNN}. 

The anchor-$k$NN graph constructed from Algorithm~\ref{anchor-graph} is an unweighted graph, which means all edges are treated equally. To better represent the pair-wise relation and control the strength of information propagation on the graph, we seek to assign weights based on the edges and the feature vectors associated with the nodes, via the same approach as described in Section~\ref{sec:LSI} by treating each node as a nonnegative linear combination of its neighbors. 
The optimization is depicted as follows, where $x_i$ denoted the original feature vector of node (word) $i$,
$w_{ij}$ is a scalar weight and $\delta$ denotes node degree. 
\begin{equation}
\begin{aligned}
& \underset{\textit{W}}{\text{argmin}}
& & \sum_{i=1}^{n}\left\|\textbf{x}_i - \sum_{j=1}^{\delta}w_{ij}\textbf{x}_j \right\|^2 \\
& \text{s.t.}
& & \sum_{j=1}^{\delta} w_{ij} = 1, i\neq j \\
& & &  w_{ij} \geq 0
\end{aligned}
\label{eqn:linear-combine}
\end{equation}

\subsection{Complexity Analyses}
As described in the previous section, the overall approach that combines APPNP (or GCN) and anchor-graph is scalable in solving the embedding imputation problem. 
In the graph construction, sampling anchors takes $O(1)$ time, the distance computation between $n$ nodes and $m$ anchors takes $O(dmn)$ time where $d$ denotes the dimension of the original feature vector, selecting $k$ $(\delta)$ nearest neighbor from $m$ anchor nodes for one node takes $O(m)$ time using a partition function and hence the overall $k$NN search for $n$ nodes takes $O(mn)$ time. In the weight matrix construction,  
solving one least square problem takes $O(dk^3)$ time
and solving $n$ NNLS problems takes $O(dnk^3)$. Imposing the non-negativity constraints doesn't alter the asymptotic complexity since this is done by projection. 
Note that if we do not construct a sparse graph using anchor-$k$NN, the complexity of
solving a least square problem is cubic in $n$. In graph neural network training, 
given a sparse graph, the complexity of evaluation is also linear in $n$. Hence, the overall complexity of approach is linear with respect to $n$, given $m$, $d$ and $k$($\delta$) are constant. 

\subsection{Convergence Analyses}
\begin{lemma}\label{convergent}
The weighted graph constructed based on Algorithm~\ref{anchor-graph} and Equation~\ref{eqn:linear-combine}
guarantees deterministic convergence in an infinite step matrix power. 
\end{lemma}

\begin{proof}
It suffices to conclude the proof if we show that all leading eigenvalues of $W$ is 1. 

In other words, 
$$
\lim_{t \rightarrow \infty} W^t\textbf{x} 
= \lim_{t \rightarrow \infty} W^t \Sigma_i^n c_i\textbf{v}_i
= \lim_{t \rightarrow \infty} \Sigma_i^n c_i W^t \textbf{v}_i
= \lim_{t \rightarrow \infty} \Sigma_i^{n} c_i \lambda_i^t \textbf{v}_i
$$
, where $\textbf{x}$ is the given representation vector which is 
a linear combination of eigenvectors $\textbf{v}_i$ of $W$ determined by coefficients $c_i$
and $\lambda_i$ are the eigenvaues of $W$. 
Suppose there are $r$ leading eigenvalues of $W$ in magnitude and $\lambda_r = 1, \forall r$, 
then we have
$$
\lim_{t \rightarrow \infty} W^t\textbf{x} 
= \Sigma_r c_r \textbf{v}_r. 
$$

Note that the graph generated by Algorithm~\ref{anchor-graph} is not necessarily connected. 
Hence, we need to show that the leading eigenvalues of all its connected components are 1. 
By permutation $W$ is block-diagonal, i.e.,
$$
W = 
\begin{pmatrix}
    \ddots\\
    &{B_{i}}\\
    &&\ddots\\
\end{pmatrix}. 
$$
The matrix power process for a block is 
$$
x_{i}^{(t+1)} = B_{i}x_{i}^{(t)}.
$$
Due to Equation~\ref{eqn:linear-combine} all the row summations of $B_i$ are 1, $\forall i$. 
This leads to the fact that the spectral radius of $B_i$ is 1, $\forall i$~\cite{seneta2006non}. 
Recall Perron-Frobenius theorem. Given a nonnegative matrix that is associated with 
a strongly connected graph whose spectral radius is 1, all of its leading eigenvalues
are 1~\cite{seneta2006non}. 
Algorithm~\ref{anchor-graph} explicitly constructs a mutual $k$NN graph which 
guarantees all the connected components are strongly connected graphs, which concludes the proof. 
\end{proof}

Note that there is a subtle difference between the construction of $W$ in~\cite{LSI}
and in this work. In the previous work, the diagonal block matrix that corresponds to 
the known embeddings are fixed to be identity in order to achieve deterministic convergence. 
However, in this work we allow information propagation within the set of known embeddings
in order to take advantage of GNN models. It is important to ensure there is no leading eigenvalue
being -1 which corresponds to a bipartite graph and makes the walk process nonconvergent.

\begin{remark}
It is guaranteed that the unlabeled nodes incorporate information from labeled nodes following Lemma~\ref{convergent}. 
\end{remark}

Because in every connected component, there is at least one labeled node due to the fact 
that the anchors are sampled from labeled node set. Otherwise, it is possible that there exists 
a connected component that doesn't include any known embedding, in which case 
the information propagation based on the graph topology is much less meaningful.

\section{Empirical Study}
The application problem setting is the same to that in Section~\ref{application}. Essentially we formulate the 
embedding imputation problem as a semi-supervised learning problem and apply the approach discussed to better solve it. 
The imputation is evaluated by both classification and regression tasks, via accuracy and mean squared error(MSE). 

The baseline models are Latent Semantic Imputation(LSI), simple multi-layer perceptron (MLP) and graph convolutional neural networks(GCN).
LSI falls within the category of conventioanl graph semi-supervised learning and does not contain any neural units. Hence, its model 
capacity is significantly smaller than the others and its performance inferiority should indicate the usefulness of the larger capacity introduced by the neural units. MLP doesn't involve any graph topology and therefore there is no information propagation or in other words, graph regularization. By setting it as a control group we can testify the role of graph. GCN allows only one-step propagation within a single layer and information diffusion is limited, while APPNP is able to propagate the information multiple steps without making the neural architecture deep. To make a fair comparison, the neural network settings in MLP, GCN and APPNP are configured in exact same way, i.e., same number of layers and same number of hidden units in each layer etc. 

The experiments are done on three general pretrained embeddings, namely word2vec~\cite{mikolov2013distributed}, GloVe~\cite{mikolov2013distributed} and fastText~\cite{bojanowski2017enriching}. The side information is the financial corpus retried from wikipedia. For a detailed data description and configuration, please refer~\cite{LSI}\cite{yao2021}. 

\textbf{Table~\ref{tab:class_large}} shows the classification accuracy comparison among different methods. 
Row-wise we have difference base embeddings and the imputation method applied. ``base" means no imputation is applied. Column-wise are different $k$ values used in a $k$NN classifier to test the sensitivity. Mean and standard deviation are reported to quantify the randomness in experiments. Bold faces are leaders in performance. Three observations can ba made: 
(1) applying any imputation method leads to significant improvement on the overall embedding quality as indicated by the comparison against base in classification accuracy 
(2) GCN often outperforms LSI and MLP 
(3) APPNP which combines power method and the rich capacity of neural net is the leader.

\begin{table}[htbp]
\centering
  \caption{Classification Accuracy (\%) on Embedding Vectors from the Small Dataset}
  \scalebox{0.7}{
  \setlength\tabcolsep{3pt}
  \centerline{
  \begin{tabular}{c | c c c c c c c }
    \toprule
    \diagbox[]{E}{$k$} & 2 & 5 & 8 & 10 & 15 & 20 & 30\\
    \midrule
    (w2v) \\
    base & 22.03 & 29.66 & 27.12 & 30.51 & 27.97 & 27.97 & 18.64 \\
    LSI & $78.01\pm0.06$ & $\bm{79.43}\pm0.15$ & $76.41\pm0.14$ & $75.52\pm0.18$ & $72.79\pm0.14$ & $68.54\pm0.35$ & $65.63\pm0.40$\\
    MLP & $73.37\pm1.51$ & $73.35\pm0.87$ & $71.50\pm0.74$ & $70.64\pm1.33$ & $68.93\pm0.98$ & $67.54\pm1.27$ & $65.89\pm0.70$\\
    GCN & $78.25\pm0.83$ & $77.84\pm1.16$ & $75.65\pm0.93$ & $74.31\pm1.18$ & $71.50\pm1.62$ & $69.43\pm1.45$ & $66.34\pm1.32$\\
    \textbf{APPNP} & $\bm{78.60}\pm0.66$ & $78.62\pm1.04$ & $\bm{76.96}\pm0.57$ & $\bm{76.53}\pm0.65$ & $\bm{73.86}\pm1.12$ & $\bm{71.83}\pm1.09$ & $\bm{69.84}\pm0.89$ \\
    \hline 
    (GloVe) \\
    base & 41.75 & 46.60 & 49.03 & 50.00 & 50.00 & 50.49 & 45.15 \\
    LSI & $77.64\pm0.11$ & $75.69\pm0.38$ & $70.68\pm0.20$ & $69.32\pm0.21$ & $65.79\pm0.28$ & $62.81\pm0.32$ & $56.04\pm0.51$ \\
    MLP & $78.83\pm0.74$ & $80.41\pm0.91$ & $79.47\pm0.91$ & $78.30\pm0.76$ & $77.06\pm0.76$ & $75.91\pm1.11$ & $72.98\pm1.04$ \\
    GCN & $\bm{80.33}\pm0.80$ & $82.51\pm0.35$ & $\bm{81.27}\pm0.51$ & $80.35\pm0.52$ & $80.00\pm0.48$ & $77.97\pm0.71$ & $74.97\pm0.79$  \\
    \textbf{APPNP} & $80.27\pm0.52$ & $\bm{82.83}\pm0.69$ & $\bm{81.27}\pm0.76$ & $\bm{81.31}\pm0.81$ & $\bm{80.37}\pm0.52$ & $\bm{78.83}\pm0.48$ & $\bm{75.44}\pm0.79$ \\
    \hline 
    (FastText) \\
    base & 44.27 & 49.62 & 52.67 & 50.00 & 51.15 & 46.95 & 44.66  \\
    LSI & $76.08\pm0.17$ & $74.15\pm0.30$ & $69.32\pm0.37$ & $67.84\pm0.52$ & $64.70\pm0.78$ & $62.14\pm0.29$ & $57.33\pm0.40$ \\
    MLP & $76.02\pm0.48$ & $76.10\pm1.14$ & $76.30\pm0.72$ & $75.30\pm0.54$ & $73.47\pm0.84$ & $71.33\pm0.99$ & $65.17\pm1.08$ \\
    GCN & $77.84\pm0.66$ & $76.98\pm0.65$ & $77.19\pm0.68$ & $76.37\pm0.93$ & $73.80\pm0.93$ & $71.29\pm0.87$ & $66.78\pm0.94$ \\
    \textbf{APPNP} & $\bm{78.30}\pm0.73$ & $\bm{77.70}\pm0.78$ & $\bm{78.23}\pm0.80$ & $\bm{77.10}\pm0.51$ & $\bm{75.38}\pm0.58$ & $\bm{71.95}\pm0.82$ & $\bm{67.13}\pm0.81$\\
    \bottomrule
  \end{tabular}
  }}
  \label{tab:class_small}
\end{table}

\begin{table}[t]
\centering 
  \caption{Classification Accuracy (\%) on Embedding Vectors from the Large Dataset.}
  \scalebox{0.7}{
  \setlength\tabcolsep{3pt}
  \centerline{
  \begin{tabular}{c | c c c c c c c }
    \toprule
    \diagbox[]{E}{$k$} & 2 & 5 & 8 & 10 & 15 & 20 & 30\\
    \midrule
    (w2v) \\
    base & 26.04 & 26.56 & 31.25 & 29.17 & 28.13 & 27.60 & 28.13\\ 
    LSI & $43.36\pm0.12$ & $47.00\pm0.12$ & $47.52\pm0.10$ & $47.60\pm0.09$ & $47.86\pm0.10$ & $48.19\pm0.11$ & $47.88\pm0.06$\\
    MLP & $41.29\pm0.36$ & $42.88\pm0.29$ & $43.03\pm0.24$ & $42.69\pm0.25$ & $42.37\pm0.22$ & $41.78\pm0.46$ & $40.55\pm0.42$\\
    GCN & $46.89\pm0.43$ & $48.59\pm0.49$ & $48.78\pm0.33$ & $48.81\pm0.33$ & $48.11\pm0.28$ & $47.57\pm0.26$ & $46.63\pm0.36$\\
    \textbf{APPNP} & $\bm{49.94}\pm0.16$ & $\bm{52.55}\pm0.45$ & $\bm{52.87}\pm0.18$ & $\bm{52.72}\pm0.18$ & $\bm{52.50}\pm0.24$ & $\bm{51.78}\pm0.30$ & $\bm{51.25}\pm0.32$ \\
    \hline 
    (GloVe) \\
    base & 31.58 & 32.83 & 34.09 & 34.09 & 34.59 & 34.59 & 33.83 \\
    LSI & $44.40\pm0.07$ & $47.38\pm0.08$ & $47.52\pm0.11$ & $48.14\pm0.08$ & $48.41\pm0.10$ & $48.17\pm0.17$ & $47.46\pm0.09$ \\
    MLP & $44.94\pm0.39$ & $47.66\pm0.27$ & $48.01\pm0.20$ & $47.95\pm0.29$ & $47.54\pm0.25$ & $46.93\pm0.18$ & $45.93\pm0.35$ \\
    GCN & $50.06\pm0.35$ & $52.87\pm0.32$ & $53.32\pm0.32$ & $53.18\pm0.26$ & $52.68\pm0.37$ & $52.09\pm0.32$ & $51.41\pm0.28$\\
    \textbf{APPNP} & $\bm{51.62}\pm0.41$ & $\bm{54.64}\pm0.17$ & $\bm{54.80}\pm0.28$ & $\bm{54.91}\pm0.33$ & $\bm{54.77}\pm0.29$ & $\bm{54.57}\pm0.33$ & $\bm{54.24}\pm0.25$ \\
    \hline 
    (FastText) \\
    base & 34.16 & 40.84 & 41.88 & 42.28 & 42.02 & 40.58 & 37.96\\
    LSI & $45.66\pm0.09$ & $47.56\pm0.06$ & $47.70\pm0.12$ & $47.86\pm0.14$ & $48.93\pm0.11$ & $48.19\pm0.08$ & $47.59\pm0.24$ \\
    MLP & $46.95\pm0.47$ & $49.22\pm0.44$ & $49.88\pm0.44$ & $49.73\pm0.48$ & $49.52\pm0.49$ & $48.84\pm0.50$ & $48.02\pm0.46$ \\
    GCN & $50.36\pm0.33$ & $53.09\pm0.38$ & $53.62\pm0.45$ & $53.65\pm0.36$ & $53.36\pm0.28$ & $53.06\pm0.25$ & $51.94\pm0.28$ \\
    \textbf{APPNP} & $\bm{52.02}\pm0.38$ & $\bm{54.99}\pm0.35$ & $\bm{55.62}\pm0.22$ & $\bm{55.60}\pm0.25$ & $\bm{55.52}\pm0.27$ & $\bm{55.40}\pm0.25$ & $\bm{55.12}\pm0.14$\\
    \bottomrule
  \end{tabular}}}
  \label{tab:class_large}
\end{table}

\textbf{Table~\ref{tab:regression}} shows the regression performance on the three general pretrained embeddings. 
The word set is decided by wordnet~\cite{miller1995wordnet}. 
$\delta$ denotes node degree in the anchor-$k$NN graph. ``w2v $\leftarrow$ GloVe" means using GloVe pre-trained embedding as the side information to do embedding imputation on word2vec pretrained embedding. The result is consistent with that in the classification task, 
i.e., all imputation methods make the overall embedding quality much better while 
APPNP always yields the best performance in terms of MSE. 

\begin{figure}[ht]
    \centering
    \includegraphics[width=\linewidth]{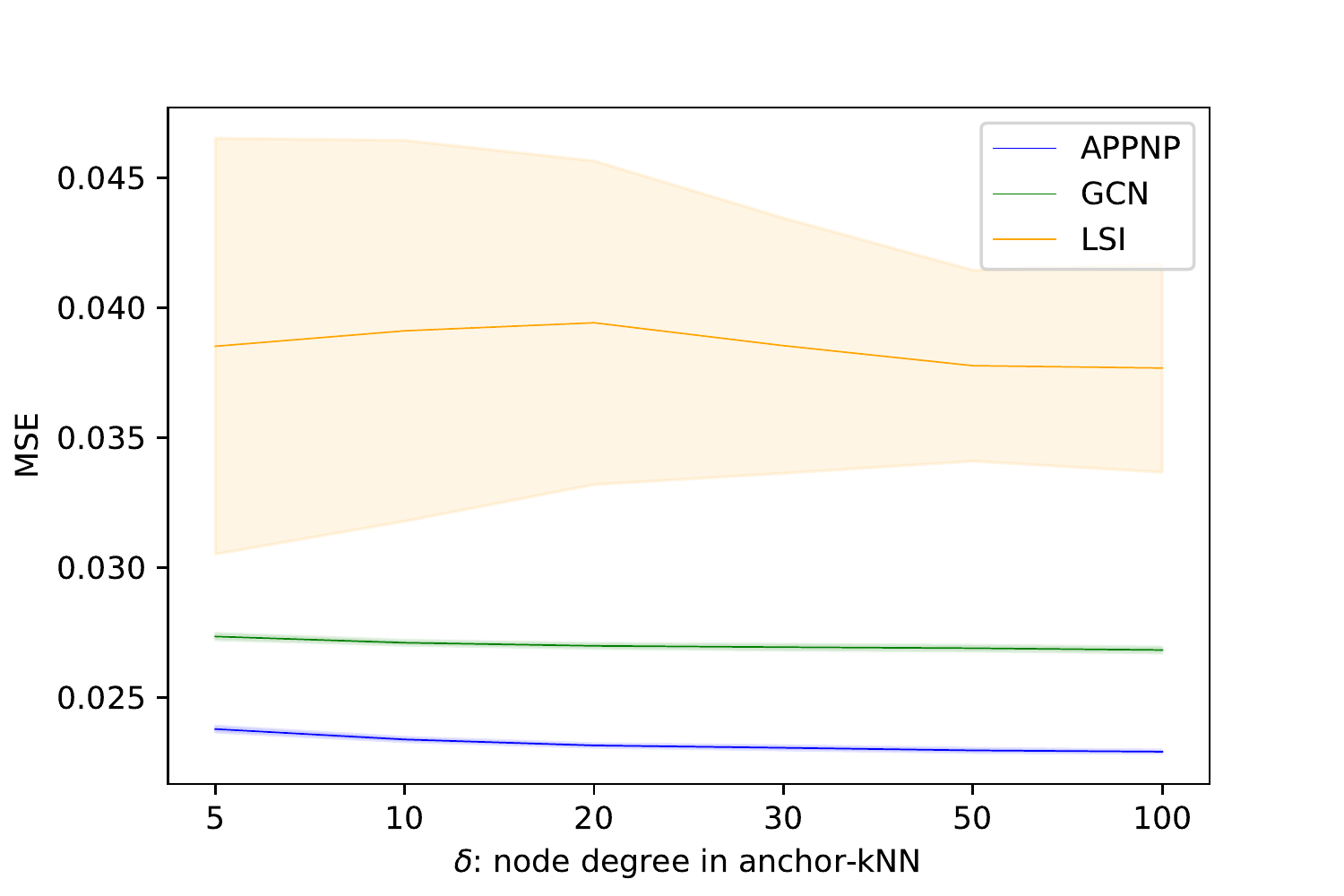}
    \caption{Sensitivity study on node degree in anchor-kNN.}
    \label{fig:sens_delta}
\end{figure}

Besides, \textbf{Table~\ref{tab:sol_comp}} shows the comparison between the exact solution of $k$NN and the approximate solution described in Algorithm~\ref{anchor-graph}. 
The observation is that the approximate solution based on the anchor-$k$NN graph yields performance as good as the exact solution based on the $k$NN graph. 
It is especially interesting to notice that when using anchor-$k$NN graph in APPNP, 
the final performance is almost identically good. A conjecture is that the neural units
to some extent compensate the graph construction randomness and variation. There is also a sensitivity study result on node degree for graph construction shown in \textbf{Figure}~\ref{fig:sens_delta}. The main message from the figure is that when combined with graph neural network, the approximate graph construction is quite robust against node degree.

\begin{table}[t]
\centering 
  \caption{MSE on Embedding Vectors for Regression Task}
  \setlength\tabcolsep{3pt}
  \renewcommand{\arraystretch}{0.8}
  \scalebox{0.8}{
  \begin{tabular}{c | c c c }
    \toprule
    \diagbox[]{E}{$\delta$} & 5 & 10 & 20 \\
    \midrule
    w2v $\leftarrow$ GloVe \\
    LSI & $0.0393\pm0.0004$ & $0.0364\pm0.0002$ & $0.0349\pm0.0002$ \\
    MLP & $0.0312\pm0.0002$ & $0.0312\pm0.0002$ & $0.0312\pm0.0002$ \\
    GCN & $0.0274\pm0.0001$ & $0.0271\pm0.0001$ & $0.0270\pm0.0001$ \\
    \textbf{APPNP} & $\bm{0.0238}\pm0.0001$ & $\bm{0.0234}\pm0.0001$ & $\bm{0.0232}\pm0.0001$ \\
    \hline 
    w2v $\leftarrow$ FastText \\
    LSI & $0.0390\pm0.0006$ & $0.0365\pm0.0004$ & $0.0352\pm0.0003$ \\
    MLP & $0.0328\pm0.0001$ & $0.0328\pm0.0001$ & $0.0328\pm0.0001$ \\
    GCN & $0.0286\pm0.0001$ & $0.0285\pm0.0001$ & $0.0284\pm0.0001$ \\
    \textbf{APPNP} & $\bm{0.0253}\pm0.0001$ & $\bm{0.0250}\pm0.0001$ & $\bm{0.0249}\pm0.0001$ \\
    \hline 
    GloVe $\leftarrow$ w2v \\
    LSI & $0.1900\pm0.0026$ & $0.1782\pm0.0022$ & $0.1710\pm0.0018$ \\
    MLP & $0.1151\pm0.0008$ & $0.1151\pm0.0008$ & $0.1151\pm0.0008$ \\
    GCN & $0.1065\pm0.0005$ & $0.1061\pm0.0004$ & $0.1058\pm0.0004$ \\
    \textbf{APPNP} & $\bm{0.1023}\pm0.0002$ & $\bm{0.1019}\pm0.0002$ & $\bm{0.1015}\pm0.0002$ \\
    \hline 
    GloVe $\leftarrow$ FastText \\
    LSI & $0.1860\pm0.0028$ & $0.1743\pm0.0019$ & $0.1681\pm0.0014$ \\
    MLP & $0.1224\pm0.0005$ & $0.1224\pm0.0005$ & $0.1224\pm0.0005$ \\
    GCN & $0.1118\pm0.0003$ & $0.1113\pm0.0003$ & $0.1110\pm0.0003$ \\
    \textbf{APPNP} & $\bm{0.1073}\pm0.0003$ & $\bm{0.1063}\pm0.0003$ & $\bm{0.1057}\pm0.0002$ \\
    \hline 
    FastText $\leftarrow$ w2v \\
    LSI & $0.0736\pm0.0015$ & $0.0692\pm0.0012$ & $0.0664\pm0.0008$ \\
    MLP & $0.0540\pm0.0004$ & $0.0540\pm0.0004$ & $0.0540\pm0.0004$ \\
    GCN & $0.0480\pm0.0002$ & $0.0478\pm0.0002$ & $0.0476\pm0.0002$ \\
    \textbf{APPNP} & $\bm{0.0436}\pm0.0001$ & $\bm{0.0432}\pm0.0001$ & $\bm{0.0430}\pm0.0001$ \\
    \hline 
    FastText $\leftarrow$ GloVe \\
    LSI & $0.0727\pm0.0012$ & $0.0675\pm0.0007$ & $0.0647\pm0.0006$ \\
    MLP & $0.0546\pm0.0003$ & $0.0546\pm0.0003$ & $0.0546\pm0.0003$ \\
    GCN & $0.0484\pm0.0003$ & $0.0480\pm0.0003$ & $0.0477\pm0.0003$ \\
    \textbf{APPNP} & $\bm{0.0427}\pm0.0003$ & $\bm{0.0421}\pm0.0002$ & $\bm{0.0417}\pm0.0002$ \\
    \bottomrule
  \end{tabular}}
  \label{tab:regression}
\end{table}

\begin{table}[t]
\centering 
  \caption{Final Performance Comparison between $k$NN Graph and Anchor-$k$NN Graph.}
  \renewcommand{\arraystretch}{0.9}
  \scalebox{0.8}{
  \begin{tabular}{c | c c  }
    \toprule
    \diagbox[]{E}{$\mathcal{G}$} & $k$NN & anchor-$k$NN  \\
    \midrule
    w2v $\leftarrow$ GloVe \\
    LSI & $0.0266\pm0.0002$ & $0.0282\pm0.0004$ \\
    APPNP & $0.0238\pm0.0002$ & $0.0234\pm0.0001$ \\
    \hline 
    w2v $\leftarrow$ FastText \\
    LSI & $0.0275\pm0.0002$ & $0.0294\pm0.0004$ \\
    APPNP & $0.0254\pm0.0002$ & $0.0251\pm0.0002$ \\
    \hline 
    GloVe $\leftarrow$ w2v \\
    LSI & $0.1437\pm0.0027$ & $0.1631\pm0.0038$ \\
    APPNP & $0.1011\pm0.0007$ & $0.1020\pm0.0007$ \\
    \hline 
    GloVe $\leftarrow$ FastText \\
    LSI & $0.1275\pm0.0005$ & $0.1358\pm0.0012$ \\
    APPNP & $0.1058\pm0.0006$ & $0.1066\pm0.0005$ \\
    \hline 
    FastText $\leftarrow$ w2v \\
    LSI & $0.0525\pm0.0008$ & $0.0581\pm0.0010$ \\
    APPNP & $0.0429\pm0.0002$ & $0.0432\pm0.0002$ \\
    \hline 
    FastText $\leftarrow$ GloVe \\
    LSI & $0.0484\pm0.0002$ & $0.0511\pm0.0005$ \\
    APPNP & $0.0427\pm0.0003$ & $0.0421\pm0.0003$ \\
    \bottomrule
  \end{tabular}}
  \label{tab:sol_comp}
\end{table}

\subsection{Language Modeling}

In this task, we still focus on GloVe~\cite{pennington2014glove}, fastText~\cite{bojanowski2017enriching} and Word2vec~\cite{mikolov2013distributed}.
We use different imputation methods and compare the quality of the imputed embeddings by looking at the language modeling perplexity.

To construct the language model corpus, we sample the first 20000 sentences of the preprocessed PubMed texts from the implementation provided by \cite{BlueBert}, with a train/validation/test split ratio of 0.8/0.1/0.1. The sampled corpus contains 28436 unique words, out of which 15984, 17473 and 13210 are available in GloVe, fastText and Word2vec, respectively. We have BioWordVec~\cite{bioword2vec} vectors as the side information to build the graph, which contains all 28436 words. We use the PyTorch official implementation 
 which trains a multi-layer LSTM on a language modeling task. 

As previously explained, we use pretrained embeddings as the weight initialization for the first layer in LSTM. 
The base performance is the embedding without any imputation, which means the missing word embeddings are randomly initialized. 
For the other methods, the missing word embeddings are imputed with the prior knowledge provided by BioWordVec. 
We train the LSTM model on the training set and report the perplexity on the test set. The embedding layer is fixed during training for all experiments and the default settings from the official PyTorch implementation are used for the LSTM. 
The number of neighbors, $\delta$, is set to 8 for graph construction. We run each experiment 20 times and compute the average perplexity and the standard deviation on the test set.

From Table~\ref{tab:language_model}, we observe that all methods are able to improve the performance of the base model and the neural network based methods are outperforming LSI. Overall, GCN slightly outperforms MLP while APPNP slightly outperforms GCN.

\begin{table}[]
\centering 
  \caption{Results for the Language Modeling Task where APPNP Slightly Outperforms other Methods.}
  \renewcommand{\arraystretch}{0.9}
  \scalebox{0.9}{
\begin{tabular}{ c|c c } 
\toprule
embedding & model & perplexity \\
\midrule
\multirow{5}{4em}{w2v} & Base & $224.902 \pm3.282$ \\ 
                        & LSI & $213.743 \pm2.117$\\ 
                        & MLP & $210.707 \pm2.725$\\
                        & GCN & $210.718 \pm1.897$ \\ 
                        & \textbf{APPNP} & $\bm{209.673}\pm2.384$\\ 
\hline
\multirow{5}{4em}{GloVe} & Base & $216.186 \pm2.185$ \\ 
                        & LSI & $208.393 \pm2.489$ \\ 
                        & MLP & $206.613 \pm1.858$  \\
                        & GCN & $206.380 \pm1.928$  \\ 
                        & \textbf{APPNP} & $\bm{206.178}\pm2.161$  \\ 
\hline
\multirow{5}{4em}{FastText} & Base & $213.586 \pm2.529$ \\ 
                            & LSI & $207.594 \pm2.518$ \\ 
                            & MLP & $205.559 \pm2.404$ \\
                            & GCN & $205.145 \pm1.769$\\ 
                            & \textbf{APPNP} & $\bm{205.004}\pm2.547$\\ 
\bottomrule

\end{tabular}}
  \label{tab:language_model}
\end{table}

%% file: chapter4.tex
\chapter{Improving GCN with Eigenvalue Perturbation}

Although graph convolutional neural network has found its wide applications in various domains and achieved promising results, one of its assumptions does not always firmly hold, i.e., the underlying graph structure is optimal in the sense of leading us to the best learning outcome in the given task. When the graph topology is given, for example, a citation network is given and we want to classify some scholarly articles into different research fields based their citation relation and the textual information, how do we know the citations are all accurate and precise, such that the citation network aids the classification process in the best way? After all, the reference part never misses its role in concerning the journal editors and paper reviewers when we submit an article or judge other people's work. Furthermore, when the graph topology is not given and instead manually constructed like in Algorithm~\ref{anchor-graph} from the previous chapter, how can we be sure such a construction is optimal in obtaining the best model performance? 

The question seems open. However, we can always try to modify the graph such that the model performance can be hopefully further improved. Moreover, rather than modifying the graph in its original domain, i.e., directly modify the graph adjacency matrix or the derived graph filter matrix, one can perturb the eigenvalues of the matrix to achieve efficient modification in a data-driven way. In this chapter, I will discuss the related topics and introduce our recent work in this direction. 

\section{Spectral Motivation}
\label{sub:ssc-gcn}

The main motivation to propose a mechanism that perturbs the eigenvalues of the graph filter matrix $S$ is that many existing hand-crafted graph filter matrices are indeed perturbing the eigenvalues of some sort. 
Typically, the graph filter matrix $S$ in GCN is obtained from the adjacency matrix $A$ or the graph Laplacian $L$. A commonly used filter is the symmetrically normalized adjacency (SNA) matrix with added self-loops \cite{kipf2016semi}, i.e., 
\begin{equation}
S_{\mathrm{SNA}} = \tilde{D}^{-\frac{1}{2}} \tilde{A} \tilde{D}^{-\frac{1}{2}},
\end{equation}
where $\tilde{A} = A+I$ and $\tilde{D} = \mbox{diag}(\tilde{A}\textbf{1})$. 
Other examples include 
higher-order polynomials with some normalization \cite{gavili2017shift}.
\cite{sun2020fisher} introduced a more sophisticated filter matrix $S_{\mathrm{DW}}$ by preprocessing the graph adjacency matrix $A$ based on DeepWalk similarities \cite{perozzi2014deepwalk} and termed the resulting Graph Convolutional Networks $\mbox{GCN}^T$.


A commonality among the existing filters is that they can all be interpreted as some manipulation of the eigenvalues of the given graph adjacency or Laplacian.
Taking a higher-order polynomial filter as an example, when the graph is undirected, it is easy to show the polynomial construction of graph filter is equivalent to the polynomial of the  eigenvalues, in that given $M = V\Lambda V^T$,
$$
\sum_{i=0}^{k}\theta_i M^i = \sum_{i=0}^{k}\theta_i (V\Lambda V^T)^i = \sum_{i=0}^{k}\theta_i V\Lambda^i V^T = V \left(\sum_{i=0}^{k}\theta_i \Lambda^i\right) V^T.
$$
Another example is the inverse shifted Laplacian \cite{jiang2019data}
$$\tilde{L}^{-1} = (\tilde{D}^{-\frac{1}{2}} (\tilde{D}-\tilde{A}) \tilde{D}^{-\frac{1}{2}})^{-1},$$
where $\tilde{A} = A + \theta I$ and $\tilde{D} = \mbox{diag}(\tilde{A}\textbf{1})$. 
Given $\tilde{L} = V \Lambda V^T$, the inverse matrix is calculated as follows:
$$
\tilde{L}^{-1} = (V \Lambda V^T)^{-1} = (V^T)^{-1} \Lambda^{-1} V^{-1} = V \Lambda^{-1} V^T,
$$
where $\Lambda^{-1}_{ii} = 1/\lambda_i$. Hence, the inverse shifted Laplacian 
as a graph filter matrix can also be interpreted as a manipulation of the eigenvalues. This observation motivates us to formulate our approach also as manipulation of the eigenvalues. Instead of working on the full spectrum, we propose to manipulate the eigenvalues of a graph filter matrix selectively to improve its performance with low computational cost.

\subsection{Optimal Low-Rank Approximations and Minimal Perturbation}
\label{sub:olra}

We develop our approach by starting with an approximation of the graph filter matrix $S$. Our point of departure is the following well-known fact regarding the optimal low-rank approximations.

\begin{lemma}\label{lem:optimality}Given a symmetric matrix $A\in \mathbb{R}^{n\times n}$ with an eigendecomposition $A=V\Lambda V^T$, where the eigenvalues in $\Lambda$ are in descending order in magnitude, i.e., $\vert \lambda_1 \vert \geq \vert \lambda_2 \vert \geq \dots \geq \vert \lambda_n \vert$, the matrix $A_k = \sum_{i=1}^{k}\lambda_i v_i v_i^T$  satisfies
\begin{equation}\label{eq:optimal-2norm}
 A_k = \arg \min_{B\in \mathbb{R}^{n\times n} \wedge \mbox{rank}(B)\le k}\Vert A - B \Vert.
\end{equation}
\end{lemma}
The lemma directly follows from the Eckart-Young theorem \cite[p. 79]{golub2013matrix}.
Lemma~\ref{lem:optimality} is more relevant to the classical graph shift operators as in \cite{bruna2013spectral} and \cite{henaff2015deep}, but it is not directly applicable in our setting, since we perturb the eigenvalues of the filter matrix. The following theorem shows that the optimal low-rank approximation is indeed a good choice for perturbing the filter.

\begin{theorem}\label{thm:minimal-pert}
Given a symmetric matrix $A\in \mathbb{R}^{n\times n}$, its optimal rank-$k$ approximation $A_k$ in Lemma~\ref{lem:optimality} also minimizes the maximum perturbation in $A-B$ for all $B\in\mathbb{R}^{n\times n}$ of rank $k$ or less and all perturbations $\delta\in\mathbb{R}^{n\times n}$ with $\Vert \delta\Vert=c$ for some constant $c>0$, i.e.,
\begin{equation}
A_k = \sum_{i=1}^{k}\lambda_i v_i v_i^T = \arg \min_{\substack{B\in \mathbb{R}^{n\times n}\\ \mbox{rank}(B)\leq k}} \max_{\substack{\delta \in \mathbb{R}^{n\times n}\\ \Vert \delta\Vert = c}} \Vert A-B+\delta\Vert.
\end{equation}
\end{theorem}

\begin{proof} Given any $B\in \mathbb{R}^{n\times n}$, let $\hat{u}_1\in \mathbb{R}^n$ and $\hat{v}_1\in {R}^n$ denote the left and right singular vectors corresponding to the largest singular value of $A-B$. Due to the Cauchy-Schwartz inequality,
$$
\Vert A - B + \delta \Vert \leq \Vert A - B \Vert + \Vert \delta \Vert,\label{eq:Cachy-Schwartz-inequality}
$$
where the inequality is an equality when $\delta =c \hat{u}_1 \hat{v}_1^T$. Hence, 
$$\min_{\substack{B\in \mathbb{R}^{n\times n}\\ \mbox{rank}(B)\leq k}} \max_{\substack{\delta \in \mathbb{R}^{n\times n}\\ \Vert \delta\Vert = c}} \Vert A-B+\delta\Vert =\min_{\substack{B\in \mathbb{R}^{n\times n}\\ \mbox{rank}(B)\leq k}} \Vert A-B\Vert + c,$$ which is minimized iff $\Vert A-B\Vert$ is minimized, i.e., $B=A_k$ due to Lemma~\ref{lem:optimality}.
\end{proof}

In plain English, Theorem~\ref{thm:minimal-pert} states that the worst-case deviation from the graph filter $S$ is minimized when the filter is constructed from the optimal low-rank approximation. It is worth noting that Lemma~\ref{lem:optimality} also holds in the Frobenius norm, so does Theorem~\ref{thm:minimal-pert}. We omit their proofs.

In practice, however, the graph filter matrix $S$ is typically asymmetric, even in the case of SNA. In general, we could apply the Eckart-Young theorem in place of Lemma~\ref{lem:optimality} and use a truncated singular value decomposition (TSVD) of an asymmetric $S$ to construct the graph filter. However, if $S$ is not too far from symmetry, it is more efficient (in terms of both computational cost and memory requirement) to use the eigenvalue decomposition of  $(S+S^T)/2$ to construct a ``near-optimal" low-rank approximation. In this work, we use the latter approach for its better efficiency. We found it to be effective for applications such as citation networks.

\subsection{Connection of Eigenvalue Perturbation with Residual Learning}
One limitation of using a low-rank approximation as 
described in Section~\ref{sub:olra} is that such low-rank approximations only contain the low-frequency signals, and a filter based on low-rank approximation alone may miss some important information in the high-frequency band. To overcome this limitation, one can introduce a ``residual unit" into the neural network, analogous to
the residual learning in HighwayNet \cite{srivastava2015highway} and ResNet \cite{he2016deep}. 

In this work, we propose a similar yet different idea.
The objective is different from the aforementioned residual learning, in that
we design a novel residual unit in the frequency domain, 
to better learn the low-frequency signals while preserving the original signals 
in both low-frequency and high-frequency bands, and in turn improve the performance. 
We accomplish the residual learning in EigLearn by training a constant number of free parameters that serve as the perturbation to the significant eigenvlaues of the graph filter matrix. One benefit of our residual learning formulation is that it does not require a complete eigendecomposition and incurs no steep computation cost.
In other words, the residual unit in EigLearn is composed of a collection of actual neurons that resembles the biases, as we will explain in detail in Section~\ref{sec:EigLearn}.

\begin{figure}[tbhp]
  \centering
    \includegraphics[scale=0.8]{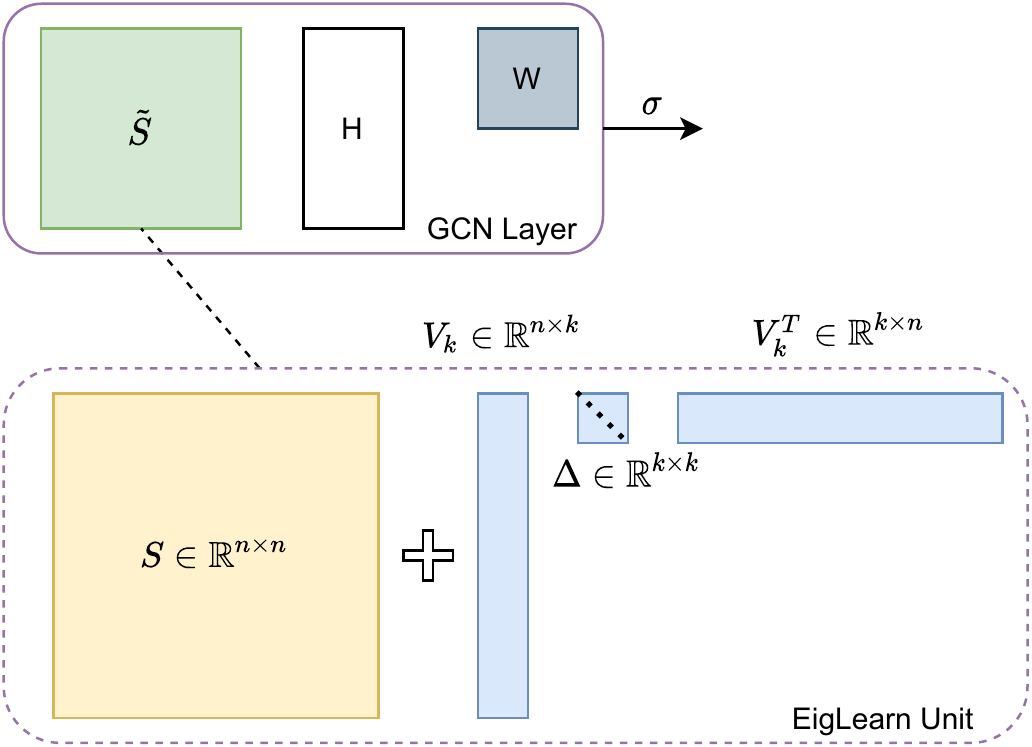}
    \caption{Schematic of one-layer of GCN with EigLearn.}
  \label{fig:eiglearn}
\end{figure}

\textbf{Figure}~\ref{fig:eiglearn} depicts the schematic of one-layer of GCN with EigLearn. $H$ denotes the feature matrix, $W$ is the linear mapping matrix and $\tilde{S} = S + V_k \Delta V^T_k$ is the perturbed graph filter matrix with $S$ being the original filter matrix. In the perturbation unit, $V_k$ is composed of $k$ significant eigenvectors of $S$. The learnable parameters in $\Delta$ represents the eigenvalue perturbation and is realized as a residual unit in the neural network.

\section{Learning Eigenvalue Perturbation}
\label{sec:EigLearn}

As discussed in Section~\ref{sub:ssc-gcn}, our approach is motivated by the fact that many GCN models exploit certain types of eigenvalue manipulation on the adjacency matrix $A$ or the Laplacian matrix $L$ and construct an effective graph filter matrix. The examples include but are not limited to, subspace constructed from the graph Laplacian with learnable coefficients \cite{bruna2013spectral}\cite{henaff2015deep}, 
graph shift operator constructions \cite{dasoulas2021learning}\cite{defferrard2016convolutional}
and inverse Laplacian \cite{jiang2019data}. We develop our approach by first raising the following questions:
\begin{itemize}
    \item Given a graph filter matrix, either hand-crafted or parameterized, 
    can we further perturb it using a data-driven approach and make it more effective in learning graph signals?
    \item The polynomial graph filter matrices are based on manipulation on all eigenvalues, 
    which can be a waste of computation and a waste of model capacity if the polynomial is parameterized. 
    Can we selectively perturb a subset of eigenvalues efficiently and effectively?
    \item Can we do this without increasing the overall complexity of GCN? 
\end{itemize}
These questions also serve as the design principles of our method.

\subsection{Perturbing the Eigenvalues}

Without loss of generality, let us assume the graph filter matrix $S \in \mathbb{R}^{n\times n}$ is symmetric, since we have addressed the general treatment of asymmetric matrices in Section~\ref{sub:olra}. Let $S = V\Lambda V^{T}$ denote the eigendecomposition of $S$.
We hypothesize that there is a sizable room for performance improvement even in those well-designed graph filter matrices for GCN, if we introduce some modifications to the graph filter matrix $S$. One might be tempted to directly treat all entries in $S$ as free parameters, but such an approach would incur at least quadratic complexity and would not reveal insights on the spectral properties.

Instead, we propose to modify the eigenvalues of $S$ and optimize the filter in the spectral domain. Because the modification is often small, conceptually it corresponds to a perturbation process, i.e., 
\begin{equation}
\tilde{\Lambda} = \Lambda + \Delta = \mbox{diag}(\lambda_1+\delta_1, ..., \lambda_n+\delta_n).
\end{equation}

One potential challenge of the proposed approach is the complexity of the eigendecomposition in the first place. A straightforward implementation based on eigendecomposition introduces cubical complexity and hinders scalability. Instead,  we can improve the GCN performance with the proposed approach with only additional linear complexity w.r.t. the number of nodes and a constant number of learnable parameters. This is done by learning the perturbation on a constant number, $k$, of significant eigenvalues.  Note that most of the graphs in real-world applications are sparse. Hence, extracting a significant eigenvector in the sparse system has linear-time complexity in the number of edges (and equivalently the number of nodes) per Lanczos iteration. Since it typically suffices to use a low-precision approximation to the leading eigenvectors, we expect the number of Lanczos iterations can be limited to a small constant. Hence, solving for $k$ significant eigenvectors in a sparse system is approximately $O(kn)$. Given $\Delta = \mbox{diag}(\delta_1, ..., \delta_k)$, 
\begin{equation}
\tilde{\Lambda} = \Lambda + \Delta = \mbox{diag}(\lambda_1+\delta_1, ..., \lambda_k+\delta_k, \lambda_{k+1},...,\lambda_n).
\end{equation}
Note that it is unnecessary to have a full matrix decomposition. The above equation is the same as follows:  
\begin{equation}
\tilde{S} = S + V_k\Delta V_k^{T}
\end{equation}
where $V_k$ are pre-computed and stored. It constitutes residual learning where $S$ is the original matrix and the perturbation is the residual component. To generalize the approach to dense matrices $S$ without suffering from quadratic complexity when solving for $V_k$, one can employ the promising matrix sparsification \cite{achlioptas2007fast} and graph sparsification \cite{spielman2011spectral,koutis2015faster} methods that preserve the spectral properties within a provable error bound. Although the sparsification on a dense system takes quadratic complexity w.r.t. the number of nodes, it is inexpensive in practice and can be done reasonably fast for very large-scale situations.

\subsection{Analysis of Complexity}

One of the most important hyperparameters in EigLearn is $k$, which should be large enough to introduce sufficient capacity to the model and yet small enough to prevent overfitting and retain the complexity. Practically, one could apply grid search to identify an optimal $k$ in the same way as finding the optimal number of neurons. In our experimental study, we found that a small $k$ (30--40) achieves optimal performance and the EigLearn is robust within a reasonable range of $k$ (20-150).

When computing $V_k$, we observe linear complexity w.r.t. $|\mathcal{E}|$, i.e., $O(kr|\mathcal{E}|)$, 
where $k$ is constant and not proportional to $|\mathcal{V}|$, $r$ is the number of Arnoldi iterations and constant with low precision, 
and $|\mathcal{E}|$ is the number of edges in the graph. 
In most applications, the graph is sparse and $|\mathcal{E}|$ is linear to $|\mathcal{V}|$, and therefore computing $V_k$ is $O(kr|\mathcal{V}|)$, i.e., linear in $|\mathcal{V}|$. 
For dense graphs, the evaluation of $A\cdot X$ or $S\cdot X$ is already quadratic, and computing $V_k$ is also quadratic. 
Hence, EigLearn does not change the asymptotic complexity of the GCN model in either dense or sparse cases. 
It’s worth mentioning that with a relatively larger error tolerance in the eigen solver, EigLearn still works well.

\subsection{Neural Architecture Setup and Model Training}

Let us first use one GCN layer as an illustration and omit the bias term for simplicity. The forward propagation is: 
\begin{equation}
H^{l+1} = \mbox{ReLU}(\tilde{S} H^{l} W).
\end{equation}
Substituting $\tilde{S} = S + V_k\Delta V_k^{T}$, we have 
\begin{equation}
H^{l+1} = \mbox{ReLU}((S + V_k\Delta V_k^{T}) H^{l} W).
\end{equation}
We follow a two-stage training procedure. Firstly $\Delta$ is initialized as $0$ and fixed.  Stage-I is equivalent to a regular GCN training and  
$$
H^{l+1} = \mbox{ReLU}(S H^{l} W). 
$$
 The parameters in $W$ (and the bias term if there is any) are updated in stage-I. In stage-II, we keep $W$ fixed and update $\Delta$.

Assume we employ empirical risk minimization learning schema given some predefined loss function and let $\mathcal{L}$ denote the empirical loss back-propagated to this GCN layer. 
The gradient descent on $W$ in stage-I training is
\begin{equation}
W \leftarrow W - \eta \cdot \nabla_{W} \mathcal{L}
\end{equation}
and the gradient descent on $\Delta$ in stage-II training is
\begin{equation}
\Delta \leftarrow \Delta - \eta \cdot \nabla_{\Delta} \mathcal{L}
\end{equation}
where $\eta$ is the learning rate. 

In our experimental study, we use a two-layer GCN for illustration. 
To further reduce the degree of freedom and prevent potential overfitting, we share $\Delta$ (of which the $k$ diagonal entries are trainable) in both layers, i.e., 
\begin{equation}
\tilde{S}^{(1)} = \tilde{S}^{(0)} = \tilde{S} = S + V_k\Delta V_k^{T}.
\end{equation}
Hence, the GCN output is
\begin{equation}
\hat{Y} = \mbox{softmax}(\tilde{S}(\mbox{ReLU}(\tilde{S}XW^{(0)}))W^{(1)}).
\end{equation}

\section{Empirical Study}
\label{sec:experiment}

To assess the effectiveness of the proposed approach against other competitors, we conducted semi-supervised node classification on three benchmark datasets, namely Cora, CiteSeer and PubMed, and compared our approach with LanczosNet \cite{liao2019lanczosnet} and FisherGCN \cite{sun2020fisher} based on their corresponding experimental setups.


\subsection{Comparison with LanczosNet}
\label{sub:compare-lanczosnet}

We first compare EigLearn with LanczosNet  \cite{liao2019lanczosnet}, since it is probably the most similar to our approach in that it also has a learnable filter based on approximate eigenvectors. The experimental study in \cite{liao2019lanczosnet} had a focus on 
multi-scale molecule regression besides semi-supervised node classification. Since the publicly available implementation of LanczosNet does not include the semi-supervised classification, we adapted the problem setup of EigLearn
based on that in \cite{liao2019lanczosnet}. In particular,
we adopted the public fixed split in this comparison.
In addition, according to the publicly available implementation, LanczosNet did not employ sparse dropout. 
For a fair comparison, we implemented EigLearn 
both with and without sparse dropout. 

\textbf{Table~\ref{tab:compare LanczosNet}} compares EigLearnGCN with and without sparse dropouts with LanczosNet and AdaLanczosNet in \cite{liao2019lanczosnet}. Since the testing loss was not given in \cite{liao2019lanczosnet}, we only report testing accuracy in \textbf{Table~\ref{tab:compare LanczosNet}}.  EigLearnGCN consistently delivered better performance than LanczosNet regardless of whether 
sparse dropout is utilized. It is worth noting that the 
results in \cite{liao2019lanczosnet} were obtained using
the fine-tuned hyperparameters for different cases, but
EigLearnGCN used the default parameters for all the cases.
These results show that EigLearn is not only more accurate
but also more robust than LanczosNet in terms of parameter tuning.
The superiority of EigLearn is probably because EigLearn
only perturbs the dominant eigenvalues to construct minimal perturbations based on Theorem~\ref{thm:minimal-pert}, whereas LanczosNet perturbs all the approximate extreme eigenvalues (i.e., the Ritz values). The additional parameters associated with the smallest eigenvalues in LanczosNet are unlikely to improve accuracy, and their presence may cause LanczosNet to be more prone to overfitting and hence lower performance.
We also noticed that the baseline performance of 
GCN in \cite{liao2019lanczosnet} was worse than that in \cite{kipf2016semi}, but no explanation was provided in
\cite{liao2019lanczosnet}.

\begin{table}[t]
  \centering
  \caption{Performance Comparison of EigLearnGCN with and without Sparse Dropout versus LanczosNet and AdaLanczosNet (without Dropout)}
    \begin{tabular}{c|ccc}
    \hline
    & \multicolumn{3}{c}{Testing Accuracy} \\
    \hline model & Cora & CiteSeer & PubMed  \\
    \hline
    LanczosNet    & $79.5\pm1.8$ & $66.2\pm1.9$ & $78.3\pm0.3$ \\
    AdaLanczosNet             & $80.4\pm1.1$ & $68.7\pm1.0$ & $78.1\pm0.4$ \\
    EigLearnGCN (w/ dropout)  & ${\bf 82.1}\pm0.2$ & $70.3\pm0.8$ & $79.2\pm0.5$ \\
    EigLearnGCN (w/o dropout) & $81.8\pm0.3$ & ${\bf 70.7}\pm0.7$ & ${\bf 79.3}\pm0.2$ \\
    \hline
    \end{tabular}
    
  \label{tab:compare LanczosNet}
\end{table}  

\subsection{Comparison with FisherGCN}
\label{sub:compare-fishergcn}
The second assessment focuses on the comparison between  EigLearn and FisherGCN \cite{sun2020fisher} that 
is arguably more sophisticated than LanczosNet. In this comparison, we adopted the same settings for data splitting and training as in \cite{sun2020fisher}. In particular, we split the data into a training set with 20 samples per class, a validation set with 500 samples, and a testing set with 1000 samples. 
We ran experiments with 20 random splits and for each data split there were 10 different initializations per split.
As thoroughly studied in \cite{shchur2018pitfalls}, a random split is a less biased evaluation setting for GCN 
performance, and hence is preferable over the public fixed split.
The other parameters remained the same in EigLearn as in 
Section~\ref{sub:compare-lanczosnet}, which is consistent with those in \cite{sun2020fisher}.

\begin{table}[htbp]
  \centering
  \caption{Comparison of Model Performances EigLearn vs. FisherGCN in Terms of Average Values and Standard Deviations}
  \setlength\tabcolsep{3pt}\label{tab:compare FisherGCN}
  \centerline{
  \scalebox{0.9}{
  \begin{tabular}{c|ccc|ccc}
    \hline
    & \multicolumn{3}{c|}{Testing Accuracy} & \multicolumn{3}{c}{Testing Loss} \\
    \hline model & Cora & CiteSeer & PubMed & Cora & CiteSeer & PubMed \\
    \hline
    GCN     & $80.5\pm2.3$ & $69.6\pm2.0$ & $78.2\pm2.4$
                                & $1.07\pm0.04$ & $1.36\pm0.03$ & $0.75\pm0.04$ \\
    FisherGCN                   & $80.7\pm2.2$ & $69.8\pm2.0$ & $78.4\pm2.4$
                                & $1.06\pm0.04$ & $1.35\pm0.03$ & $0.74\pm0.04$ \\
    GCN (PT)           & $79.8\pm2.1$ & $69.7\pm1.9$ & $78.3\pm2.2$
                                & $0.66\pm0.04$ & $0.96\pm0.04$ & $0.58\pm0.05$ \\
    \textbf{ELGCN}        & $\bm{81.4}\pm1.5$ & $\bm{70.1}\pm1.8$ & $\bm{78.9}\pm2.1$
                                & $\bm{0.59}\pm0.04$ & $\bm{0.94}\pm0.04$ & $\bm{0.57}\pm0.05$ \\
                                \hline
    GCN$^T$ & $81.2\pm2.3$ & $70.3\pm1.9$ & $79.0\pm2.6$
                                & $1.04\pm0.04$ & $1.33\pm0.03$ & $0.70\pm0.05$ \\
    FisherGCN$^T$               & $81.5\pm2.2$ & $70.5\pm1.7$ & $79.3\pm2.7$
                                & $1.03\pm0.03$ & $1.32\pm0.03$ & $0.69\pm0.04$ \\
    GCN$^T$ (PT)       & $81.1\pm1.8$ & $70.5\pm1.7$ & $79.3\pm2.0$
                                & $0.62\pm0.04$ & $0.92\pm0.04$ & $0.54\pm0.04$ \\
    \textbf{ELGCN$^T$}    & $\bm{82.2}\pm1.4$ & $\bm{70.6}\pm1.6$ & $\bm{79.8}\pm1.8$
                                & $\bm{0.56}\pm0.04$ & $\bm{0.90}\pm0.04$ & $\bm{0.53}\pm0.04$ \\
    \hline
  \end{tabular}  
  }
    }
\end{table}    

\textbf{Table~\ref{tab:compare FisherGCN}} reports a comparison among the baseline GCN, FisherGCN, and EigLearn using the graph filter matrices based on $S_\mathrm{SNA}$ and $S_\mathrm{DW}$, respectively. 
We used both classification accuracy and loss on testing set as the evaluation metrics. 
For the baselines, we report the performances of both TensorFlow-based implementations of GCN and $\mbox{GCN}^T$ in \cite{sun2020fisher} and our PyTorch-based implementation.\footnote{There are some subtle differences between the implementations in TensorFlow and PyTorch that lead to the performance discrepancies in different baseline implementations. For example,  TensorFlow realizes the weight decay with $L_2$ regularization and includes the penalty  in the total loss, while PyTorch implements the weight decay in the ADAM optimizer and excludes the penalty from the total loss.}
\textbf{Table~\ref{tab:compare FisherGCN}} shows that 
EigLearn consistently improved the baseline performances and reduced testing losses. Furthermore, EigLearnGCN and $\mbox{EigLearnGCN}^T$ noticeably outperformed FisherGCN and $\mbox{FisherGCN}^T$ correspondingly. EigLearn also yielded larger improvements to the baselines than both FisherGCN and $\mbox{FisherGCN}^T$.
These larger improvements are significant, since compared to FisherGCN, EigLearn is easier to implement and fits more naturally into the neural network architectures. In addition, EigLearnGCN is significantly faster than FisherGCN in that the additional cost associated with EigLearn is only a small fraction of the cost for training a regular GCN, while FisherGCN is several times slower than GCN as reported in \cite{sun2020fisher}. It is also worth noting that EigLearn  reduced the standard deviations of testing accuracy consistently, while $\mbox{FisherGCN}^T$ increased the standard deviation for PubMed. Hence, these experiment results confirm that EigLearn is more accurate and robust than $\mbox{FisherGCN}$ in terms of standard generalization.

\begin{figure}[t]
    \makebox[\textwidth][c]{\includegraphics[width=1.2\textwidth]{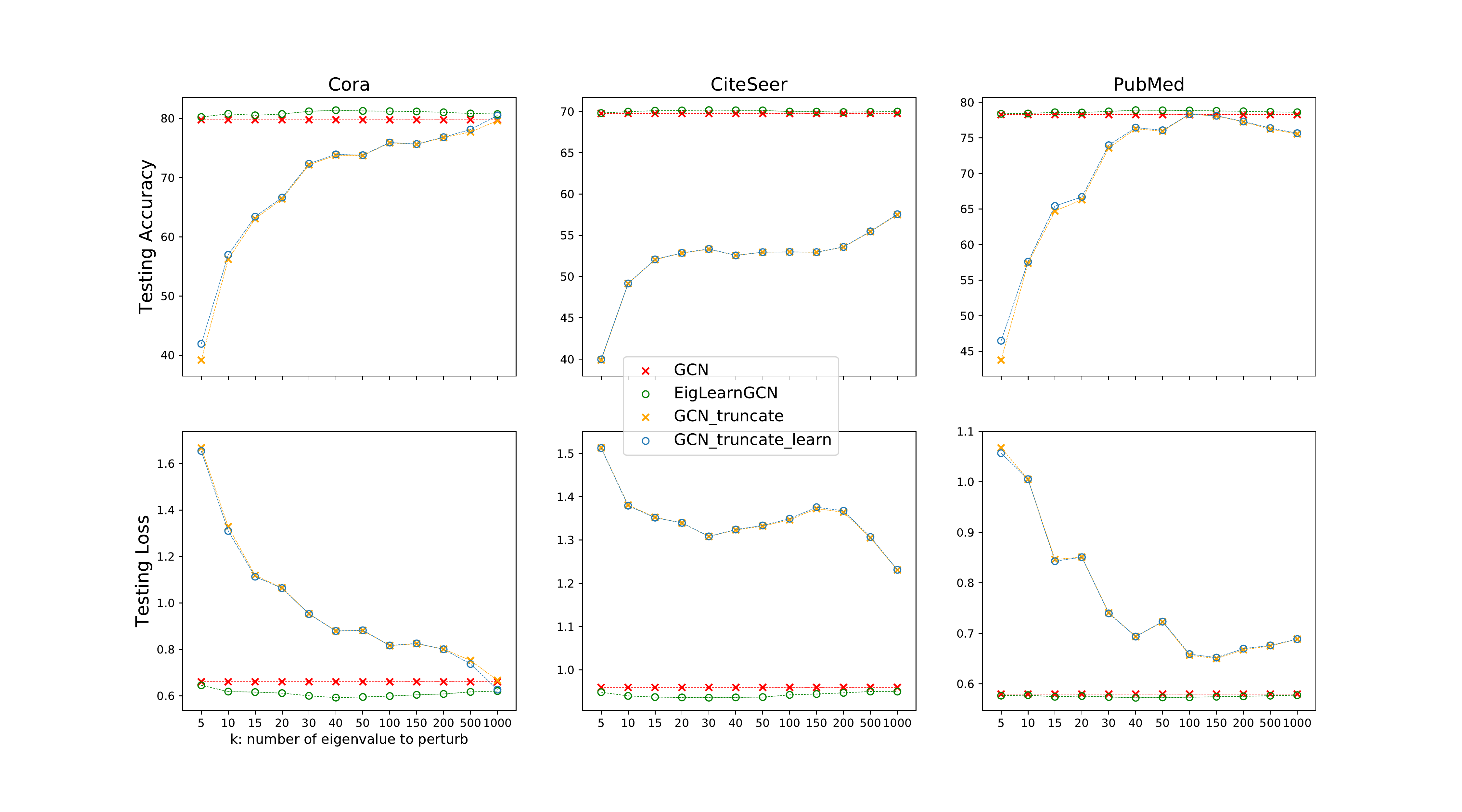}}
      \vspace{-0.5in}
     \caption{Performance comparison between EigLearn and TruncateTrain.}
    \label{fig:truncate_learn}
\end{figure}

\subsection{Apply EigLearn on ChebyNet and SGCN}
To testify the applicability of the EigLearn unit, we also inject it into ChebyNet~\cite{defferrard2016convolutional} 
and Simplified-GCN~\cite{wu2019simplifying}, 
and observe performance improvement on SGC and ChebyNet with the EigLearn unit as shown in \textbf{Table~\ref{tab:chebynet and sgc}}
using the same settings in the previous section.

The idea of SGCN is to first linearize the internal GCN layers by removing the activation functions, i.e., for a $k$-layer SGCN we have 
\begin{equation}
    H = S^kX\Pi_{i=1}^kW_{(i)},
\end{equation}
where $H$ is the feature map, $S$ is the graph filter matrix, $X$ is the input feature matrix and $W_{(i)}$ are the trainable weight matrix. The second difference between SGCN and GCN is to remove the extra weight matrices to prevent overfitting induced by large amount of free parameters, i.e., 
\begin{equation}
    H = S^kXW.
\end{equation}

ChebyNet has the graph convolution operation defined as 
\begin{equation}
    y = g_{\theta}(L)x = \Sigma_{i=1}^k \theta_i T_i(\tilde{L}) x,
\end{equation}
where $\theta_i$ is learnable, $\tilde{L} = 2L / \lambda_{max} - I$, $L$ is the original Laplacian, $\lambda_{max}$ is the largest eigenvalue of $L$ and $I$ is identity matrix, 
given the Chebyshev polynomials being $T_i(a) = 2aT_{i-1}(a) - T_{i-2}(a)$ with $T_0 = 1$ and $T_1=a$. 

\begin{table}[htbp]
  \centering
  \caption{Comparison of Model Performances Applying EigLearn on ChebyNet and SGC in Terms of Average Values and Standard Deviations}
  \label{tab:chebynet and sgc}
  \setlength\tabcolsep{3pt}\centerline{
  \scalebox{0.9}{
  \begin{tabular}{c|ccc|ccc}
    \hline
    & \multicolumn{3}{c|}{Testing Accuracy} & \multicolumn{3}{c}{Testing Loss} \\
    \hline model & Cora & CiteSeer & PubMed & Cora & CiteSeer & PubMed \\
    \hline
    ChebyNet                    & $76.5\pm2.4$          & $68.5\pm2.0$          & $75.9\pm2.6$
                                & $0.74\pm0.05$         & $0.98\pm0.04$         & $0.65\pm0.06$ \\
    \textbf{EL-ChebyNet}        & $\bm{78.5}\pm2.0$     & $\bm{69.1}\pm2.0$     & $\bm{76.3}\pm2.5$
                                & $\bm{0.68}\pm0.05$    & $\bm{0.96}\pm0.04$    & $\bm{0.65}\pm0.06$ \\
                                \hline
    SGC                         & $78.1\pm2.7$          & $69.2\pm2.1$          & $77.6\pm2.5$
                                & $0.81\pm0.03$         & $1.08\pm0.03$         & $0.58\pm0.03$ \\
    \textbf{EL-SGC}             & $\bm{81.0}\pm1.5$     & $\bm{69.6}\pm2.0$     & $\bm{79.2}\pm2.3$
                                & $\bm{0.61}\pm0.04$    & $\bm{0.97}\pm0.03$    & $\bm{0.55}\pm0.04$ \\
    \hline
  \end{tabular}  
  }
  }
\end{table}   

The ChebyNet model is based on an order-2 Chebyshev polynomial of the shifted and rescaled Laplacian. 
The SGC model is based on a power-2 graph filter matrix.\footnote{
The SGC performance in the original paper is based on the public fix split. 
We use the available implementation (\url{https://github.com/Tiiiger/SGC})
and perform random splits, and observe a performance degradation. 
The performance degradation on random split also happens to ChebyNet. 
}

\subsection{Comparison with TruncateTrain}
In the methodology section we argued that the residual formulation to learn the eigenvalue perturbation is necessary. 
To validate it, we also conducted experiments to compare EigLearn and the direct eigenvalue perturbation learning based on a truncated
eigendecomposition of the graph filter matrix (denote it as TruncateTrain). As shown in \textbf{Figure~\ref{fig:truncate_learn}},
two observations are apparent. Firstly, only when we used a large number of eigenvectors from the graph filter matrix, 
could we preserve the model performance. This observation invalidates the linear complexity assumption in TruncateTrain. 
Secondly, directly learning the eigenvalue perturbation could not boost the model performance as effectively as EigLearn utilizing the residual formulation. Although we do not provide theoretical analyses on this, the straightforward 
explanations are: the residual formulation anchors the base level of the eigenvalues and performance; on top of that, the perturbations are learned on individual eigenvalues without changing the meaningful attenuation pattern of the spectrum.

\subsection{Experiment on Large Dataset}
To demonstrate the scalability
and generalization of EigLearn, we also run experiments on a significantly larger dataset,
the arxiv network, which is roughly 10 times larger than PubMed in the number of nodes with more than 1 million edges,
from the Open Graph Benchmark~\cite{hu2021open}. We follow the detailed GCN settings provided in 
the official implementation from the OGB project. We train the eigenvalue perturbation with a learning rate of 0.002 
for 50 epochs without regularization or early stopping. For computer memory issue, we use an order-3 filter matrix in $\mbox{GCN}^T$.
We aggregate the results from ten runs (for both data random split and trainable parameter random initialization) 
in \textbf{Table~\ref{tab:arxiv}}. EigLearn  consistently improves on GCN with both filter matrices 
and across different numbers of eigenvalues. 
In term of algorithm efficiency, we compare the overall training time of GCN and EL-GCN (including computing $V_k$) 
and do not observe a significant time increase. The detailed run time comparison (repeated 10 times measured in seconds) is shown in \textbf{Table~\ref{tab:run_time}}.

\begin{table}[htbp]
  \centering
  \caption{Performance Improvement by EigLearn on ogbn-arxiv Dataset}
  \label{tab:arxiv}
\setlength\tabcolsep{1pt}  \centerline{
    \scalebox{0.9}{
  \begin{tabular}{c|ccc|ccc}
    \hline
    & \multicolumn{3}{c|}{Training Accuracy} & \multicolumn{3}{c}{Testing Accuracy} \\
    \hline model & k=20 & k=50 & k=100 & k=20 & k=50 & k=100 \\
    \hline
    GCN            & $78.84\pm0.57$ & $79.17\pm0.45$ & $78.94\pm0.59$
                                & $73.86\pm0.11$ & $73.60\pm0.10$ & $73.74\pm0.09$ \\
    \textbf{ELGCN}        & $\bm{80.27}\pm0.06$ & $\bm{80.45}\pm0.07$ & $\bm{80.27}\pm0.12$
                                & $\bm{74.14}\pm0.08$ & $\bm{73.86}\pm0.09$ & $\bm{73.98}\pm0.11$ \\
                                \hline
    GCN$^T$        & $79.80\pm0.42$ & $79.42\pm0.41$ & $79.73\pm0.49$
                                & $73.67\pm0.10$ & $73.61\pm0.11$ & $73.62\pm0.16$ \\
    \textbf{ELGCN$^T$}    & $\bm{81.45}\pm0.07$ & $\bm{81.43}\pm0.10$ & $\bm{81.13}\pm0.12$
                                & $\bm{74.08}\pm0.11$ & $\bm{74.02}\pm0.07$ & $\bm{73.99}\pm0.07$ \\
    \hline
  \end{tabular}  
  }
}
\end{table}

\begin{table}[htbp]
  \centering
  \caption{Run Time Comparison between GCN and ELGCN on Different Datasets}
  \label{tab:run_time}
\setlength\tabcolsep{3pt}  \centerline{
    \scalebox{1}{
  \begin{tabular}{c|cccc}
    \hline
    & \multicolumn{4}{c}{Dataset} \\
    \hline model & Cora & CiteSeer & PubMed & ogbn-arxiv \\
    \hline
    GCN            & 3.06 & 3.44 & 6.44 & 213.9  \\
    ELGCN       & 4.18 & 4.16 & 7.83 & 273.6   \\
    \hline
  \end{tabular}  
  }
}
\end{table}

\subsection{Other Experiment Results}
\begin{figure}[t]
    \makebox[\textwidth][c]{\includegraphics[width=1\textwidth]{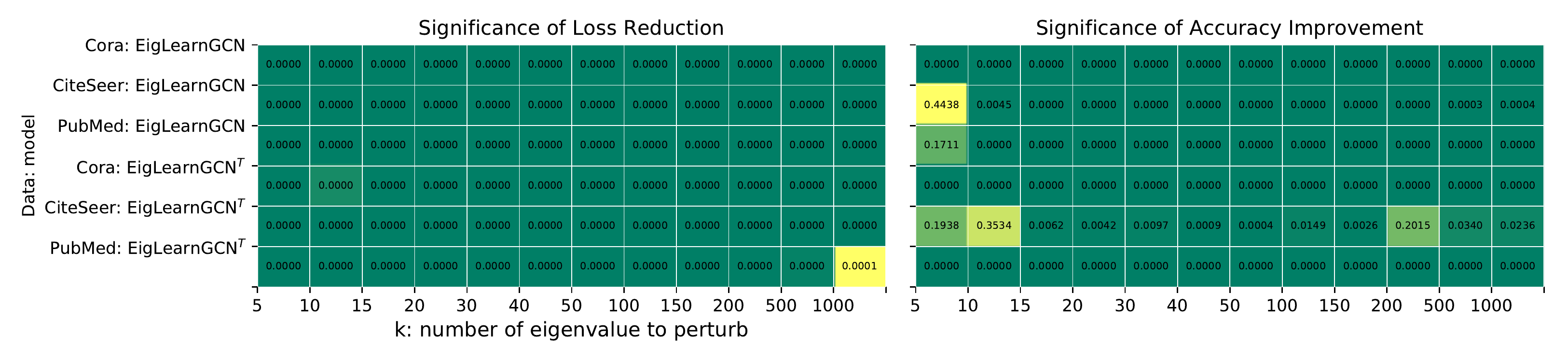}}
    \caption{Pair-Sample t-test on EigLearn performance improvement.}
    \label{fig:significance}
\end{figure}

To further demonstrate that the improvement induced by EigLearn is systematic, we conducted
pair-sample t-tests across different datasets, graph filter matrices 
(GCN means symmetrically normalized adjacency and GCN$^T$ is a higher-order polynomial) with perturbation on different $k$ eigenvalues. As shown in \textbf{Figure~\ref{fig:significance}}, the t-tests reveal that the majority of experiment settings
are statistically significant except for a few corner cases (e.g., when $k$ is small). 

\begin{figure}[t]
    \makebox[\textwidth][c]{\includegraphics[width=1.2\textwidth]{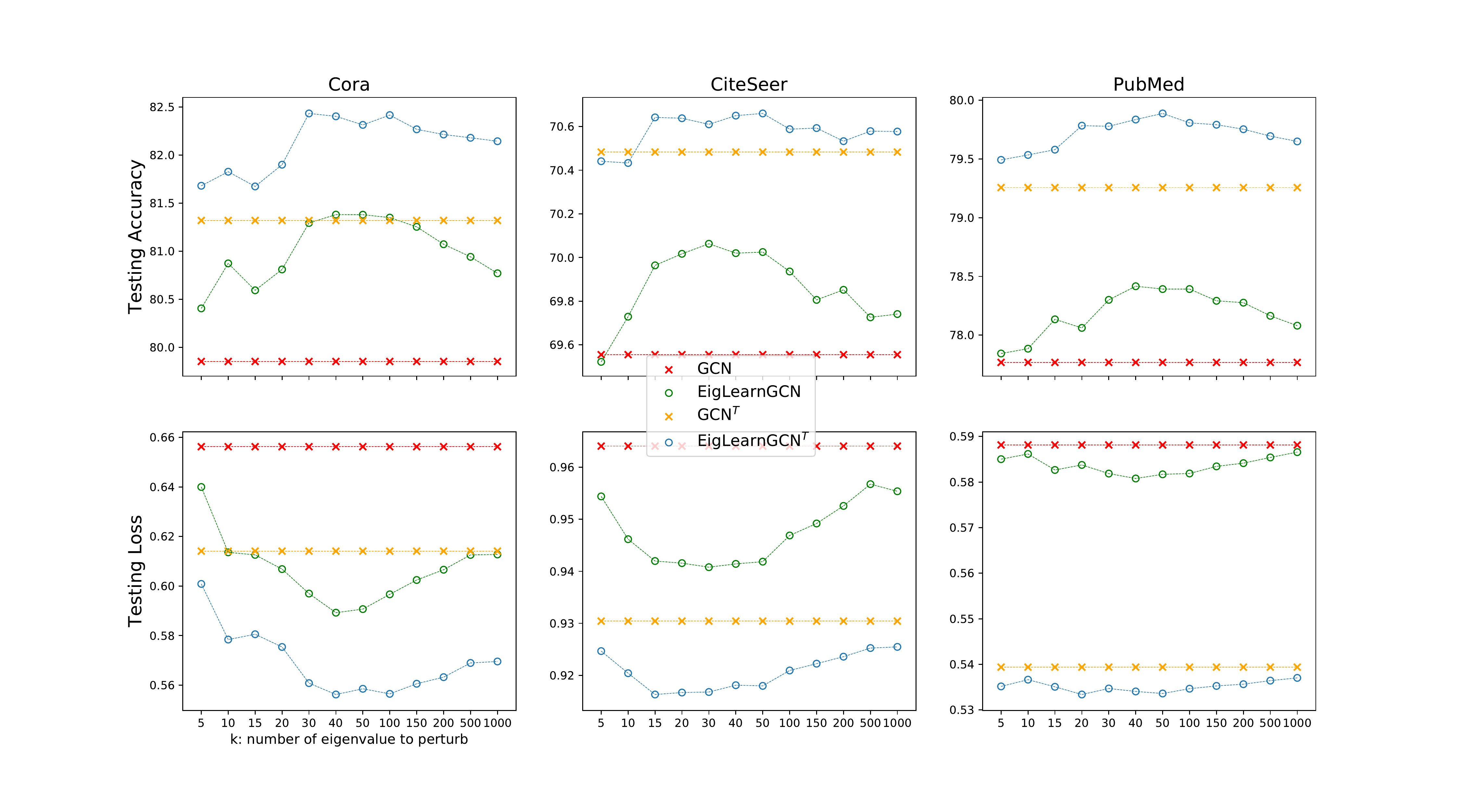}}
    \vspace{-0.5in}
    \caption{EigLearn sensitivity on $k$.}
    \label{fig:k_sensitivity}
\end{figure}

Next, we show the sensitivity analysis on the EigLearn in \textbf{Figure~\ref{fig:k_sensitivity}}. Essentially, EigLearn extracts the 
perturbation in the subspace spanned by $k$ dominant eigenvectors of the graph filter matrix. 
One observation from \textbf{Figure~\ref{fig:k_sensitivity}} is that the performance induced by EigLearn is relatively consistent within a reasonably 
wide band of $k$ value (e.g., 20 to 150).  It also confirms that we do not need a large number of eigenvectors to make EigLearn 
achieve high effectiveness. On the contrary, when $k$ is large, the performance improvement diminishes. 
It is most probably due to overfitting when many trainable parameters are introduced to the model. 
When $k$ is too small, the extra capacity added to the model is not sufficient for EigLearn to make 
sizable improvement. Besides, as a common practice, we conduct perturbation in the subspace spanned by 
the dominant eigenvectors instead of the least significant eigenvectors. We also ran experiments with the least 
significant eigenvectors.  These eigenvectors barely made any performance improvement. 
This provides some experimental justification of using the significant eigenvectors and supports 
our argument that polynomial-based approaches may waste computation and model capacity on a large number of insignificant eigenvalues when the implicit perturbation is applied to the full spectrum.   

\begin{figure}[htbp]
    \makebox[\textwidth][c]{\includegraphics[width=1.2\textwidth]{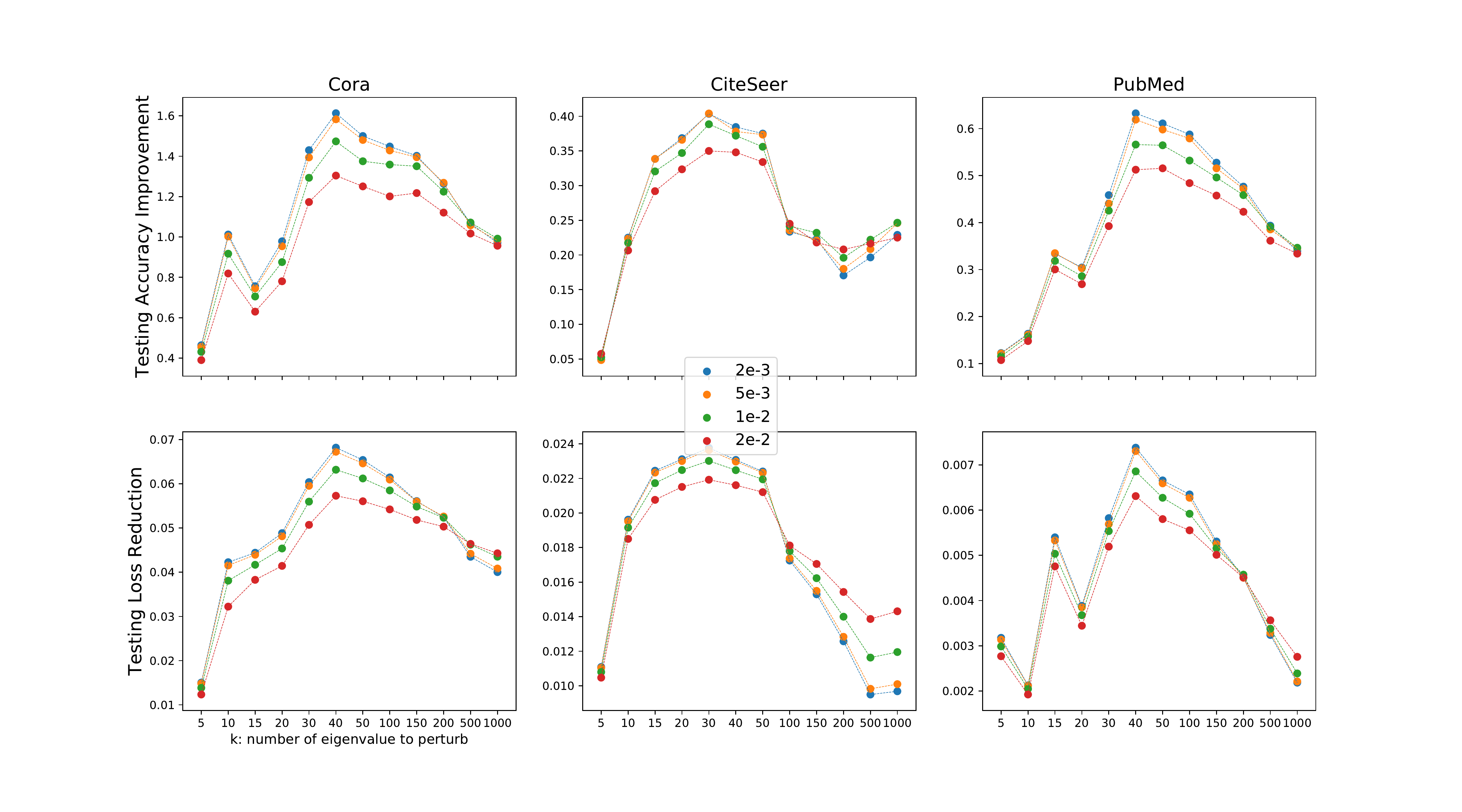}}
    \caption{EigLearn sensitivity on learning rate.}
    \label{fig:lr_sensitivity}
\end{figure}
\begin{figure}[htbp]
    \makebox[\textwidth][c]{\includegraphics[width=1.2\textwidth]{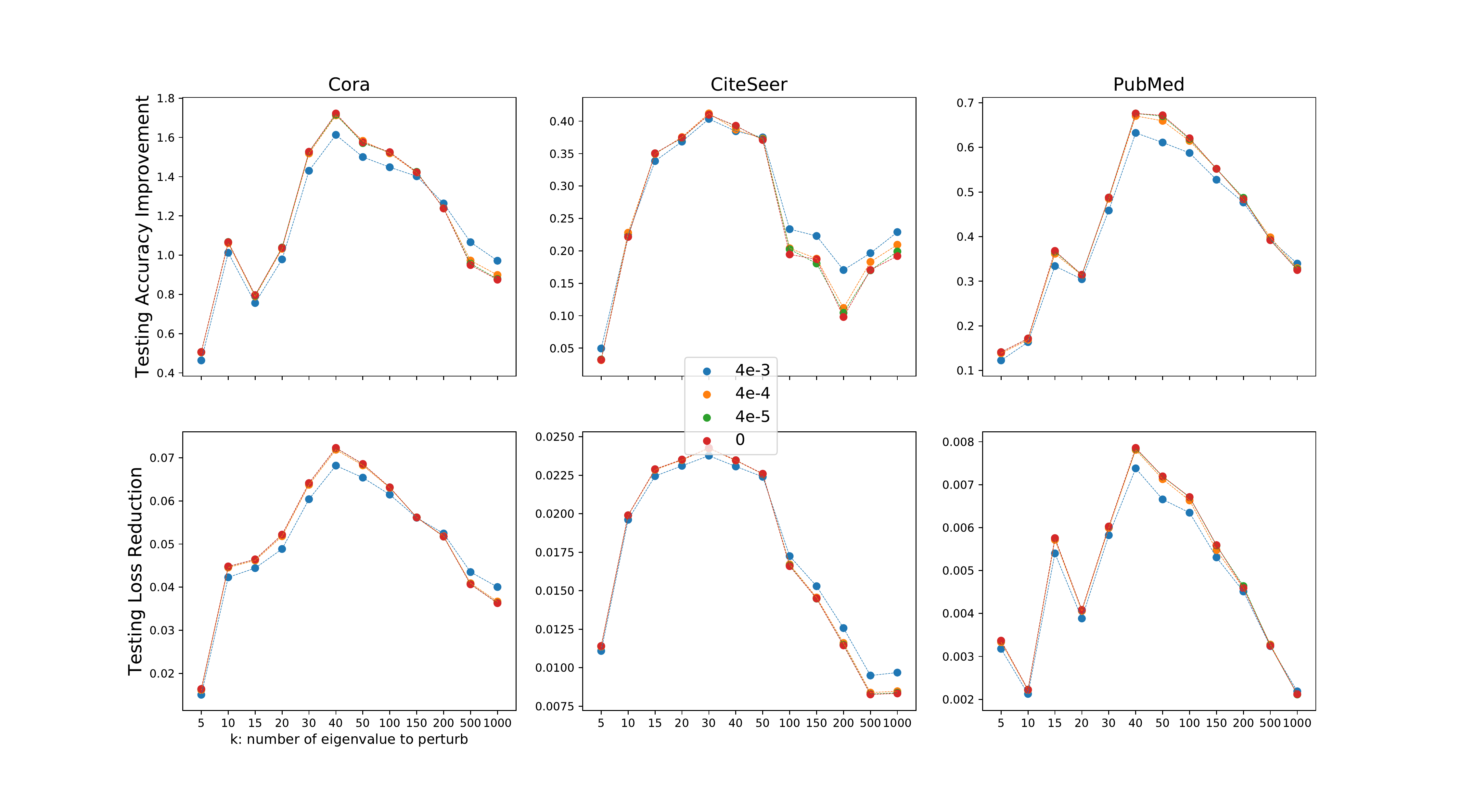}}
    \caption{EigLearn sensitivity on regularization.}
    \label{fig:reg_sensitivity}
\end{figure}

We also check the sensitivity on residual learning rate and regularization (weight decay),
as shown in \textbf{Figures~\ref{fig:lr_sensitivity}} and  \textbf{\ref{fig:reg_sensitivity}}, 
where we have GCN performance improvement versus hyperparameter value. 
In general EigLearn is quite robust toward these two hyperparameters. 
From the experimental study we find out indeed a smaller learning rate is helpful in 
learning the perturbation and this is consistent with the assumption that the perturbations
are small and should be learnt with a smaller learning rate. As for regularization, 
it turns out that smaller weight decay or even no weight decay leads to slightly better result,
although practically this does not make a big difference. 

\begin{figure}[htbp]
    \makebox[\textwidth][c]{\includegraphics[width=1.2\textwidth]{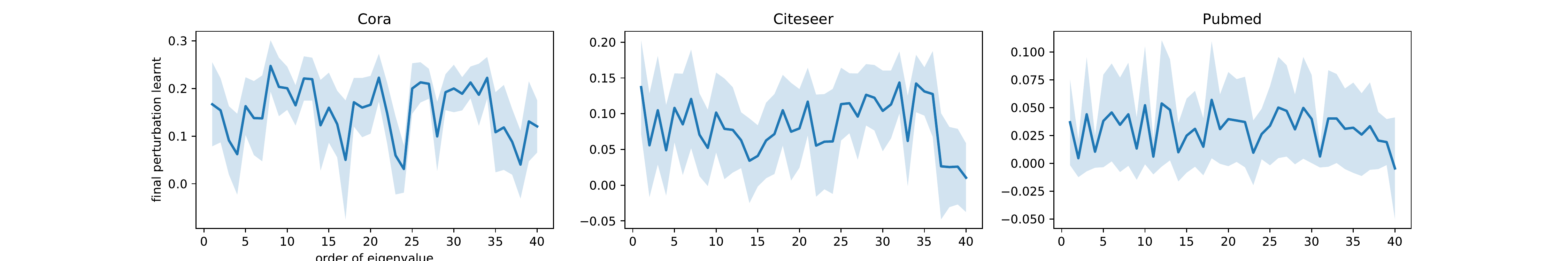}}
    \caption{Final perturbations on $S_{SNA}$.}
    \label{fig:residual}
\end{figure}

We also checked the final perturbation as shown in \textbf{Figure~\ref{fig:residual}}. Although there is some variance due to random data split and initialization, the pertubations tend to reside on one side of zero instead of reversing to zero. This also evidences that EigLearn provides systematic improvement rather than just randomly changing the eigenvalues. 
It is also worth pointing out that almost all perturbations tend to be positive. This behavior coincides with the philosophical design of low-pass filter \cite{nt2019revisiting}, 
which enhances the low-frequency band signals to a certain extent. 

%% file: chapter5.tex
\chapter{Neural Network Pruning for Better Efficiency}

\section{Motivation of Neural Network Pruning}
Neural network models have achieved remarkable performance in various application domains.  Nevertheless, a large number of weights in pre-trained large neural networks prohibit them from being efficiently deployed, especially on devices with limited resources such as smartphones and embedded systems.  It is highly desirable to obtain lightweight versions of neural networks for inference. 

Neural network pruning aims at removing a large number of parameters without significantly deteriorating the performance while benefiting from the reduced storage footprints for pre-trained networks and computing power. Formally, given a dense layer $\textbf{z}_{t} = \sigma (\textbf{z}_{t-1}^T A + \textbf{b})$,  where $\textbf{z}_{t-1} \in \mathbb{R}^m$ is the input signal, 
$\textbf{z}_{t} \in \mathbb{R}^n$ is the output signal, 
$A \in \mathbb{R}^{m\times n}$ is the weight matrix, 
$\textbf{b} \in \mathbb{R}^n$ is the bias, $\sigma$ denotes some activation function, we desire to obtain a sparse version of $A$ denoted by $\Tilde{A}$ such that $A$ and $\Tilde{A}$ have similar spectral structure and $\textbf{z}_{t-1}^T \Tilde{A}$ is as close to $\textbf{z}_{t-1}^T A$ as possible. Similarly for a convolution $z_t = T * z_{t-1}$ we want to find a sparse version of $T$ such that the convolution result is as close as possible, where $T$ could be a vector, matrix or higher-order array depending on the order of input signal and the number of output channel. By closeness, we use norms as metric. 

In this chapter, I will identify the close connection between matrix sparsification and neural network pruning for dense and convolutional layers, and argue that weight pruning is essentially a matrix sparsification process to preserve the spectrum. Based on the analysis, I also propose a matrix sparsification algorithm tailored for neural network pruning that yields better pruning result, and therefore provide a consolidated viewpoint for neural network pruning and enhance the interpretability of deep neural networks by identifying and preserving the critical neural weights.

\section{Neural Network Pruning and Matrix Sparsification}

\subsection{Neural Network Pruning as Spectrum Preserving Process}
In a dense layer, we focus on the $\textbf{z}^T A$ part since it contains most parameters.
Neural network is essentially a function simulator that learns some artificial features,
which is achieved by linear mappings, nonlinear activations, and some other customized units
(e.g., recurrent unit).
For the linear mapping, the analysis is usually done on the spectral domain. 

Recall that Singular Value Decomposition (SVD) is optimal under both spectral norm \cite{eckart1936approximation} and Frobenius norm \cite{mirsky1960symmetric}. 
The weight matrix $A \in \mathbb{R}^{m\times n}$ as a linear operator can be decomposed as
$$
A = U\Sigma V^T = \sum_{i=1}^{min(m,n)}\sigma_i \textbf{u}_i \textbf{v}_i^T
$$
where $U = [\textbf{u}_1,\textbf{u}_2,...,\textbf{u}_m] \in \mathbb{R}^{m \times m}$ is 
the left singular matrix, 
$V = [\textbf{v}_1, \textbf{v}_2,..., \textbf{v}_n] \in \mathbb{R}^{n\times n}$ is the right singular matrix and and $\Sigma$ contains the singular values $diag(\sigma_1, \sigma_2,..., \sigma_n)$
in a non-increasing order.

Note that a input signal $\textbf{z} \in \mathbb{R}^m$ can be written into
a linear combination of $\textbf{u}_i$, i.e., $\textbf{z}_i = \sum_i^m c_i \textbf{u}_i$ where
$c_i$ are the coefficients.
Thus, the mapping from $\textbf{z} \in \mathbb{R}^m$ to $\textbf{z}^{\prime} \in \mathbb{R}^n$ is
$$
\textbf{z}^{\prime} = \textbf{z}^TA 
= \sum_{i=1}^m c_i\textbf{u}_i^T \sum_{j=1}^{min(m,n)}\sigma_j \textbf{u}_j \textbf{v}_j^T
= \sum_{j=1}^{min(m,n)} c_j\sigma_j \textbf{v}_j
$$
where $\sigma_j$ are non-increasingly ordered, since
$\textbf{u}_i^T\textbf{u}_j=0, \forall i\neq j$
and $\textbf{u}_i^T\textbf{u}_i=1, \forall i$.

When we prune a neural network, we would like to preserve the spectrum of its weight matrix in order to preserve the neural network performance. In other words, we want to obtain a sparse $\Tilde{A}$ that has similar singular values to $A$. How to measure the wellness of spectrum preservation? We can use the spectral norm (2-norm) $\|A\|_2 = \sigma_1$ which is the largest singular value since we care about the dominant principle component, and the Frobenius norm (F-norm)
$$
\|A\|_F = (\sum_{i=1}^m\sum_{j=1}^n A_{ij}^2)^{1/2} = (Tr(A^TA))^{1/2} = (\sum_{i=1}^{min(m,n)} \sigma_i^2)^{1/2}
$$
which is usually considered an aggregation of the whole spectrum. 
Note that $\|A\|_2 \leq \|A\|_F$ and $\|A\|_F \leq \sqrt{min(m,n)}\|A\|_2$. 

Therefore, the goal is to find a sparse $\Tilde{A}$ such that 
$\|A - \Tilde{A}\|_2 \leq \epsilon$ or $\|A - \Tilde{A}\|_F \leq \epsilon$.

\subsection{Matrix Sparsification Algorithms}
Matrix sparsification is important in many numerical problems, e.g., low-rank approximation, semi-definite programming and matrix completion, which widely exist in data mining and machine learning problems. Matrix sparsification is to reduce the number of nonzero entries in a matrix without altering its spectrum. The original problem is NP-hard \cite{mccormick1983combinatorial}\cite{gottlieb2010matrix}. The study of approximation solutions to this problem was pioneered by \cite{achlioptas2007fast}, and further expanded in \cite{achlioptas2013near} \cite{arora2006fast} \cite{achlioptas2013matrix} \cite{nguyen2009matrix} \cite{drineas2011note}. An extensive study on the error bound was done in \cite{gittens2009error}. 

Since the spectrum of the sparsified matrix does not deviate significantly from that of the original matrix, serving as a linear operator the matrix retains its functionality,  i.e., 
$A\in \mathbb{R}^{m\times n}$ is a mapping $\mathbb{R}^m \rightarrow \mathbb{R}^n$. 
We can define the matrix sparsification process as the following  optimization problem: 
\begin{equation}
\begin{aligned}
& \underset{}{\text{min}}
& & \| \Tilde{A} \|_{0} \\
& \text{s.t.}
& & \| A - \Tilde{A}\| \leq \epsilon \\
\end{aligned}
\label{eqn:linear-combine2}
\end{equation}
where $A$ is the original matrix, $\Tilde{A}$ is the sparsified matrix, 
$\|\cdot\|_{0}$ is the $0$-norm that equals the number of non-zero entries in a matrix, $\|\cdot\|$ denotes matrix norm, $\epsilon \geq 0$ is the error tolerance. 

In matrix sparsification, we often use the spectral norm (2-norm) $\|\cdot\|_{2}$ and the Frobenius norm (F-norm) $\|\cdot\|_{F}$ to measure the deviation of the sparsified matrix from the original one.

Magnitude-based neural network pruning have attracted a lot of attention and show supprisingly simplicity and superior efficacy. 
In the context of matrix sparsification, this is a straightforward approach, namely magnitude-based matrix sparsification or hard thresholding. 
Given a matrix $A$, let $\Tilde{A}$ denote its sparsifier. Entry-wise we have
\[\Tilde{A}_{ij} = \begin{cases}
                A_{ij} &  |A_{ij}| > t \\
                0 &   else
                \end{cases}
\].
\begin{remark}\label{f_opt}
Magnitude based thresholding always achieves sparsification optimality in terms of F-norm. 
\end{remark}
The fact can be trivially verified since using $|A_{ij}|$ and using $A_{ij}^2$
(on which F-norm is based)
are equivalent in terms of deciding small entries in a matrix.
However throwing away small entries does not always guarantee
the optimal sparsification result in terms of 2-norm. And in many situations,
we care more about the dominant singular value instead of the whole spectrum.

In randomized matrix sparsification, each entry is sampled according to some distribution independently and then rescaled. For example, each entry is sampled according to a Berboulli distribution, and we either set it to zero or rescale it. 
\[\Tilde{A}_{ij} = \begin{cases}
                A_{ij}/p_{ij} &  p_{ij} \\
                0 &   1-p_{ij}
                \end{cases}
\]
where $p_{ij}$ can be a constant or positively correlated to the magnitude of the entry.
The following theorem provides the justification to this type of matrix sparsification.

\begin{theorem}\label{the1}
A matrix where each entry is sampled from a zero-mean bounded-variance distribution possesses weak spectrum with large probability. 
\end{theorem} 
By weak spectrum, it means small matrix norm. To be more concrete, since matrix norm is a metric and triangle inequality applies,
we have
$$
\|A\| \leq \|\Tilde{A}\| + \|A-\Tilde{A}\|.
$$
We need to show that
$N = A-\Tilde{A}$ falls within the category of matrices described in Theorem \ref{the1}. Since
$$
E(N_{ij}) = E(A_{ij} - \Tilde{A}_{ij}) = A_{ij} - A_{ij}/p_{ij}\cdot p_{ij} = 0
$$
and
$$
var(N_{ij}) = var(\Tilde{A}_{ij}) = (A_{ij}/p_{ij})^2 \cdot p_{ij} = A_{ij}^2 / p_{ij},
$$
as long as $A_{ij}^2 / p_{ij}$ is upper-bounded, which is true most of time,
$var(N_{ij})$ is bounded. Therefore, the
randomized matrix sparsification can guarantee the error bound.

\subsection{Customize Matrix Sparsification Algorithm for Neural Network Pruning}

In this section, we propose a customized matrix sparsification algorithm to show the potential of 
designing a better spectrum preservation process in neural network pruning. We do not intend to
present a new state-of-the-art neural network pruning algorithm. There are two important points in our proposed algorithm: truncation and sampling based on the principal components of  explicit truncated SVD.
To the best of our knowledge,   sampling based on probability proportional to principal components is employed for the first time in designing matrix sparsification for neural network pruning . 

First, we adopt the truncation trick that is common in existing work. As clearly pointed out by \cite{achlioptas2013matrix}, the spectrum of the random matrix $N = A - \Tilde{A}$ is determined by its variance bound. Usually, the larger the variance, the stronger the spectrum of the random matrix. Existing works took advantage of the finding and proposed truncation \cite{arora2006fast}\cite{drineas2011note} in sparsification, i.e., to set small entries to zero while leaving large entries as is and sampling on the remaining ones. 
\[\Tilde{A}_{ij} = \begin{cases}
                A_{ij} & |A_{ij}| > t \\
                0  & p_{ij} < c \\
                A_{ij}/p_{ij}\cdot Bern(p_{ij}) & else 
                \end{cases}
\]
where $p_{ij} \propto |A_{ij}|$, $t$ is decided by the quantile (leave large entries as is), 
and $c$, the lower threshold for zeroing weights, as a constant could be set manually, 
and $Bern(\cdot)$ denotes Bernoulli distribution.

Second, instead of sampling based on the probability calculated from the magnitude of the original matrix entry, 
we do sampling based on the probability calculated from the principal component matrix entry magnitude
with a little compromise on complexity, 
in order to better preserve the dominant singular values. 
Matrix sparsification was originally proposed for fast low-rank approximation on very large matrices, 
due to the fact that sparsity accelerates matrix-vector multiplication in power iteration. Essentially, 
we desire to find the sparse sketch $\Tilde{A}$ of $A$ that preserves the dominant singular values well. 
This coincides with the goal of layer-wise neural network pruning from the spectrum preserving viewpoint --
we desire to preserve the dominant singular values, based on the fact that we often consider
information lies in the low-frequency domain while noises are in the high-frequency domain. 
The major difference is that weight matrices in neural network, either from dense layers or convolutional layers,
are usually not too large, 
and therefore explicit SVD or truncated SVD on them is fairly affordable. Once we have access to the principle components
of the weight matrices, we are able to preserve them better in the sparsification process. 
Note that preserving dominant singular values is a harmonic approach between preserving the 2-norm and the F-norm, 
since $\|A\|_2 = \lim_{p\to\infty} (\sum_i \sigma_i^p)^{1/p}$ and $\|A\|_F = \lim_{p\to2} (\sum_i \sigma_i^p)^{1/p}$. 

The crutial part is to find the low-rank approximation $B$ to $A$, where 
$
B = \sum_{i=1}^K \sigma_i\textbf{u}_i\textbf{v}_i^T
$ and $\sigma_i \textbf{u}_i \textbf{v}_i^T$ are from SVD on A.
We set the entry-wise sampling probability based on $|B_{ij}|$, i.e., $p_{ij} \propto |B_{ij}|$. 
Algorithm \ref{Sparsify} presents the sparification algorithm. The \textit{partition} function is the one used in \textit{quicksort}. 
\begin{algorithm}[ht]
    \SetKwFunction{Union}{Union}\SetKwFunction{FindCompress}{FindCompress}\SetKwInOut{Input}{input}\SetKwInOut{Output}{output}
    
    \Input{($A$, $c$, $q$, $K$)} 
    \Output{$\Tilde{A}$} 
    \BlankLine  \tcp*{$q$: quantile above which remain unchanged}
    $B =$ \textit{truncated-SVD($A$, $K$)}; \tcp*{$B$: low-rank approx to $A$}
    $m,n$ = shape of $B$; \\
    $t$ = partition($\{|B_{ij}|\}$, int($m\times n \times q$)); \\
    \For{$i\gets1$ \KwTo $m$}{
        \For{$j\gets1$ \KwTo $n$}{
            \uIf{$|B_{ij}| < t$}{
                $p_{ij} = (B_{ij}/t)^2$; \\
                \uIf{$p_{ij} < c$}{
                    $A_{ij} = 0$; \\
                }
                \uElse{
                    $A_{ij} = A_{ij}/p_{ij}\cdot Bern(p_{ij})$; \\
                }
            }
        }
    }
    \caption{Sparsify}\label{Sparsify}
\end{algorithm}

Here we provide a high level proof on sparsification error being upper bounded. 
Let $D$ denote the sparse sketch generated by setting smallest entries in $A$ to 0 and $\Tilde{A}$ as usual the final sparsfied result. 
From Fact \ref{f_opt} we know
that $D$ is the optimal sketch of $A$ in terms of F-norm, i.e., $\|A-D\|_F = \epsilon_*$.
Based on Theorem \ref{the1} and its illustration we know that 
$N=\Tilde{A}-D$ satisfies the zero-mean
and bounded-variance condition. Hence, $\|\Tilde{A}-D\|_F \leq \epsilon_0$. Therefore, 
if we apply triangle equality given matrix norm is a metric, 
$$
\|A-\Tilde{A}\|_F \leq \|A-D\|_F + \|D-\Tilde{A}\|_F \leq \epsilon_* + \epsilon_0. 
$$
Some other techniques, e.g., quantization\cite{gong2014compressing}\cite{han2015deep}, 
can be used together  with 
sparsification to further compress matrices and neural networks. Essentially they are 
also spectrum preservation techniques \cite{achlioptas2007fast}\cite{arora2006fast}.

\section{Generalization to Convolution}

Extensive literatures argue that convolutional layers compression can be formalized as tensor algebra problems \cite{lebedev2014speeding}\cite{denton2014exploiting} \cite{kim2015compression}\cite{liu2015sparse}\cite{sun2016sparsifying}. 
However, it is advantageous to explain convolutional layer pruning from the matrix viewpoint since the linear algebra have many nice properties that do not hold for multilinear algebra. We want to ask: can we still provide theoretical support to convolutional layer pruning 
using linear algebra we have discussed so far?

\begin{figure}[htbp]
  \centering
  \includegraphics[width=0.9\textwidth]{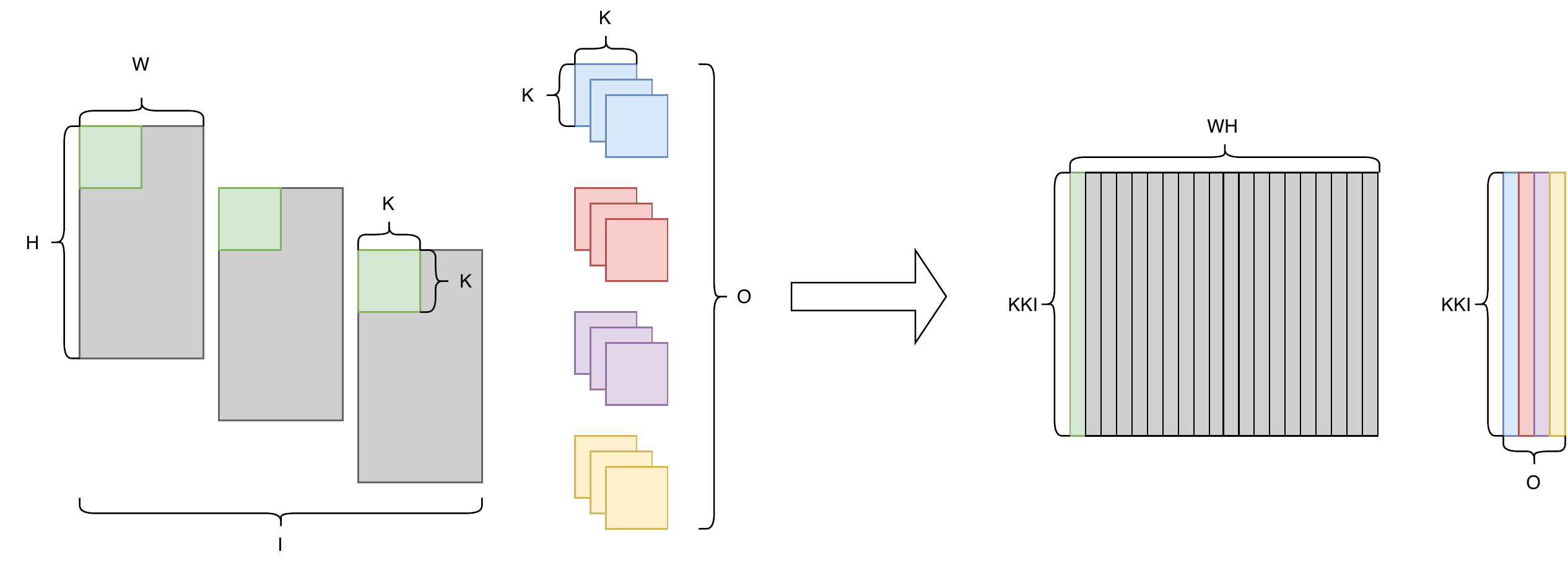}
  \caption{Convolution as dense matrix multiplication.}
  \label{cnn as mat}
\end{figure}

\subsection{Pruning on Convolutional Filters}
In this section we state and illustrate the following fact. 
\begin{remark}\label{CNN mat mult}
Discrete convolution in neural networks can be represented by dot product between two dense matrices. 
\end{remark}

To see this, suppose we have a convolutional layer with input signal size of $W\times H$ as width by height, and with $C$ input channels and $O$ output channels. Here we consider a 2-d convolution on the signal. The kernel is of size $C \times K \times K$ and there are $O$ such kernels. For the sake of simplicity in notations, suppose the striding step is 1, half-padding is applied and there is no dilation (for even $W$ and $H$ the above setting results in output signal of size $W\times H$ as width by height). 2-d convolution means that the kernel is moving in two directions. Fact \ref{CNN mat mult} has been utilized to optimize lower-level implementation of CNN on hardware \cite{chellapilla2006high}\cite{chetlur2014cudnn}. Here we take advantage of the idea to unify neural network pruning on dense layers and convolutional layers with matrix sparsification. 

Let us focus on one single output channel, one step of the convolution operation is the summation of element-wise product of two higher-order array, i.e., the kernel $G \in \mathbb{R}^{C\times K\times K}$ and 
the receptive field of the signal of the same size $X \in \mathbb{R}^{C\times K\times K}$. Note that taking the summation of element-wise product is equivalent to
vector inner product. Therefore, if we unfold the kernel for a single output channel to a vector and rearrange the receptive field of the signal accordingly to another vector, a single convolution step can be treated as two vector inner product, i.e., $G*X = \textbf{g}^T\textbf{x}$ where $\textbf{g}, \textbf{x} \in \mathbb{R}^{CKK}$. Since we have $O$ output channels in total, there are $O$ such kernels of the same size. All of them being unfolded, we then can convert the convolution into a matrix product $Z^TA$, where $A \in \mathbb{R}^{CKK \times O}$ being the kernels and $Z \in \mathbb{R}^{CKK \times WH}$ being the rearranged input signals. And consequently the output signal $Y \in \mathbb{R}^{WH \times O}$ 
(as mentioned before, stride 1, half padding and no dilation result in 
input signal and output signal being in the same shape). 
\textbf{Figure \ref{cnn as mat}} visualizes convolution as matrix multiplication.

The matrix multiplication representation of convolution discussed above generalizes to any other convolution settings. 
Also note that the way we unfold the filters does not affect the spectrum of the resulting matrix, 
since row and column permutations do not change matrix spectrum. 
Therefore, all the analyses based on simple linear algebra we have discussed so far generalize to convolutional layer pruning. 
Another way to characterize the discrete convolution is to leverage the doubly-block circulant matrix, and the singular values can be calculated using fast fourier transform\cite{sedghi2018singular}. This is another line of work under investigation for neural network pruning.

\subsection{Convolutional Filter Channel Pruning}

Entry-wise pruning almost always results in unstructured sparsity
that requires specific data structure design in network deployment in order
to realize the complexity reduction from pruning. Therefore, it is desirable 
to prune entire channels from convolutional layers to achieve higher efficiency.  There is another important work on pruning channels \cite{li2016pruning}. The approach is to take small $\sum_{ijk}|T_{ijk}|$ where $T$ denotes the filter for a specific channel. This is equivalent to remove a column in $A$ we just discussed with small-magnitude values. It is also a spectrum preserving process as $\sum_{ijk}|T_{ijk}|$ is fairly a proximity to $\sum_{ijk}(T_{ijk})^2$ on which the F-norm is based. Hence, pruning the whole filter with small $\sum_{ijk}|T_{ijk}|$ is to preserve the F-norm of the convolution matrix we discussed in the previous subsection.

\section{Graph Sparsification in GCN}
In additional to removing parameters in neural units, one can also remove edges from the graph or elements from the associated graph filter matrix~\cite{srinivasa2020fast} leveraging graph sparsification techniques~\cite{spielman2011spectral}\cite{koutis2015faster} in order to speed up graph neural network training, inference and broaden the application scenarios. Concretely, given a graph $\mathcal{G}$, the graph sparsification process generates another graph $\mathcal{\Tilde{G}}$ with fewer edges that is spectrally similar to the original graph $\mathcal{G}$. The spectral similarity is defined on quadratic form, i.e., for all vestors $x$ and any $\epsilon > 0$, we have
\begin{equation}
    (1-\epsilon)x^T L_{\mathcal{G}}x < x^T L_{\Tilde{\mathcal{G}}}x < (1+\epsilon)x^T L_{\mathcal{G}}x, 
\end{equation}
where $L_{\mathcal{G}}$ and $L_{\Tilde{\mathcal{G}}}$ are the Laplacian matrices of $\mathcal{G}$ and $\Tilde{\mathcal{G}}$ respectively. The key in graph sparsification is the fast computation of effective resistance, which is a meaningful measurement in electronic networks. Combined with sampling strategies based on effective resistance, one can remove unimportant edges and therefore reduce the number of elements in the graph filter matrix while preserving the algebraic properties of the graph filter matrix serving as an operator in GCN. 

\section{Empirical Study}

The experiments are mainly based on LeNet \cite{lecun1998gradient} on MNIST and VGG19 \cite{simonyan2014very} on CIFAR10 dataset \cite{krizhevsky2009learning}.  We trained the neural networks from scratch based on the official PyTorch \cite{paszke2019pytorch} implementation.  Then we conducted our experiments based on the pre-trained neural networks. 

We trained LeNet with a reduced number of epochs of 10. All other hyperparameter settings are the ones used in the original implementation of the PyTorch example. The VGG19 was trained with the following hyperparameter setting: batch size 128, momentum 0.9, weight decay $5e^{-4}$,  and the learning rates of 0.1 for 50 epochs, 0.01 of 50 epochs, and 0.001 of another 50 epochs.  The final testing accuracy for LeNet on MNIST and VGG19 on FICAR10 was 99.14\% and 92.66\%, respectively.

\begin{figure}[t]
  \centering
  \includegraphics[width=.7\textwidth]{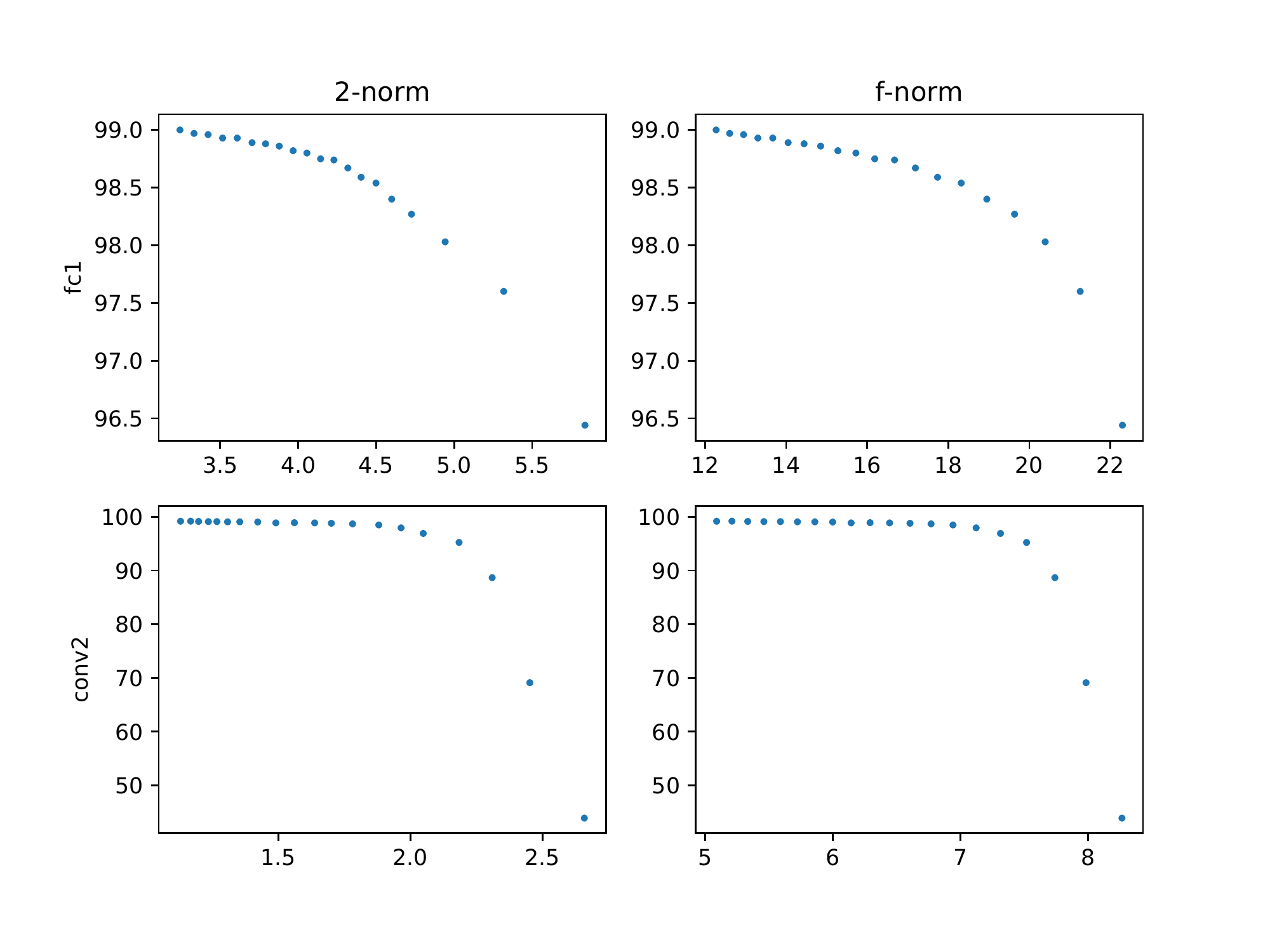}
  \caption{$\|A-\Tilde{A}\|_2$ and $\|A-\Tilde{A}\|_F$ vs. 
  neural network accuracy as the sparsity increases (LeNet on MNIST).}
  \label{lenet sparsity acc}
\end{figure}

\begin{figure}[htbp]
  \centering
  \includegraphics[width=.7\textwidth]{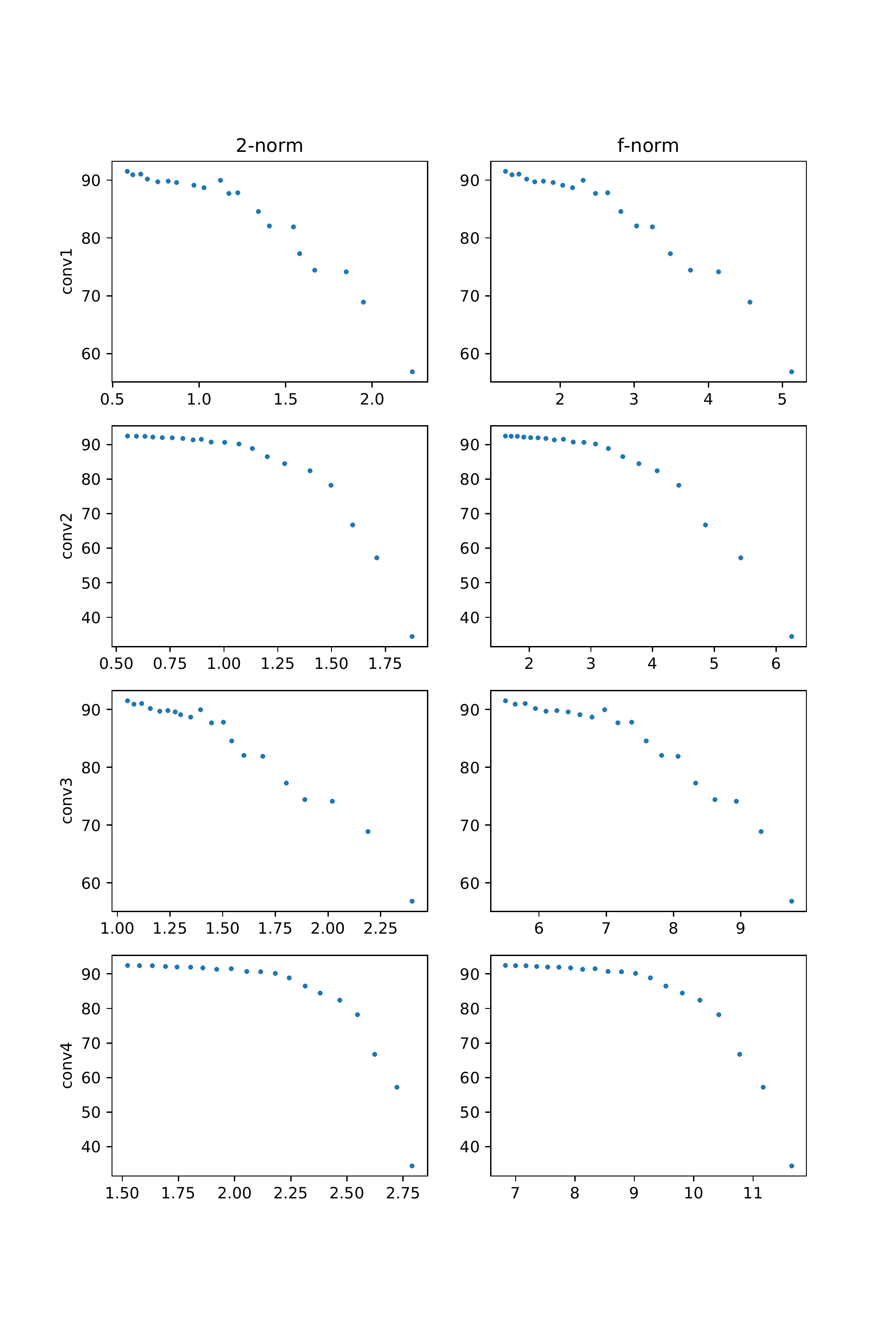}
  \caption{$\|A-\Tilde{A}\|_2$ and $\|A-\Tilde{A}\|_F$ vs. 
  neural network accuracy as sparsity increases (VGG19 on CIFAR10).}
  \label{vgg sparsity acc}
\end{figure}

\begin{figure}[htbp]
  \centering
  \includegraphics[width=0.8\textwidth]{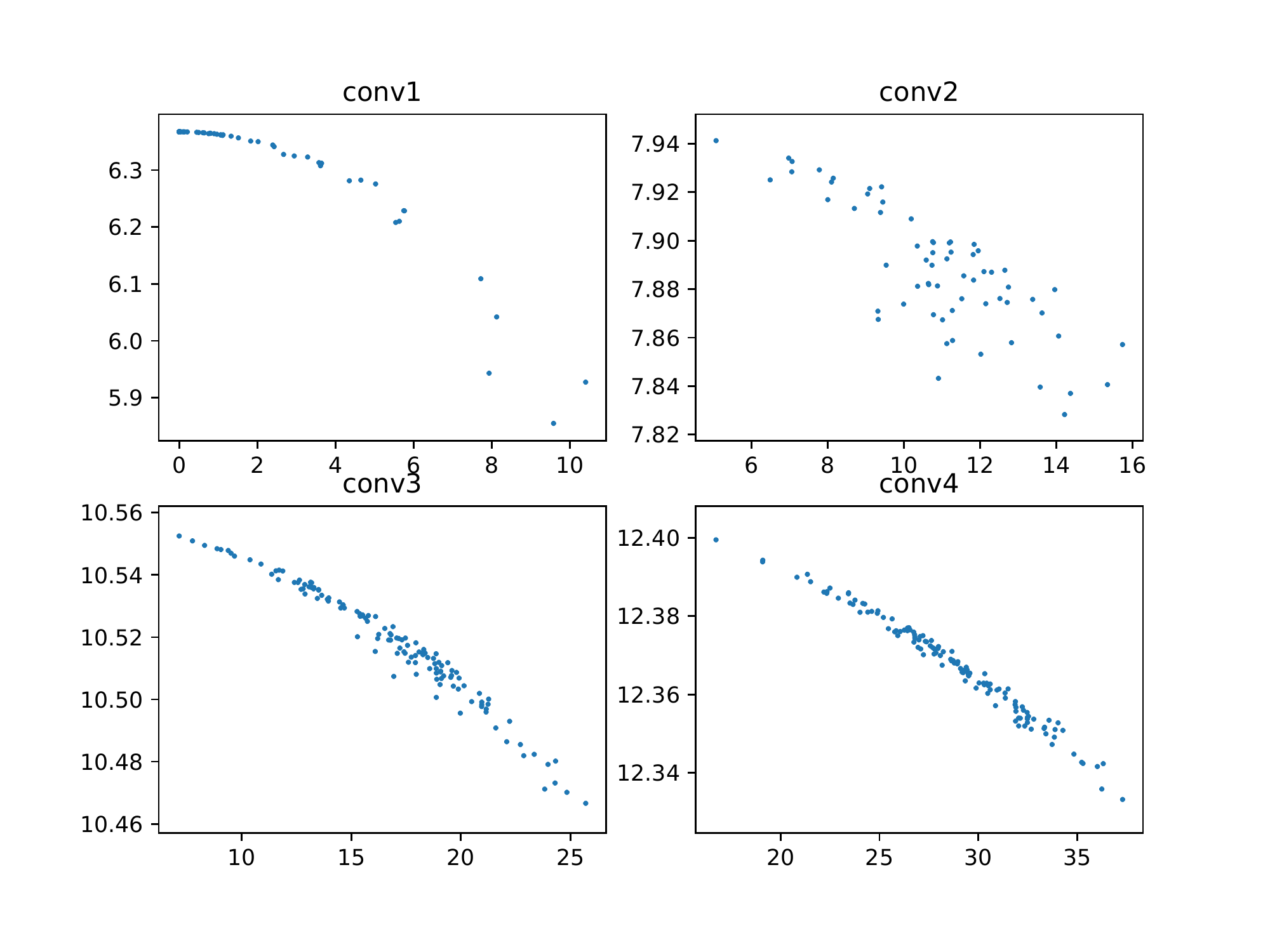}
  \caption{VGG19 channel pruning based on $\sum_{i}|T_i|$. 
 }
  \label{channel pruning}
\end{figure}

We investigated the relationship between spectrum preservation and the performance of the pruned neural network.  Matrix sparsification (hard thresholding) was employed to prune LeNet and VGG19.  We varied the percentage of parameter preservation from 20\% to 1\% to get different sparsities (the sparser,  the larger $\|A-\Tilde{A}\|_2$ and $\|A-\Tilde{A}\|_F$).  We perform dense layer pruning (\textbf{Figure~\ref{lenet sparsity acc}}), convolution layer pruning (\textbf{Figure~\ref{vgg sparsity acc}}) and convolution channel pruning (\textbf{Figure~\ref{channel pruning}}) respectively, and check the corresponding 2-norm and F-norm. 

\textbf{Figure~\ref{lenet sparsity acc}} shows the experiment result of LeNet on MNIST. We prune a dense layer denoted by ``fc1" and a convolution layer denoted by ``conv2". The x-axis denotes the norm and y-axis denotes the accuracy of the pruned neural network. The figure shows a consistent pattern that as the sparsity increases, the spectrum measured by both 2-norm $\|A-\Tilde{A}\|_2$ and F-norm $\|A-\Tilde{A}\|_F$ deviates from its origin, and the neural network performance keeps decreasing. \textbf{Figure~\ref{vgg sparsity acc}} shows the experiment result of VGG19 on CIFAR10. We prune 4 of its convolution layers denoted by ``conv1" to ``conv4". The pattern is similar to that in \textbf{Figure~\ref{lenet sparsity acc}}. \textbf{Figure~\ref{channel pruning}} shows the result of convolution channel pruning based on $\sum_{i}|T_i|$.  y-axis denotes $\|\Tilde{A}\|_F$ and x-axis denotes $\sum_{i}|T_i|$. The smaller $\sum_{i}|T_i|$, the smaller $\|A-\Tilde{A}\|_F$,  the larger $\|\Tilde{A}\|_F$, the better spectra are preserved. Hence, our analysis bridges the gap between small $\sum_{i}|T_i|$ and good neural network performance preservation.

We applied Algorithm~\ref{Sparsify} on all convolutional layers in VGG19 at the same time, varied algorithm
settings to get different sparsities, recorded the corresponding testing performance of the pruned network, 
and compared with the performance of the pruned network via thresholding at the same sparsity level. 
Due to the randomness in our proposed algorithm, the sparsity in different layers is also different. 
We present the aggregated sparsity, i.e., the total number of nonzero parameters divided by total number of parameters 
in all convolutional weight matrices, in our empirical study result. To ease the implementation and focus on 
our arguments, we fixed parameter $c=0.5$, varied the quantile parameter $q$ and the number of principal components $K$. 

\begin{figure*}[htbp]
  \centering
  \includegraphics[width=\textwidth]{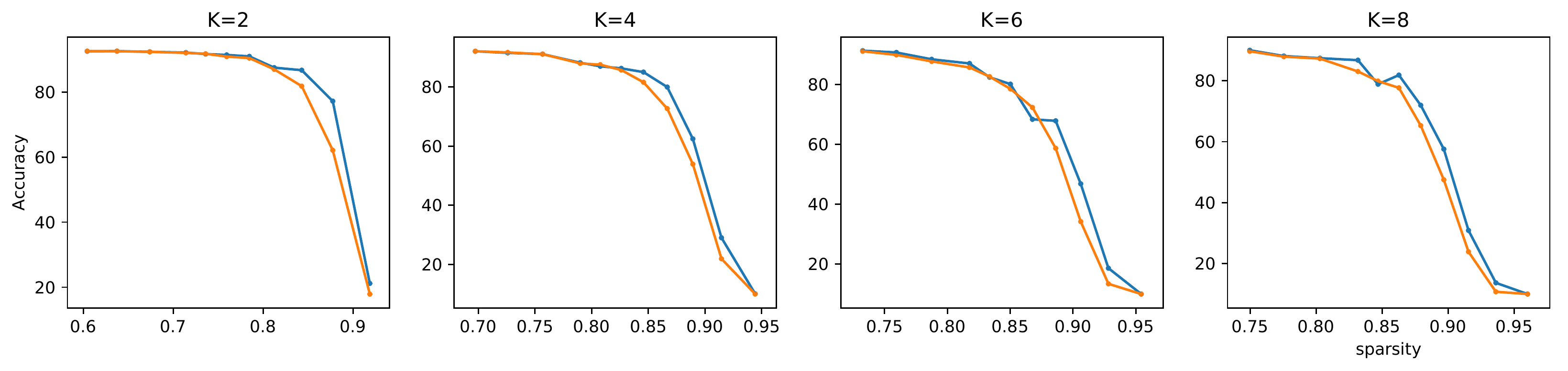}
  \caption{Pruned network testing performance given by magnitude-based thresholding (orange) vs. Algorithm\ref{Sparsify} (blue). }
  \label{acc compare}
\end{figure*}

From \textbf{Figure~\ref{acc compare}} we can see that, our proposed algorithm almost always leads to better pruned network 
generalization performance without retraining compared to that given by thresholding, at different sparsity levels.
This demonstrates the potential of designing and customizing matrix sparsification algorithms 
for better neural network pruning approaches. In addition, we also observed Algorithm~\ref{Sparsify} almost always 
yield smaller sparsification error compared to thresholding in terms of 2-norm, which is exactly the motivation of the algorithm design.

%% file: chapter6.tex
\chapter{Discussion and Future Work}

Machine learning has enabled the wide application of big-data based recommendation systems in many business scenarios and helped us with high-efficiency decision making in all aspects. However, the applicability of such approaches can be severely limited due to knowledge scarcity in the specific domains of interest. In this work, we tackle such challenges by first formulating the problem of cross-domain knowledge transfer, where the targeting domain leverages the rich information readily available from another domain. We then illustrate how graph based methods can solve the problem by capturing the complex data distribution and propose new methods to better solve concrete application problems, such as embedding imputation in financial data analyses. We dive into the methodological study for the related graph-related methods, analyze the mechanism from the spectral graph perspective, and further improve the advances such as graph neural networks. In addition to efficacy, we also address the efficiency issue that naturally exists in graph learning and large neural networks with fast graph-related algorithms and neural network pruning, such that the methods can be leveraged in more daily business operations where data are tremendously growing. 

For future work, we are especially interested in the following directions: 
\begin{itemize}
    \item We are interested in spectral regularization in deep neural networks. The goal here is to improve neural network generalization performance via spectral manipulation. 
Some works have shown promising results, for example, bounding the Lipschitz constant of neural network via bounding layer-wise spectral norm for more stable training and better generalization\cite{yoshida2017spectral}\cite{gouk2018regularisation}, performing dropout in the spectral domain instead of the regular domain for better neural network regularization\cite{khan2019regularization}, imposing constraints (e.g., fix the set of singular vectors) on weight matrix singular vectors to improve neural network performance\cite{jia2017improving}.
    \item We are also interested in interpretibility~\cite{du2019techniques} in machine learning. As machine learning models become more complicated and neural networks are widely adopted, it is essential to understand the mechanisms, such that the model design can be further motivated. The interpretibility of machine learning also guides us what methods fit what data and problems. 
    \item Finally, we want to apply the recent advances in machine learning research and our study in more business problems and further demonstrate their practical values. To be more specific, we want to apply network-based approach to solve the problems related to (1) cold-start problems in recommendation system on social network where new users do not have much information (2) user taste exploration in content consumption where we leverage the community interests to improve user consumption experience. 
\end{itemize}